\documentclass[twoside]{article}
\usepackage{ecj,palatino,epsfig,latexsym,natbib}
\usepackage{hyperref}
\usepackage{amsmath}
\usepackage{amssymb}
\usepackage{tabulary}
\usepackage{multirow}
\usepackage{booktabs}

\newcommand{\sphinxcode}[1]{{\ttfamily #1}}
\newcommand{\sphinxstyleemphasis}[1]{{\emph{#1}}}

\newcommand{\sphinxstylestrong}[1]{{\bf #1}}
\newcommand{\sphinxstyleliteralintitle}[1]{{\ttfamily #1}}
\newcommand{\sphinxincludegraphics}[2]{\sphinxincludegraphics[#1]{#2}}
\newcommand{\code}[1]{{\ttfamily #1}}

\begin{document}

\ecjHeader{x}{x}{xxx-xxx}{201X}{
%   5   10        20        30        40   45
Using SO-Functions in MOO Test Suites
}{D.~Brockhoff et al.}
\title{\bf Using Well-Understood Single-Objective Functions in Multiobjective Black-Box Optimization Test Suites}  

\author{\name{\bf Dimo Brockhoff} \hfill \addr{dimo.brockhoff@inria.fr}\\ 
        \addr{Inria, research centre Saclay and\\
              CMAP UMR 7641 \'{E}cole Polytechnique CNRS, France}
\AND
       \name{\bf  Tea Tu\v{s}ar} \hfill \addr{tea.tusar@ijs.si}\\
        \addr{Jo\v{z}ef Stefan Institute, Ljubljana, Slovenia}
\AND
			 \name{\bf  Anne Auger} \hfill \addr{anne.auger@inria.fr}\\
        \addr{Inria, research centre Saclay and\\
              CMAP UMR 7641 \'{E}cole Polytechnique CNRS, France}
\AND
			 \name{\bf  Nikolaus Hansen} \hfill \addr{nikolaus.hansen@inria.fr}\\
        \addr{Inria, research centre Saclay and\\
              CMAP UMR 7641 \'{E}cole Polytechnique CNRS, France}
}

\maketitle

\begin{abstract}
Several test function suites are being used for numerical benchmarking of multiobjective
optimization algorithms. While they have some
desirable properties, like well-understood Pareto sets and Pareto fronts of various shapes,
most of the currently used functions possess characteristics
that are arguably under-represented in real-world problems. They
mainly stem from the easier construction of such functions and result in improbable
properties such as separability, optima located exactly at
the boundary constraints, and the existence of variables that solely control the distance between a
solution and the Pareto front. Here, we propose an alternative way to constructing multiobjective
problems---by combining existing single-objective problems from the literature. We describe
in particular the \sphinxcode{bbob-biobj} test suite with 55 bi-objective functions in
continuous domain, and its extended version with 92 bi-objective functions (\sphinxcode{bbob-biobj-ext}).
Both test suites have been implemented in the \href{https://github.com/numbbo/coco}{COCO} platform for black-box optimization benchmarking.
Finally, we recommend a general procedure for creating test suites for an arbitrary number of
objectives. Besides providing the formal function definitions and presenting their
(known) properties, this paper also aims at giving the rationale behind our
approach in terms of groups of functions with similar properties, objective space normalization,
and problem instances. The latter allows us to easily compare the
performance of deterministic and stochastic solvers, which is an often overlooked issue in
benchmarking.

\end{abstract}


\section{Introduction}
\label{\detokenize{index:introduction}}
Numerical benchmarking is an important part of (black-box) optimization that helps to understand
algorithm behavior and recommend algorithms. In order to obtain meaningful results, a
benchmarking experiment should be (i) based on a thorough, well-documented and well-understood
methodology and (ii) either be conducted on real-world problems of interest or a collection of artificial test
functions that possess comprehensible difficulties observed in practical optimization problems. This holds true
for both single- and multiobjective problems but for the latter, the methodology is less advanced
at the moment.

Many artificial test functions that are frequently used in multiobjective optimization have been derived
by setting up the Pareto front shape first without relating it to the intrinsic difficulties of
the objective functions. Such an approach has the advantage that the analytical forms of the Pareto
front and the Pareto set can be exploited to facilitate the performance assessment. Another aspect of
state-of-the-art test suites for multiobjective optimization is the fact that not much progress
has been made to avoid the overrepresentation of functions that are too simple or have questionable
properties. Several existing (and still frequently used) multiobjective test suites, for example,
contain a large share of functions that are separable, have the Pareto set on the domain boundary, or contain distance
and position variables %
\footnote{
A function is said to have a \sphinxstyleemphasis{distance variable} if changing this variable only results in
dominating or dominated solutions. In other words, a distance variable determines solely the
distance of a solution from the Pareto front. A \sphinxstyleemphasis{position variable}, in turn, only results in
incomparable solutions when changed (see \cite{hhbw2006a} for details).
}
---artificial features not reflecting well the difficult black-box
problems observed in practice.

The most complete reference for multiobjective test function suites is the work of \citet{hhbw2006a}. The authors not only review all available test suites at the time of writing
(not many new have been introduced thereafter), but also give general advice on the desired properties of
multiobjective test functions. Based on these recommendations, Huband et al. finally
propose a generic test function generator and use it to create the WFG function suite with
nine scalable test functions. These test functions, however, are constructed in a similar manner as the
above mentioned suites with the shape of the Pareto front being the first design criterion. After that,
transformations in the search and objective space give the functions some desired properties,
like non-separability and multimodality.

In the context of single-objective algorithm benchmarking, a lot of progress has been made in
recent years in the design of artificial test functions that represent a wide range of difficulties
observed in practice. The black-box optimization benchmarking test suite (\sphinxcode{bbob}, \citet{hafr2009a}) in
particular has received wide acceptance as its 24 test functions have various advantages over
previous test suites. The functions are well understood and expose algorithms to a variety of
real-world difficulties such as multimodality, ill-conditioning, non-separability of the variables,
and non-linearities. The \sphinxcode{bbob} functions are grouped into five function groups with functions
within a group sharing similar difficulties (such as multimodality with weak global structure) and
with the aim to not overemphasize certain difficulties. Each function has one or several concrete
scientific questions associated with it that can be answered by looking at algorithm performance results on
that function (or in combination with another function). General statements beyond the tested
concrete functions are possible by testing invariance properties of algorithms such as scaling,
rotation and affine invariance. The \sphinxcode{bbob} functions also come in the form of instances which
allows us to easily compare deterministic and stochastic algorithms (see
Section~\ref{sec:bbobinstances}).

In contrast to the previously mentioned approaches to building multiobjective test suites, we suggest to
focus on introducing the known difficulties of real-world problems into the test suite.
This is analogous to the single-objective case, but has the disadvantage that analytical formulas for
the Pareto front and Pareto set might not be available. The motivation behind this approach is
that in practice, multiobjective problems are constructed in exactly this way---with each objective
corresponding to a separate single-objective function.
Concretely, we propose a generic way to combine the well-understood single-objective functions
from the \sphinxcode{bbob} test suite \citep{hafr2009a}. Using the well-established \sphinxcode{bbob} test functions as building
blocks allows us to build upon a careful statistical choice of the functions (without
overrepresenting a certain type of problem) as well as comprehensive difficulties. In particular,
our proposal fulfills all five recommendations for benchmark suites mentioned by
Huband et al. (\citet{hhbw2006a}, page 485). We showcase our idea by implementing two bi-objective test
suites within the COCO platform \citep{han2016coco} that supports automated benchmarking. The
disadvantage of having no analytical expressions for the Pareto sets and Pareto fronts in our
approach is addressed by visually displaying the approximations of Pareto sets and Pareto fronts
coming from many numerical experiments with a large variety of algorithms.
The corresponding hypervolume values are available online for
performance assessment %
\footnote{
The best known hypervolume values for all supported test instances are available via the
COCO platform at
\url{https://github.com/numbbo/coco/blob/master/code-experiments/src/suite\_biobj\_best\_values\_hyp.c}
}.
Moreover, the non-existence of analytical forms of Pareto set and Pareto front in our approach
can be even seen as an advantage: the combination of existing single-objective test functions
allows, in a controlled way, to mimic the typical constructions of real-world problems and
to empirically investigate the resulting Pareto set and Pareto front shapes from such
constructions.

The proposed multiobjective benchmark functions come in the form of pseudo-random \sphinxstyleemphasis{instances},
which allows to deal easily with the following two, otherwise non-trivial, tasks in performance
assessment:
\begin{itemize}
\item {} 
the comparison of algorithms with different success probabilities (in the sense of reaching
certain quality levels of the Pareto set approximations) and

\item {} 
the comparison of deterministic and stochastic approaches.

\end{itemize}

The paper is organized as follows. We start by outlining the fundamental definitions in
multiobjective optimization and benchmarking in Section~\ref{sec:preliminaries} before reviewing
existing multiobjective benchmark suites and their properties in Section~\ref{sec:stateoftheart}.
Section~\ref{sec:singleobj} then introduces the main concepts behind the well-known
single-objective \sphinxcode{bbob} test suite and discusses the ideas of function groups, objective
normalization and problem instances. Next, Section~\ref{sec:biobj} proposes the concrete \sphinxcode{bbob-biobj}
and \sphinxcode{bbob-biobj-ext} test suites and showcases some of the their functions by visualizing
the best found solutions in the objective and search space. Detailed descriptions and links to
visualizations for all proposed functions are provided in an accompanying extended version of this article, which can be found at \url{http://bbobbiobj.gforge.inria.fr/bbob-biobj-functions.pdf}. The
paper concludes with a proposal for creating test suites for an arbitrary number of
objectives in Section~\ref{sec:anyobj} and concluding remarks in Section~\ref{sec:conclusions}.

\section{Preliminaries}
\label{sec:preliminaries}
In the following, we consider bi-objective, unconstrained
\sphinxstylestrong{minimization} problems of the form
\begin{equation*}
\begin{split}\min_{x \in \mathbb{R}^n} F(x)=(f_\alpha(x),f_\beta(x)),\end{split}
\end{equation*}
where \(n\) is the number of variables of the problem (also called
the problem dimension), \(f_\alpha: \mathbb{R}^n \rightarrow \mathbb{R}\)
and \(f_\beta: \mathbb{R}^n \rightarrow \mathbb{R}\) are the two
objective functions, and the \(\min\) operator is related to the
standard \sphinxstyleemphasis{dominance relation}. A solution \(x\in\mathbb{R}^n\)
is thereby said to \sphinxstyleemphasis{dominate} another solution \(y\in\mathbb{R}^n\) if
\(f_\alpha(x) \leq f_\alpha(y)\) and \(f_\beta(x) \leq f_\beta(y)\) hold and at
least one of the inequalities is strict. Note that we adopt the notation
\(f_\alpha\) for the first objective (resp. \(f_\beta\) for the second
objective) instead of \(f_1\) and \(f_2\) to avoid confusion with
notations adopted within the single-objective \sphinxcode{bbob} test suite.

Solutions which are not dominated by any other solution in the search
space are called \sphinxstyleemphasis{Pareto-optimal} or \sphinxstyleemphasis{efficient solutions}. All
Pareto-optimal solutions constitute the \sphinxstyleemphasis{Pareto set} of which an
approximation is sought. The Pareto set's image in the
objective space \(F(\mathbb{R}^n)\) is called the \sphinxstyleemphasis{Pareto front}.

Two specific points in the objective space are important to mention. The \sphinxstyleemphasis{ideal point}
is defined as the vector in objective space that contains the optimal \(F\)-value for each
objective \sphinxstyleemphasis{independently}. More precisely, if
\(f_\alpha^{\rm opt}:= \inf_{x\in \mathbb{R}^n} f_\alpha(x)\) and
\(f_\beta^{\rm opt}:= \inf_{x\in \mathbb{R}^n} f_\beta(x)\), the ideal point is given by
\begin{quote}
    \begin{equation*}
    z_{\rm ideal}  =  (f_\alpha^{\rm opt},f_\beta^{\rm opt}).
\end{equation*}\end{quote}

The \sphinxstyleemphasis{nadir point} (in objective space) consists in each objective of
the worst value obtained by a Pareto-optimal solution. More precisely,
if we denote the set of Pareto optimal points by \(\mathcal{P}\), the
nadir point satisfies
\begin{quote}
    \begin{equation*}
    z_{\rm nadir}  =   \left( \sup_{x \in \mathcal{P}} f_\alpha(x),
 \sup_{x \in \mathcal{P}} f_\beta(x)  \right).
\end{equation*}\end{quote}

In the specific case where each of two objective functions has a unique global
minimum (that is, a single point in the search space which maps to the global minimum
function value),
 \begin{equation*}
 z_{\rm nadir}  =   \left( f_\alpha(x^{\rm opt}_{\beta}),
     f_\beta(x^{\rm opt}_{\alpha})  \right),
 \end{equation*}
where \(x^{\rm opt}_{\alpha}= \arg \min f_\alpha(x)\) and
\(x^{\rm opt}_{\beta}= \arg \min f_\beta(x)\).

Note that all given definitions generalize trivially to problems with more than two
objectives. When solving an unconstrained multiobjective problem as the above,
often the goal is to find,
with as few evaluations of \(F\) as possible, a set of non-dominated solutions which is
(i) as large as possible and (ii) has objective values as close to the Pareto front as possible. %
\footnote{
The distance in objective space is defined here in such a way that the nadir and ideal points
have in each coordinate the distance of one. Note
also that finding a set of non-dominated solutions
as large as possible might not always be the ultimate goal, in particular
if the number of objective functions is large.
}
Alternatively, the goal can also be to maximize a given quality indicator, for example
the hypervolume \citep{ztlf2003a} of the set of all non-dominated solutions found so far \citep{bro2016biperf}.

When an optimization algorithm approaches the above minimization problem, it actually
does not solve the generic function \(F\), but a concrete \sphinxstyleemphasis{instance} of \(F\)
with a concrete problem dimension and potentially other concrete inherent parameters.
Each generic multiobjective function \(F\) should therefore be seen as a
parametrized function
\(F^\theta: \mathbb{R}^n \to \mathbb{R}^m\) with parameter value \(\theta \in \Theta\),
a concrete problem dimension \(n\), and a concrete number of objectives \(m\), here
\(m=2\).
The parameter value \(\theta\) determines a so-called \sphinxstyleemphasis{function instance}.
For example, \(\theta\) might encode the location of the optimum of a
single-objective (parametrized) function \(f^{\theta}\), which means that
different instances have shifted optima:
\begin{equation*}
\begin{split}& f^\theta: \mathbb{R}^n \to \mathbb{R}\\
& \text{with } f^{x^{\rm opt}}(x) = || x - x^{\rm opt} ||^2\end{split}
\end{equation*}
where \(x\) and \(\theta=x^{\rm opt}\) live
both in \(\mathbb{R}^n(=\Theta)\).
Despite simple shifts in the search space as in the above example, other transformations
such as search space rotations and shifts in the objective values might
be defined by instances as well as it is done, for example, in the \href{https://github.com/numbbo/coco}{COCO} platform. In order to simplify the
handling of instances, we have a mapping
from a problem's parameter \(\theta\) to an integer such that we can talk about the
first, second, third, \(\ldots\) instance of a problem where the integer instance number is then
mapped to a concrete \(\theta\) parameter.
In the proposed multiobjective test suites, the multiobjective function instances are
furthermore determined by the instances of the underlying single-objective functions.

\section{Review of Existing Multiobjective Test Suites}
\label{sec:stateoftheart}
Many multiobjective test suites have been proposed throughout the years. Here, we in particular
discuss those that are scalable in the problem dimension and that are unconstrained or
box-constrained and defined in the continuous domain---the focus of our proposal for a new benchmark
suite.

The (evolutionary) multiobjective optimization field first performed numerical comparisons of
algorithms on single, independently proposed test problems such as the problems by
\citet{kurs1991a} and \citet{ff1995a}, see for example \citep{ttlk2002a}, or on actual
real-world studies, see for example \citep{vl1998a} for an early overview. A first attempt to create a
consistent multiobjective test function \sphinxstyleemphasis{suite} with several problems with desired properties was,
to the best of our knowledge, the work of \citet{vl1999a,vl1999b}. Van
Veldhuizen and Lamont clearly stated the need for scalable test suites and emphasized that problems
of a test suite should contain practically relevant features.

In the years that followed, several other scalable test suites have been proposed, of which the
most established ones are
\begin{itemize}
\item {} 
the ZDT suite of \citet{zdt2000a}, scalable in the number of variables but
with only two objectives,

\item {} 
its rotated version, the IHR problem suite of \citet{ihr2007a},

\item {} 
the DTLZ suite of \citet{dtlz2005a}, with seven problems, all
scalable in the number of variables and objectives,

\item {} 
the WFG suite of \citet{hhbw2006a} with nine scalable problems of various difficulties,

\item {} 
the LZ suite of \citet{lz2009a} containing problems with more complicated Pareto sets,

\item {} 
the CEC2007 suite, combining and extending 13 existing test functions from the literature \citep{hqdz2007a},

\item {} 
the CEC2009 suite with 13 problems overall \citep{zzzs2009a}, and finally

\item {} 
the CEC2017 suite with 17 collected problems, tailored towards many-objective optimization \citep{cec2017}.

\end{itemize}

Most of these test suites have some desirable properties like well-understood Pareto sets and Pareto
fronts with shapes of various kinds (linear, convex, concave, discontinuous). But they also possess
artificial characteristics that stem from the easier construction
of such problems---overrepresenting properties such as no or only few dependencies among
variables, Pareto sets located exactly at the boundary constraints, and the differentiation between
position and distance variables. Although, for example, the importance of non-separable test
functions in single-objective test suites is unquestioned and even Deb (\citet{deb2001a}, page 353f.)
states its significance, most proposed multiobjective test problems are still separable or mostly
separable in the sense that a function is separable if it can be optimized variable by variable.
Even though all test suites in the above list are scalable in the
problem dimension, we rarely see performance studies that investigate the scaling of the
algorithms with the problem dimension.

The arguably most complete paper on the topic of multiobjective benchmark problems to date is still
the work of \citep{hhbw2006a} where the authors (i) identify important properties
test functions should have, (ii) discuss in detail all other available test suites at that time
with respect to these properties, and (iii) finally propose a new, well-motivated test suite that
avoids many pitfalls of other test suites. In particular, Huband et al. (see \cite{hhbw2006a}, page 485)
recommend that multiobjective test suites should, in addition to recommendations for
single-objective test suites:
\begin{enumerate}
\item {} 
contain a few unimodal test problems to test convergence velocity relative to different
Pareto optimal geometries and bias conditions,

\item {} 
cover the three core types of Pareto optimal geometries: degenerate Pareto optimal fronts,
disconnected Pareto optimal fronts, and disconnected Pareto optimal sets,

\item {} 
have a majority of its test problems multimodal with a few deceptive problems,

\item {} 
have the majority of problems nonseparable, and

\item {} 
contain problems that are both nonseparable and multimodal to be representative of
real-world problems.

\end{enumerate}

All five recommendations are fulfilled for the test suites proposed in this paper.
Similar to the single-objective \sphinxcode{bbob} test functions, the WFG suite of \citep{hhbw2006a}
employs problem transformations that change the properties like \mbox{(non-)}separability, bias,
and the shape of the Pareto front of underlying \sphinxstyleemphasis{raw} objective functions.

One common property of the above mentioned test suites is that
their Pareto sets can be described in analytical form. This certainly has an advantage for
performance assessment but it also restricts the types of real-world problem characteristics that
can be captured with such functions.

However, in practice, multiobjective optimization problems are typically
constructed by combining objective functions that are defined (and understood) independently such
as cost and performance of a new product. The objective functions might thereby come from different
domains and share or do not share common properties such as uni-/multimodality, (non-)separability,
asymmetry, etc.

The idea of defining multiobjective test problems by combining single-objective test functions is
therefore straightforward and has been proposed before, for example by \citet{ihr2007a} and \citet{hwbw2015a}.
To create a benchmark suite with challenging properties observed in practice, we follow here the
same path and combine some of the existing, well-established, and well-understood test functions
of the \sphinxcode{bbob} test suite to create new multiobjective test suites.

\section{The Single-objective \sphinxstyleliteralintitle{bbob} Functions}
\label{sec:singleobj}
The main idea behind the multiobjective test suites proposed in this paper is to take
well-understood single-objective test functions with problem properties observed in practice
and to combine them to form multiobjective problems. We choose the functions from the
well established \sphinxcode{bbob} test suite since they all
\begin{itemize}
\item {} 
have associated scientific questions that can be answered by running numerical experiments on them,

\item {} 
are categorized into function groups depending on their properties, and finally

\item {} 
already come in the form of function instances.

\end{itemize}

This section first discusses properties of real-world problems and how the \sphinxcode{bbob} test suite
balances different problem difficulties. It then gives more details about the \sphinxcode{bbob} functions,
their function groups and instances.

\subsection{Real-World Function Properties}
\label{sec:real-world-function-properties}
We present here in short the
general properties of objective functions that
are related to difficulties observed in real-world problems.
It depends on these properties whether an optimization problem
is easy or hard to solve. They build the basis of the function groups described later.

A \sphinxstyleemphasis{separable} function does not have any dependencies among its
variables and can therefore be optimized by applying \(n\) independent
one-dimensional optimizations along each coordinate axis while
keeping the other variables fixed. Difficult optimization problems are
typically not separable and thus, \sphinxstyleemphasis{non-separable} optimization problems
should be considered. The
typical well-established technique to generate non-separable benchmark
functions from separable ones is the application of a rotation matrix.
That is, if \(g(x)\) is a separable function with respect to \(x\) and
\(\mathbf R \in \mathbb{R}^{n \times n}\) is a rotation matrix, then \(g(\mathbf{R} x)\)
will generally be non-separable with respect to \(x\).

A \sphinxstyleemphasis{unimodal} function has only one local minimum which is at the same
time also its global one.
A \sphinxstyleemphasis{multimodal} function has more local minima which is highly common in
practical optimization problems. We consider a multimodal function to have
\sphinxstyleemphasis{weak global structure} if the qualities (the \(f\)-values) of the local
optima are only weakly related with their locations in search space,
e.g. when neighboring optima do not generally have similar quality values.

\sphinxstyleemphasis{Ill-conditioning} is another typical challenge of real-parameter
optimization and, besides multimodality, probably the most common one.
The condition number measures, loosely speaking, how strongly the steepness
of the gradient depends on the position within a level set.
The condition number measures in essence a variation of sensitivity,
with a minimal value of 1.
A small condition number means that the function is well-conditioned,
while a large condition number indicates an ill-conditioned function.
Conditioning can be rigorously formalized in the
case of convex quadratic functions (with optimum in zero WLOG),
\(f(x) = \frac{1}{2} x^THx\) where \(H\) is a symmetric
positive definite matrix, as the condition number of the Hessian matrix
\(H\). Since contour lines associated to a convex quadratic function
are ellipsoids, the condition number corresponds to the squared ratio
between the largest and the shortest axis length of the ellipsoid.

The \sphinxcode{bbob} test suite contains ill-conditioned functions
with a typical conditioning of \(10^6\). We believe this is a realistic
requirement, while we have seen practical problems with conditioning
as large as \(10^{10}\) \citep{cdhq2010a}.

\subsection{Balancing Problem Difficulties}
\label{sec:balancing-problem-difficulties}
It is worth noting that in several existing single-objective test suites, some of the easier
properties are overrepresented. For example, in the CUTEr/CUTEst test suite
\citep{got2005a}, 202 (54\%) out of the 375 functions, that are labeled as unconstrained
or bound constrained, are of the ``sum of squares'' type, a further 58 (15\%) are
quadratic. Furthermore, out of the 191 problems with a fixed dimension,
there are 49 (26\%) that have only two variables while only 31 (16\%)
have a dimension larger than 10.

Such an overrepresentation is not a big problem per se, but when making
statements on algorithm performance aggregated over all functions in a suite,
one has to keep in mind that the performance of the better algorithms
might simply come from the fact that they are tailored towards simpler
problems.

With the \sphinxcode{bbob} test suite, all problems are scalable in dimension and
belong to a certain problem group, sharing similar difficulties. It is
therefore possible to aggregate performance data only over a subset of the
functions sharing the same properties. Having all problem groups of
similar size also avoids problems of overfitting to certain difficulties
if aggregated results are presented.

\subsection{Function Groups}
\label{sec:function-groups}
Related to the mentioned problem difficulties above, the \sphinxcode{bbob} test suite
comes with 24 functions, split into five function groups:
\begin{itemize}
\item {} 
Group 1 \sphinxstylestrong{Separable} contains only separable functions:
\begin{itemize}
\item {} 
Sphere function (\href{http://coco.lri.fr/downloads/download15.03/bbobdocfunctions.pdf\#page=5}{\(f_1\) in the \sphinxcode{bbob} suite})

\item {} 
Separable ellipsoid function (\href{http://coco.lri.fr/downloads/download15.03/bbobdocfunctions.pdf\#page=10}{\(f_2\) in the \sphinxcode{bbob} suite})

\item {} 
Separable Rastrigin function (\href{http://coco.lri.fr/downloads/download15.03/bbobdocfunctions.pdf\#page=15}{\(f_3\) in the \sphinxcode{bbob} suite})

\item {} 
B\"{u}che-Rastrigin function (\href{http://coco.lri.fr/downloads/download15.03/bbobdocfunctions.pdf\#page=20}{\(f_4\) in the \sphinxcode{bbob} suite})

\item {} 
Linear slope function (\href{http://coco.lri.fr/downloads/download15.03/bbobdocfunctions.pdf\#page=25}{\(f_5\) in the \sphinxcode{bbob} suite})

\end{itemize}

\item {} 
Group 2 \sphinxstylestrong{Moderate} consists of functions with low or moderate conditioning, including multi-modal functions:
\begin{itemize}
\item {} 
Attractive sector function (\href{http://coco.lri.fr/downloads/download15.03/bbobdocfunctions.pdf\#page=30}{\(f_6\) in the \sphinxcode{bbob} suite})

\item {} 
Step ellipsoid function (\href{http://coco.lri.fr/downloads/download15.03/bbobdocfunctions.pdf\#page=35}{\(f_7\) in the \sphinxcode{bbob} suite})

\item {} 
Original Rosenbrock function (\href{http://coco.lri.fr/downloads/download15.03/bbobdocfunctions.pdf\#page=40}{\(f_8\) in the \sphinxcode{bbob} suite})

\item {} 
Rotated Rosenbrock function (\href{http://coco.lri.fr/downloads/download15.03/bbobdocfunctions.pdf\#page=45}{\(f_9\) in the \sphinxcode{bbob} suite})

\end{itemize}

\item {} 
Group 3 \sphinxstylestrong{Ill-conditioned} contains unimodal functions with high conditioning:
\begin{itemize}
\item {} 
Ellipsoid function (\href{http://coco.lri.fr/downloads/download15.03/bbobdocfunctions.pdf\#page=50}{\(f_{10}\) in the \sphinxcode{bbob} suite})

\item {} 
Discus function (\href{http://coco.lri.fr/downloads/download15.03/bbobdocfunctions.pdf\#page=55}{\(f_{11}\) in the \sphinxcode{bbob} suite})

\item {} 
Bent cigar function (\href{http://coco.lri.fr/downloads/download15.03/bbobdocfunctions.pdf\#page=60}{\(f_{12}\) in the \sphinxcode{bbob} suite})

\item {} 
Sharp ridge function (\href{http://coco.lri.fr/downloads/download15.03/bbobdocfunctions.pdf\#page=65}{\(f_{13}\) in the \sphinxcode{bbob} suite})

\item {} 
Sum of different powers function (\href{http://coco.lri.fr/downloads/download15.03/bbobdocfunctions.pdf\#page=70}{\(f_{14}\) in the \sphinxcode{bbob} suite})

\end{itemize}

\item {} 
Group 4 \sphinxstylestrong{Multi-modal} comprises multi-modal functions with adequate global structure:
\begin{itemize}
\item {} 
Rastrigin function (\href{http://coco.lri.fr/downloads/download15.03/bbobdocfunctions.pdf\#page=75}{\(f_{15}\) in the \sphinxcode{bbob} suite})

\item {} 
Weierstrass function (\href{http://coco.lri.fr/downloads/download15.03/bbobdocfunctions.pdf\#page=80}{\(f_{16}\) in the \sphinxcode{bbob} suite})

\item {} 
Schaffer F7 function with condition number 10  (\href{http://coco.lri.fr/downloads/download15.03/bbobdocfunctions.pdf\#page=85}{\(f_{17}\) in the \sphinxcode{bbob} suite})

\item {} 
Schaffer F7 function with condition number 1000 (\href{http://coco.lri.fr/downloads/download15.03/bbobdocfunctions.pdf\#page=90}{\(f_{18}\) in the \sphinxcode{bbob} suite})

\item {} 
Composite Griewank-Rosenbrock function F8F2 (\href{http://coco.lri.fr/downloads/download15.03/bbobdocfunctions.pdf\#page=95}{\(f_{19}\) in the \sphinxcode{bbob} suite})

\end{itemize}

\item {} 
Group 5 \sphinxstylestrong{Weakly-structured} consists of multi-modal functions with weak global structure:
\begin{itemize}
\item {} 
Schwefel \(x \sin{x}\) function (\href{http://coco.lri.fr/downloads/download15.03/bbobdocfunctions.pdf\#page=100}{\(f_{20}\) in the \sphinxcode{bbob} suite})

\item {} 
Gallagher 101 peaks function (\href{http://coco.lri.fr/downloads/download15.03/bbobdocfunctions.pdf\#page=105}{\(f_{21}\) in the \sphinxcode{bbob} suite})

\item {} 
Gallagher 21 peaks function  (\href{http://coco.lri.fr/downloads/download15.03/bbobdocfunctions.pdf\#page=110}{\(f_{22}\) in the \sphinxcode{bbob} suite})

\item {} 
Katsuura function (\href{http://coco.lri.fr/downloads/download15.03/bbobdocfunctions.pdf\#page=115}{\(f_{23}\) in the \sphinxcode{bbob} suite})

\item {} 
Lunacek bi-Rastrigin function (\href{http://coco.lri.fr/downloads/download15.03/bbobdocfunctions.pdf\#page=120}{\(f_{24}\) in the \sphinxcode{bbob} suite})

\end{itemize}

\end{itemize}

See \citep{hafr2009a} for the exact problem formulations.
Problem difficulty is typically increasing from the first to the last group,
but there are exceptions, for example,
solving the B\"{u}che-Rastrigin function from the first group is quite difficult for most algorithms.

The main idea behind these hand-assigned function groups is that algorithm performance can be
easily aggregated over all functions within a group in order to make
meaningful statements on subsets of all 24 functions. If, for example,
an application engineer knows that her/his real-world problem
is multi-modal and also shows some global structure, a recommendation
about which algorithm will perform well on that problem can be made
mostly by looking at the multi-modal function group.

\subsection{Function Instances}
\label{sec:bbobinstances}
All \sphinxcode{bbob} functions come naturally in the form of instances.
That is to say, each function optimized by an algorithm takes the form:
\begin{equation*}
\begin{split}f(x) =  H_1 \circ \ldots \circ H_{k_1} (f_{\rm raw}(T_1 \circ \ldots \circ T_{k_2}(x)))\end{split}
\end{equation*}
where \(f_{\rm raw}\) is a raw function---usually the simplest representative of the function class (like the sphere function with optimum in zero)---and
where \(T_i: \mathbb{R}^n \to \mathbb{R}^n\) are search space transformations and \(H_i: \mathbb{R} \to \mathbb{R}\) are function value transformations that are applied to the raw function. For example search space transformations can be rotations or translations of the optimum and for example, a function-value transformation can be translating the function by a scalar. Each of those transformations applied to the raw function are actually (pseudo)-random, e.g. when applying a translation in the search space, the vector by which the search point is shifted is randomly sampled. They can be seen as instances of a parametrized transformation.

In an abstract manner, the functions optimized are instances of a parametrized function \(F^\theta\) (as introduced in Section~\ref{sec:preliminaries});
the parameter \(\theta\) is instantiated (pseudo)-randomly from an integer number, the so-called instance number (and potentially function number).
We refer to a function class as a set of functions \(\{ F^\theta : \theta \in \Theta \}\) and we often name the function class after its raw function.

Transformations that are shared
by all \sphinxcode{bbob} functions are shifts
in the optimal function value and a pseudo-random location of the optimum. In addition, several of
the non-separable functions are created by pseudo-random rotations of the search space and many of
the simpler functions are made less regular by non-linear transformations in both search and
objective space. See \citep{hafr2009a} for more details.

Though the potential set of instances for a given \sphinxcode{bbob} function is unbounded (and can be indexed
by any positive integer), numerical benchmarking experiments are typically advised on 10--15 of
those instances. Default instances in the \href{https://github.com/numbbo/coco}{COCO} implementation might change from year to year to
avoid overfitting. Note also that in some cases, single instances might be more
difficult/easier to solve than others. However, in general, the difficulties
among instances of the same \sphinxcode{bbob} function are more similar than between different functions.

Performing numerical benchmarking experiments on a set of different instances of a parametrized
function instead of experiments on a single fixed function has an immediate advantage:
deterministic algorithms and stochastic algorithms can be compared easily in the same way stochastic
algorithms are naturally compared. Running a deterministic algorithm on different instances of the
same parametrized function introduces stochasticity of the runtime to reach certain target
difficulties among runs in the same way than the combined stochasticity from the instance generation
and the random events within a stochastic algorithm. Care, however, has to be taken that the
variation of problem difficulty among instances is relatively low compared to the variation of
difficulty between the actual benchmark functions %
\footnote{
An assumption that does not always hold for all instances of highly multi-modal functions
and that is not the case at all for instances of most combinatorial optimization problems.
}.

\subsection{Domain Bounds}
\label{sec:domain-bounds}
All functions provided in the \sphinxcode{bbob} suite are unbounded,
i.e., defined on the entire real-valued space \(\mathbb{R}^n\).
The search domain of interest, however, is defined as \([-5,5]^n\).
With the exception of the linear slope (function \href{http://coco.lri.fr/downloads/download15.03/bbobdocfunctions.pdf\#page=25}{\(f_5\) in the \sphinxcode{bbob} suite}),
the optimal solutions and hyperballs of radius 1 around them lie within the domain of interest for all instances in all dimensions.

\subsection{Normalization and Target Difficulties}
\label{sec:normalization-and-target-difficulties}
All \sphinxcode{bbob} functions are normalized in the sense that the given target function
values/difficulties around the optimal function value are comparable
over functions and instances.
Functions are provided with an \(f\)-offset such that the optimal function
value is, loosely speaking, a realization of a Cauchy distribution with median
zero and interquartile range \(200\).
The optimal function value is furthermore rounded to two decimal places and
set to \(\pm 1000\) if its absolute value exceeds \(1000\) \citep{hafr2009a}.
The target difficulties are computed as a set of differences to the optimal function value.
The differences are equally spaced on the log scale and the same for all functions and instances.
Algorithms however are not allowed to use or exploit any of this information \citep{han2016ex}.

\section{The Proposed Bi-Objective Test Suites}
\label{sec:biobj}
The main contribution of this paper is the definition of
multiobjective test suites by combining single-objective functions. For the bi-objective case and given
the 24 single-objective \sphinxcode{bbob} functions from \citet{hafr2009a}, it is natural to combine all
of them in pairs---resulting in \(24^2=576\) bi-objective functions overall.
We however assume that
multiobjective optimization algorithms are not sensitive to permutations of
the objective functions, so that there is no need to include the bi-objective function
\((f_\beta,f_\alpha)\) if \((f_\alpha,f_\beta)\) is already present in the suite.
This results in the total of \(\binom{25}{2} = 300\) function combinations---the number
of \(2\)-combinations with repetitions, or \(2\)-multicombinations of \(24\)
objective functions %
\footnote{
The general formula to compute the number \(N_{k,s}\) of
\(k\)-combinations with repetitions drawn from a ground set
with \(s\) entries is \(N_{k,s} = \binom{s+k-1}{k}\).
}.

While a benchmarking suite should contain a large number of different problems to avoid
overfitting of algorithms to the problem suite, first tests in \citet{bro2015gecco} showed that having 300 functions is
impracticable in terms of the overall running time of a benchmarking
experiment. Therefore, a subset of these 300 functions needed to be selected.

This section presents two such selections---the \sphinxcode{bbob-biobj} test suite with 55 functions and its
extension, the \sphinxcode{bbob-biobj-ext} test suite with 92 functions. We also provide visualizations
for some of the functions from the two suites here, showing different Pareto set and front shapes,
while the plots for all 92 functions are collected in the accompanying paper at \url{http://bbobbiobj.gforge.inria.fr/bbob-biobj-functions.pdf}.

\subsection{The \sphinxstyleliteralintitle{bbob-biobj} Test Suite}
\label{sec:bbobbiobjsuite}
The bi-objective \sphinxcode{bbob-biobj} test suite is created by exploiting the organization of the \sphinxcode{bbob}
functions into groups. More precisely, only two (representative) functions from each of the \sphinxcode{bbob}
function groups are chosen.
This way, we do not introduce any bias towards a specific group. In addition,
within each group, the functions are chosen to be the most
representative without repeating similar functions. For example,
only one Ellipsoid, one Rastrigin, and one Gallagher function are
included in the \sphinxcode{bbob-biobj} suite although they appear in
multiple versions in the \sphinxcode{bbob} suite.

Our choice of
10 \sphinxcode{bbob} functions for creating the \sphinxcode{bbob-biobj} test suite is the following:
\begin{itemize}
\item {} 
Separable functions:
\begin{itemize}
\item {} 
Sphere function (\href{http://coco.lri.fr/downloads/download15.03/bbobdocfunctions.pdf\#page=5}{\(f_1\) in the \sphinxcode{bbob} suite})

\item {} 
Separable ellipsoid function (\href{http://coco.lri.fr/downloads/download15.03/bbobdocfunctions.pdf\#page=10}{\(f_2\) in the \sphinxcode{bbob} suite})

\end{itemize}

\item {} 
Functions with low or moderate conditioning
\begin{itemize}
\item {} 
Attractive sector function (\href{http://coco.lri.fr/downloads/download15.03/bbobdocfunctions.pdf\#page=30}{\(f_6\) in the \sphinxcode{bbob} suite})

\item {} 
Original Rosenbrock function (\href{http://coco.lri.fr/downloads/download15.03/bbobdocfunctions.pdf\#page=40}{\(f_8\) in the \sphinxcode{bbob} suite})

\end{itemize}

\item {} 
Unimodal functions with high conditioning
\begin{itemize}
\item {} 
Sharp ridge function (\href{http://coco.lri.fr/downloads/download15.03/bbobdocfunctions.pdf\#page=65}{\(f_{13}\) in the \sphinxcode{bbob} suite})

\item {} 
Sum of different powers function (\href{http://coco.lri.fr/downloads/download15.03/bbobdocfunctions.pdf\#page=70}{\(f_{14}\) in the \sphinxcode{bbob} suite})

\end{itemize}

\item {} 
Multi-modal functions with adequate global structure
\begin{itemize}
\item {} 
Rastrigin function (\href{http://coco.lri.fr/downloads/download15.03/bbobdocfunctions.pdf\#page=75}{\(f_{15}\) in the \sphinxcode{bbob} suite})

\item {} 
Schaffer F7 function with condition number 10  (\href{http://coco.lri.fr/downloads/download15.03/bbobdocfunctions.pdf\#page=85}{\(f_{17}\) in the \sphinxcode{bbob} suite})

\end{itemize}

\item {} 
Multi-modal functions with weak global structure
\begin{itemize}
\item {} 
Schwefel \(x \sin{x}\) function (\href{http://coco.lri.fr/downloads/download15.03/bbobdocfunctions.pdf\#page=100}{\(f_{20}\) in the \sphinxcode{bbob} suite})

\item {} 
Gallagher 101 peaks function (\href{http://coco.lri.fr/downloads/download15.03/bbobdocfunctions.pdf\#page=105}{\(f_{21}\) in the \sphinxcode{bbob} suite})

\end{itemize}

\end{itemize}

Using the previously described pairwise combinations, this results in
having only \(\binom{11}{2} = 55\) bi-objective functions in
the final \sphinxcode{bbob-biobj} suite (denoted as {\hyperref[index:f1]{\(F_{1}\)}} to {\hyperref[index:f55]{\(F_{55}\)}}
in the rest of the paper). They are all scalable in the search space dimension and come in
the form of instances as it is the case with the original \sphinxcode{bbob} suite.

\begin{figure}[htbp]
	\centering
  \noindent\includegraphics[scale=0.8]{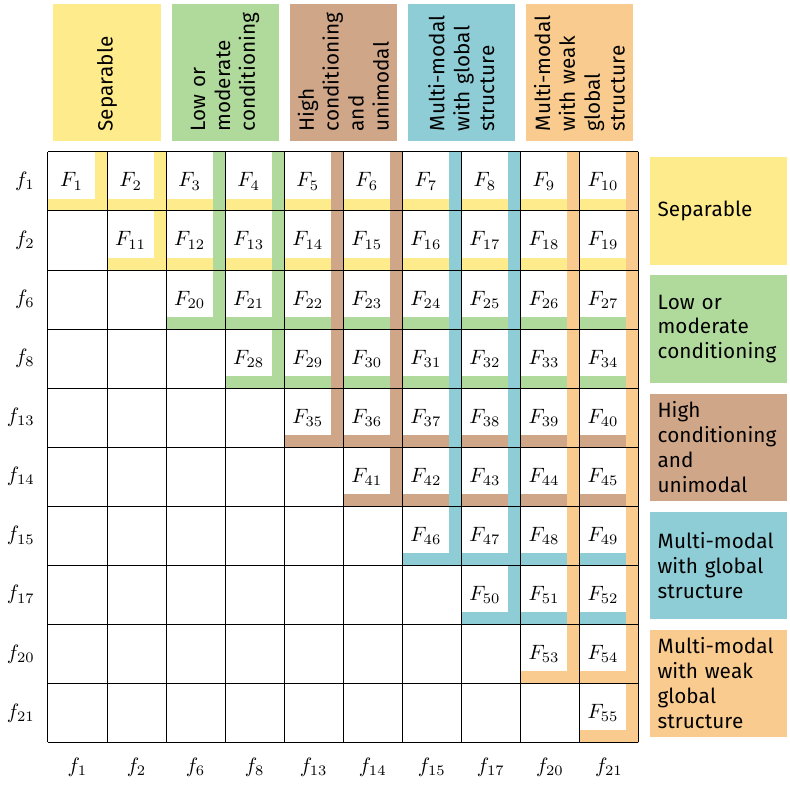}
	\caption{The functions of the \sphinxcode{bbob-biobj} test suite ({\hyperref[index:f1]{\(F_{1}\)}} to
	         {\hyperref[index:f55]{\(F_{55}\)}}, in the table cells) together
					 with the information about which single-objective \sphinxcode{bbob} 
					 functions are used to define them (left and bottom annotations).}
	\label{fig:bbobbiobjfunctionsoverview}
\end{figure}

In the following, we specify the common properties of the \sphinxcode{bbob-biobj} functions
and the main rationale behind them
while concrete details on each of
the 55 functions are given in the accompanying paper at \url{http://bbobbiobj.gforge.inria.fr/bbob-biobj-functions.pdf}.
See Fig.~\ref{fig:bbobbiobjfunctionsoverview} for an overview of how the single-objective functions (denoted with \(f_i\))
are combined to form the bi-objective functions of the \sphinxcode{bbob-biobj} test suite.

\subsubsection{Domain and Region of Interest}
\label{sec:domain-and-region-of-interest}
Since we use the single-objective \sphinxcode{bbob} functions to construct the \sphinxcode{bbob-biobj} suite,
all functions are unbounded and the extreme solutions of the Pareto set
are guaranteed to lie within \([-5,5]^n\).

Note that the Pareto set can partially lie outside of this area but that the major part of
the Pareto set is expected to lie within it. Moreover, it is highly unlikely that any
non-dominated solutions would be found outside of the \sphinxstyleemphasis{region of interest} \([-100,100]^n\).
In other words, we believe that the region of interest contains the entire Pareto set, but
due to the nature of the \sphinxcode{bbob-biobj} function definitions, there is no
guarantee that this is indeed always the case.

\subsubsection{Function Groups}
\label{sec:functiongroups}
By combining the original \sphinxcode{bbob} function groups, we obtain 15 function
groups to structure the 55 bi-objective functions of the \sphinxcode{bbob-biobj} test
suite. Each function group contains three or four functions. We are listing
below the function groups and in parenthesis the functions that belong to
the respective groups (see also Fig.~\ref{fig:bbobbiobjfunctionsoverview}):

\begin{enumerate}
\item {} 
separable - separable (functions {\hyperref[index:f1]{\(F_{1}\)}}, {\hyperref[index:f2]{\(F_{2}\)}}, {\hyperref[index:f11]{\(F_{11}\)}})

\item {} 
separable - moderate ({\hyperref[index:f3]{\(F_{3}\)}}, {\hyperref[index:f4]{\(F_{4}\)}}, {\hyperref[index:f12]{\(F_{12}\)}}, {\hyperref[index:f13]{\(F_{13}\)}})

\item {} 
separable - ill-conditioned ({\hyperref[index:f5]{\(F_{5}\)}}, {\hyperref[index:f6]{\(F_{6}\)}}, {\hyperref[index:f14]{\(F_{14}\)}}, {\hyperref[index:f15]{\(F_{15}\)}})

\item {} 
separable - multi-modal ({\hyperref[index:f7]{\(F_{7}\)}}, {\hyperref[index:f8]{\(F_{8}\)}}, {\hyperref[index:f16]{\(F_{16}\)}}, {\hyperref[index:f17]{\(F_{17}\)}})

\item {} 
separable - weakly-structured ({\hyperref[index:f9]{\(F_{9}\)}}, {\hyperref[index:f10]{\(F_{10}\)}}, {\hyperref[index:f18]{\(F_{18}\)}}, {\hyperref[index:f19]{\(F_{19}\)}})

\item {} 
moderate - moderate ({\hyperref[index:f20]{\(F_{20}\)}}, {\hyperref[index:f21]{\(F_{21}\)}}, {\hyperref[index:f28]{\(F_{28}\)}})

\item {} 
moderate - ill-conditioned ({\hyperref[index:f22]{\(F_{22}\)}}, {\hyperref[index:f23]{\(F_{23}\)}}, {\hyperref[index:f29]{\(F_{29}\)}}, {\hyperref[index:f30]{\(F_{30}\)}})

\item {} 
moderate - multi-modal ({\hyperref[index:f24]{\(F_{24}\)}}, {\hyperref[index:f25]{\(F_{25}\)}}, {\hyperref[index:f31]{\(F_{31}\)}}, {\hyperref[index:f32]{\(F_{32}\)}})

\item {} 
moderate - weakly-structured ({\hyperref[index:f26]{\(F_{26}\)}}, {\hyperref[index:f27]{\(F_{27}\)}}, {\hyperref[index:f33]{\(F_{33}\)}}, {\hyperref[index:f34]{\(F_{34}\)}})

\item {} 
ill-conditioned - ill-conditioned ({\hyperref[index:f35]{\(F_{35}\)}}, {\hyperref[index:f36]{\(F_{36}\)}}, {\hyperref[index:f41]{\(F_{41}\)}})

\item {} 
ill-conditioned - multi-modal ({\hyperref[index:f37]{\(F_{37}\)}}, {\hyperref[index:f38]{\(F_{38}\)}}, {\hyperref[index:f42]{\(F_{42}\)}}, {\hyperref[index:f43]{\(F_{43}\)}})

\item {} 
ill-conditioned - weakly-structured ({\hyperref[index:f39]{\(F_{39}\)}}, {\hyperref[index:f40]{\(F_{40}\)}}, {\hyperref[index:f44]{\(F_{44}\)}}, {\hyperref[index:f45]{\(F_{45}\)}})

\item {} 
multi-modal - multi-modal ({\hyperref[index:f46]{\(F_{46}\)}}, {\hyperref[index:f47]{\(F_{47}\)}}, {\hyperref[index:f50]{\(F_{50}\)}})

\item {} 
multi-modal - weakly structured ({\hyperref[index:f48]{\(F_{48}\)}}, {\hyperref[index:f49]{\(F_{49}\)}}, {\hyperref[index:f51]{\(F_{51}\)}}, {\hyperref[index:f52]{\(F_{52}\)}})

\item {} 
weakly structured - weakly structured ({\hyperref[index:f53]{\(F_{53}\)}}, {\hyperref[index:f54]{\(F_{54}\)}}, {\hyperref[index:f55]{\(F_{55}\)}})

\end{enumerate}

\subsubsection{Normalization of Objectives}
\label{sec:normalization-of-objectives}
None of the 55 \sphinxcode{bbob-biobj} functions is explicitly normalized and the
optimization algorithms therefore have to cope with objective values in
different ranges. Typically, different orders of magnitude
between the objective values can be observed.

However, to facilitate comparison of algorithm performance over different functions,
we suggest to normalize the objectives based on the ideal and nadir points
before calculating the hypervolume indicator \citep{bro2016biperf}.
Both points can be computed, because the global
optimum is known and is unique for the used 10 \sphinxcode{bbob} base functions.
In the black-box optimization benchmarking setup,
the algorithm is allowed to use the values of the nadir point as
an upper bound on the region of interest in objective space.

\subsubsection{Instances}
\label{sec:bbobbiobjinstances}
Our proposed test functions are parametrized and their instances are instantiations of the
underlying parameters as is done for the \sphinxcode{bbob} functions (see \citet{han2016coco}).
The instances for the bi-objective
functions are obtained using instances of each single-objective function composing the
bi-objective one. In addition, we assert two conditions:
\begin{quote}

1. The Euclidean distance between the two single-objective optimal solutions (also called the
extreme solutions) in the search space is at least \(10^{-4}\).

2. The Euclidean distance between the ideal and the nadir point in the non-normalized
objective space is at least \(10^{-1}\).
\end{quote}

We associate to each function instance an integer instance ID. The relation between the
instance ID, \(K^{F}_{\rm ID}\), of a bi-objective function \(F = (f_\alpha, f_\beta)\)
and the instance IDs, \(K_{\rm ID}^{f_\alpha}\) and \(K_{\rm ID}^{f_\beta}\), of its
underlying single-objective functions \(f_\alpha\) and \(f_\beta\) is the following:
\begin{itemize}
\item {} 
\(K_{\rm ID}^{f_\alpha} = 2 K^{F}_{\rm ID} + 1\) and

\item {} 
\(K_{\rm ID}^{f_\beta} = K_{\rm ID}^{f_\alpha} + 1\)

\end{itemize}

If we find that the two above conditions are not satisfied for all dimensions and
functions in the \sphinxcode{bbob-biobj} suite, we increase the instance ID of the
second objective successively until both properties are fulfilled.
For example, the \sphinxcode{bbob-biobj} instance ID
8 corresponds to the instance ID 17 for the first objective and instance ID 18 for
the second objective. For the \sphinxcode{bbob-biobj} instance ID 9, on the contrary, the
first instance ID is 19 but for the second objective, instance ID 21 is chosen
instead of instance ID 20 in order to conform with both conditions.

Exceptions to the above rule are, for historical reasons, the
\sphinxcode{bbob-biobj} instance IDs 1 and 2 in order to match the instance IDs
1 to 5 with the ones proposed in \citet{bro2015gecco}. The \sphinxcode{bbob-biobj}
instance ID 1 contains the single-objective instance IDs 2 and 4 and
the \sphinxcode{bbob-biobj} instance ID 2 contains the two instance IDs 3 and 5.

For each bi-objective function and given dimension, the \sphinxcode{bbob-biobj} suite
contains by default 15 instances. %
\footnote{
In principle, as for the instance generation for the \sphinxcode{bbob} suite,
the number of possible instances for the \sphinxcode{bbob-biobj} suite is unlimited
\citep{han2016coco}.
However, running some tests with too few instances will render the
potential statistics and their interpretation problematic while even the
tiniest observed difference can be made statistically significant with a
high enough number of instances. A good compromise to avoid either pitfall
seems to lie between, say, 9 and 19 instances.
}

Problem instances in the multiobjective case can differ more
wildly than those in the single-objective case. Even when
for each single objective the different instances are similar,
different combinations of them can result in different shapes of the Pareto front
(for example continuous vs.\ discontinuous) or in different difficulties to
solve such problems (the orientation of level sets, for example, might be
in accordance between the objectives or perpendicular---resulting in
significantly different multiobjective problems when two highly-conditioned
functions are combined). %
\footnote{
While we cannot give a guarantee about the maximal difference in difficulty
between instances, numerical experiments show that performance
differences up to two orders of magnitude (in the number of function
evaluations to reach a certain hypervolume indicator precision) can be
observed in some cases. Differences of more than one order of magnitude
happen in maximally 30\% of the function/dimension pairs with typical
algorithms on the proposed \sphinxcode{bbob-biobj} test suite. Due to the higher
amount of multimodal functions in the \sphinxcode{bbob-biobj-ext} suite,
differences among instances are more common.
}
Consequently, we do not adopt the technique from the single-objective case
to compare results from different instance sets.
Yet it may well be possible to cherry-pick instances carefully to generate
multiple sets with sufficiently uniform characteristics.

\subsection{The \sphinxstyleliteralintitle{bbob-biobj-ext} Test Suite}
\label{sec:bbobbiobjextsuite}
Having all combinations of only a subset of the single-objective \sphinxcode{bbob} functions in a test suite
like the above \sphinxcode{bbob-biobj} one has its
advantages but also a few disadvantages. Using only a subset of the 24 \sphinxcode{bbob} functions
introduces a bias towards the chosen functions and reduces the amount of different difficulties
a bi-objective algorithm is exposed to in the benchmarking exercise. Allowing all combinations of
(a subset of the) \sphinxcode{bbob} functions also increases the percentage of problems for which both
objectives are from different \sphinxcode{bbob} function groups. In practice, however, it can often be
assumed that both objective functions come from a similar ``function domain''.

The rationale behind the extended test suite, denoted as \sphinxcode{bbob-biobj-ext},
is therefore to reduce the mentioned effects. To this end, we add all within-group combinations of \sphinxcode{bbob}
functions which are not already in the \sphinxcode{bbob-biobj} suite and which do not combine a function
with itself. For technical reasons, we also remove the Weierstrass function (\href{http://coco.lri.fr/downloads/download15.03/bbobdocfunctions.pdf\#page=80}{\(f_{16}\)} in the
\sphinxcode{bbob} suite) because its optimum is not necessarily unique and computing the nadir point is
therefore technically more challenging than for the other functions.
This extension adds \(3 \cdot (4+3+2+1-1) + 2\cdot (3+2+1-1) = 3\cdot 9+2\cdot 5=37\) functions, resulting in
92 functions overall.

\hyperref[\detokenize{index:fig92}]{Fig.\@ \ref{\detokenize{index:fig92}}} details which single-objective \sphinxcode{bbob} functions (left and bottom annotations) are
contained in the 92 \sphinxcode{bbob-biobj-ext} functions. Note that the numbers of the first 55
\sphinxcode{bbob-biobj-ext} functions are the same as in the
\sphinxcode{bbob-biobj} test suite for compatibility reasons.
\begin{figure}[htbp]
\centering

\noindent\includegraphics[scale=0.65]{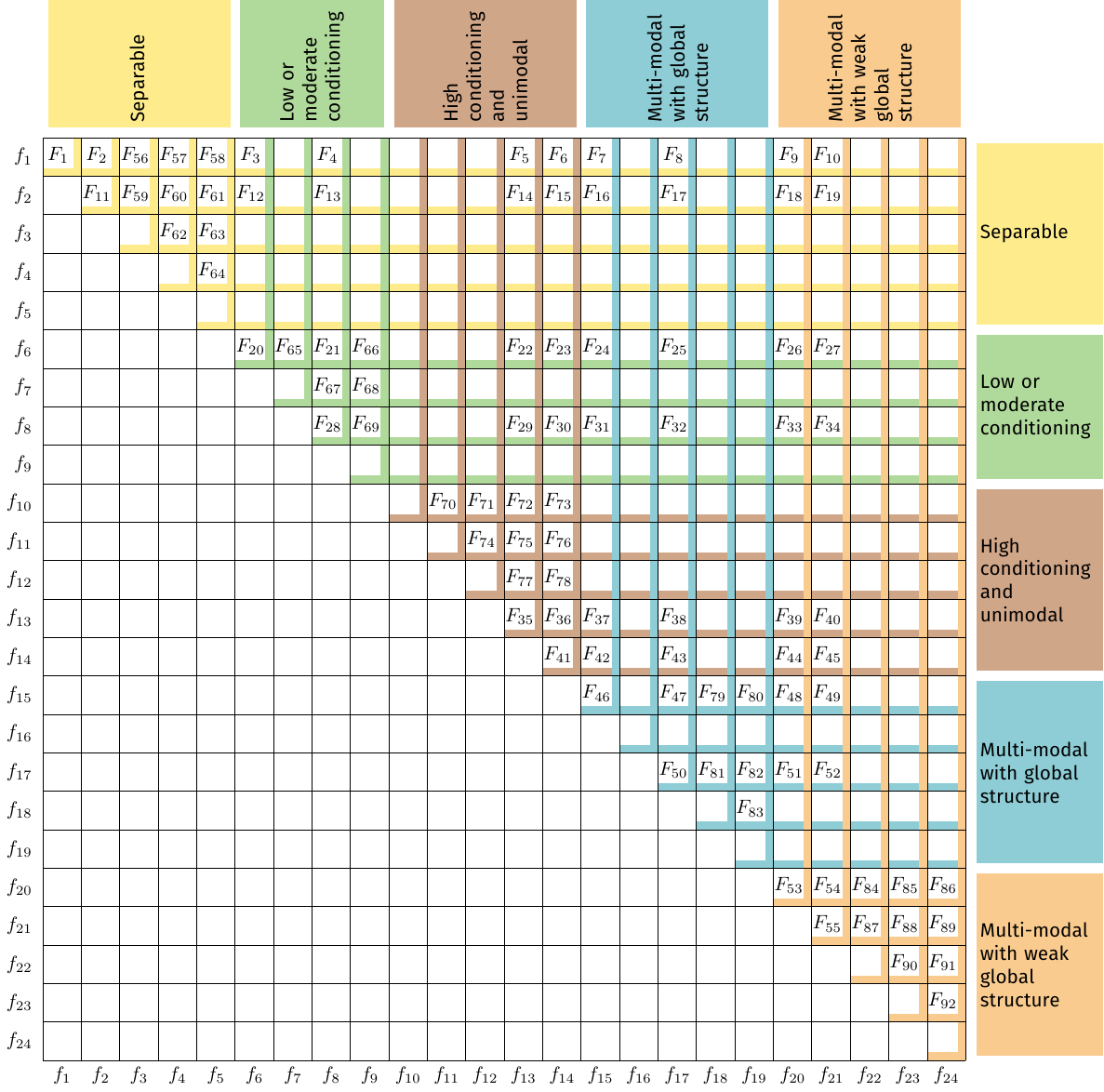}
\caption{The functions of the \sphinxcode{bbob-biobj-ext} test suite ({\hyperref[index:f1]{\(F_{1}\)}} to {\hyperref[index:f92]{\(F_{92}\)}}, in the table cells) together
with the information about which single-objective \sphinxcode{bbob} functions are used
to define them (left and bottom annotations).}\label{index:fig92}\label{index:id72}\end{figure}

\subsubsection{Function Groups}
\label{sec:functiongroupsext}
Like for the \sphinxcode{bbob-biobj} test suite, we obtain 15 function
groups to structure the 92 bi-objective functions of the \sphinxcode{bbob-biobj-ext} test
suite. Depending on whether a function group combines functions from the same or
from different \sphinxcode{bbob} function groups, each function group contains
8, 12 or just four functions. We are listing
below the function groups and in parenthesis the functions that belong to
the respective group:
\begin{enumerate}
\item {} 
separable - separable (12 functions: {\hyperref[index:f1]{\(F_{1}\)}}, {\hyperref[index:f2]{\(F_{2}\)}}, {\hyperref[index:f11]{\(F_{11}\)}}, {\hyperref[index:f56]{\(F_{56}\)}}-{\hyperref[index:f64]{\(F_{64}\)}})

\item {} 
separable - moderate ({\hyperref[index:f3]{\(F_{3}\)}}, {\hyperref[index:f4]{\(F_{4}\)}}, {\hyperref[index:f12]{\(F_{12}\)}}, {\hyperref[index:f13]{\(F_{13}\)}})

\item {} 
separable - ill-conditioned ({\hyperref[index:f5]{\(F_{5}\)}}, {\hyperref[index:f6]{\(F_{6}\)}}, {\hyperref[index:f14]{\(F_{14}\)}}, {\hyperref[index:f15]{\(F_{15}\)}})

\item {} 
separable - multi-modal ({\hyperref[index:f7]{\(F_{7}\)}}, {\hyperref[index:f8]{\(F_{8}\)}}, {\hyperref[index:f16]{\(F_{16}\)}}, {\hyperref[index:f17]{\(F_{17}\)}})

\item {} 
separable - weakly-structured ({\hyperref[index:f9]{\(F_{9}\)}}, {\hyperref[index:f10]{\(F_{10}\)}}, {\hyperref[index:f18]{\(F_{18}\)}}, {\hyperref[index:f19]{\(F_{19}\)}})

\item {} 
moderate - moderate (8 functions: {\hyperref[index:f20]{\(F_{20}\)}}, {\hyperref[index:f21]{\(F_{21}\)}}, {\hyperref[index:f28]{\(F_{28}\)}}, {\hyperref[index:f65]{\(F_{65}\)}}-{\hyperref[index:f69]{\(F_{69}\)}})

\item {} 
moderate - ill-conditioned ({\hyperref[index:f22]{\(F_{22}\)}}, {\hyperref[index:f23]{\(F_{23}\)}}, {\hyperref[index:f29]{\(F_{29}\)}}, {\hyperref[index:f30]{\(F_{30}\)}})

\item {} 
moderate - multi-modal ({\hyperref[index:f24]{\(F_{24}\)}}, {\hyperref[index:f25]{\(F_{25}\)}}, {\hyperref[index:f31]{\(F_{31}\)}}, {\hyperref[index:f32]{\(F_{32}\)}})

\item {} 
moderate - weakly-structured ({\hyperref[index:f26]{\(F_{26}\)}}, {\hyperref[index:f27]{\(F_{27}\)}}, {\hyperref[index:f33]{\(F_{33}\)}}, {\hyperref[index:f34]{\(F_{34}\)}})

\item {} 
ill-conditioned - ill-conditioned (12 functions: {\hyperref[index:f35]{\(F_{35}\)}}, {\hyperref[index:f36]{\(F_{36}\)}}, {\hyperref[index:f41]{\(F_{41}\)}}, {\hyperref[index:f70]{\(F_{70}\)}}-{\hyperref[index:f78]{\(F_{78}\)}})

\item {} 
ill-conditioned - multi-modal ({\hyperref[index:f37]{\(F_{37}\)}}, {\hyperref[index:f38]{\(F_{38}\)}}, {\hyperref[index:f42]{\(F_{42}\)}}, {\hyperref[index:f43]{\(F_{43}\)}})

\item {} 
ill-conditioned - weakly-structured ({\hyperref[index:f39]{\(F_{39}\)}}, {\hyperref[index:f40]{\(F_{40}\)}}, {\hyperref[index:f44]{\(F_{44}\)}}, {\hyperref[index:f45]{\(F_{45}\)}})

\item {} 
multi-modal - multi-modal (8 functions: {\hyperref[index:f46]{\(F_{46}\)}}, {\hyperref[index:f47]{\(F_{47}\)}}, {\hyperref[index:f50]{\(F_{50}\)}}, {\hyperref[index:f79]{\(F_{79}\)}}-{\hyperref[index:f83]{\(F_{83}\)}})

\item {} 
multi-modal - weakly structured ({\hyperref[index:f48]{\(F_{48}\)}}, {\hyperref[index:f49]{\(F_{49}\)}}, {\hyperref[index:f51]{\(F_{51}\)}}, {\hyperref[index:f52]{\(F_{52}\)}})

\item {} 
weakly structured - weakly structured (12 functions: {\hyperref[index:f53]{\(F_{53}\)}}-{\hyperref[index:f55]{\(F_{55}\)}}, {\hyperref[index:f84]{\(F_{84}\)}}-{\hyperref[index:f92]{\(F_{92}\)}})

\end{enumerate}

\subsubsection{Normalization and Instances}
\label{sec:normalization-and-instances-ext}
Normalization of the objectives and instances for the \sphinxcode{bbob-biobj-ext} test suite is handled in the
same manner as for the \sphinxcode{bbob-biobj} suite, i.e., the objective functions are not normalized
and 15 instances are prescribed for a typical experiment.

\subsection{Search Space and Objective Space Plots}
\label{sec:search-space-and-objective-space-plots}
In order to better understand the properties of the 55 \sphinxcode{bbob-biobj} functions, we visualize
the best known Pareto set and Pareto front approximations in the search and objective space, respectively,
providing two plots for each (the approximations are depicted as black points).
The first plot of the search space shows the projection onto a coordinate-axes-parallel cut
defined by two variables, while the second plot contains the projection onto a random cutting plane which contains both
single-objective optima. This second plot additionally shows the contour lines for both objective functions.
Next, the first plot of the objective space is in original scaling (as seen by the algorithm), while the second is in
log-scale, normalized so that the ideal point is at \((0,0)\) and the nadir point is at \((1,1)\).

In addition to the best known Pareto set/Pareto front approximations (in black), all plots show various cuts through the search space:
(i) along a random direction through each single-objective optimum (in blue),
(ii) along each coordinate axis through each single-objective optimum (blue dotted lines),
(iii) along the line connecting both single-objective optima (in red),
(iv) two fully random lines %
\footnote{
of random direction and with a support vector, drawn uniformly at random in \([-4,4]^n\)
} (in yellow), and
(v) a random line in the random projection plane going through both optima %
\footnote{
with a random direction within the plane and a support vector, drawn uniformly at random in \([-4,4]\)
in the coordinate system of the cutting plane
} (in green).

All lines are normalized (of length 10, where the support vector is in the middle). Ticks along the lines indicate the ends of line segments of the same length in search space. Thicker points on the lines
depict solutions that are non-dominated with respect to all points on the same line.
Furthermore, the search space plots highlight the projected region \([-5,5]^n\) as a gray-shaded area while
the gray-shaded area in the objective space plots denotes the region of interest between the ideal (\(\times\)) and
nadir points (\(+\)). Note that, to keep the plots at a manageable size, the Pareto set and Pareto front
approximations are carefully downsampled such that only one solution per grid point is shown---with the
precision of 2 decimals for the search space plots and 3 decimals for the objective space plots to define
the grid. The number of all available and actually displayed solutions is indicated in the legend of each plot.

The figure below shows exemplary plots for three functions of the \sphinxcode{bbob-biobj} suite, the double sphere
problem (\(F_{1}\)) with a continuous Pareto front and a straight line as the Pareto set, the sphere/Gallagher problem
(\(F_{10}\)) with a continuous Pareto front but a gap in the Pareto set, and the double Rastrigin
problem (\(F_{46}\)) for which both Pareto set and Pareto front are discontinuous.
Due to downsampling, the number of displayed points (\(\sim 10000\) or less) is much smaller than the
number of non-dominated solutions contained in the Pareto set approximation (\(2.9\times10^6\) for \(F_{1}\),
\(1.5\times10^6\) for \(F_{10}\) and \(3.1\times10^5\) for \(F_{46}\)).
\begin{figure}
  \centering
\includegraphics[trim={1.8cm 0 2.6cm 0},clip,width=0.32\linewidth]{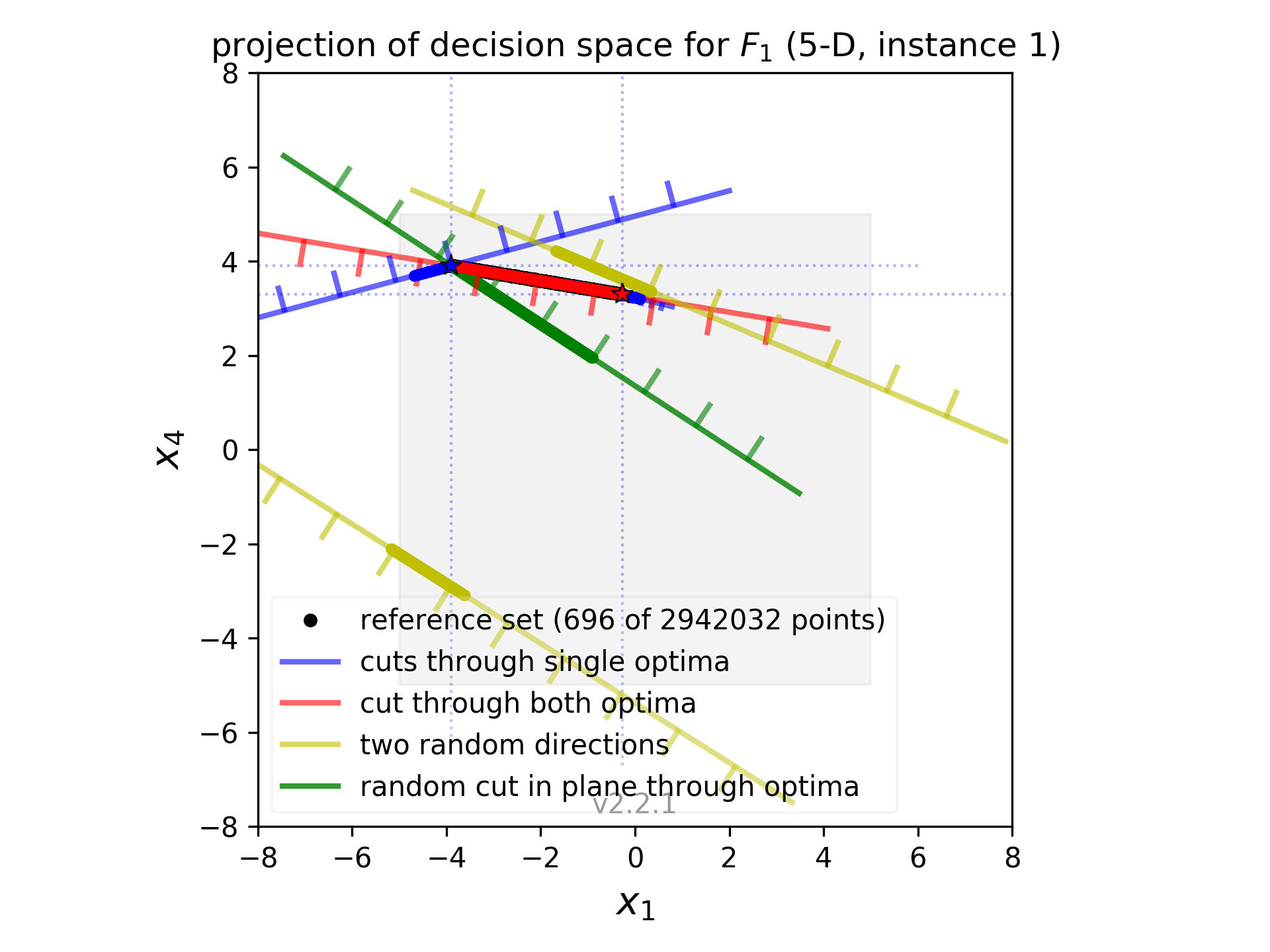} 
\includegraphics[trim={1.8cm 0 2.6cm 0},clip,width=0.32\linewidth]{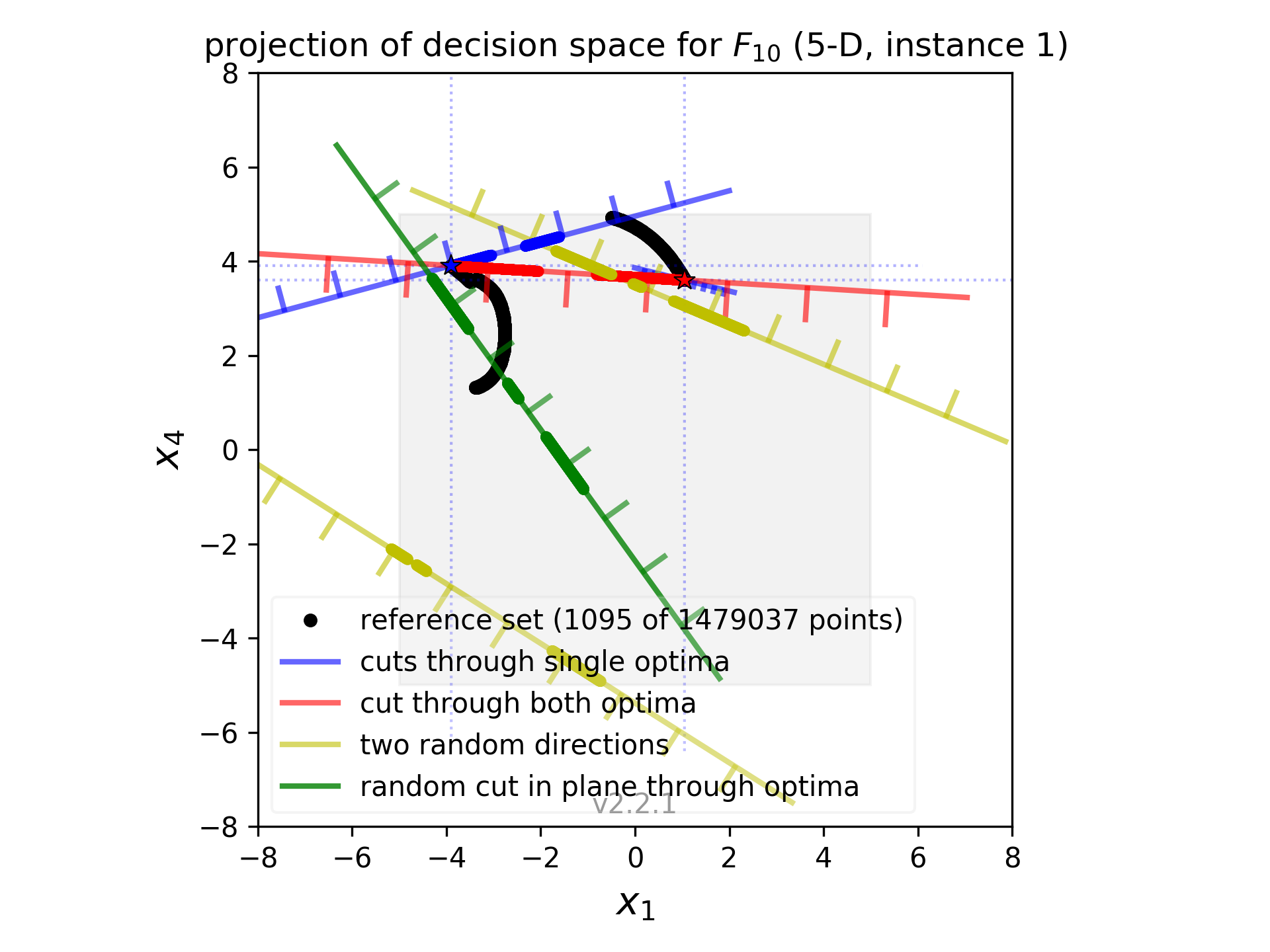} 
\includegraphics[trim={1.8cm 0 2.6cm 0},clip,width=0.32\linewidth]{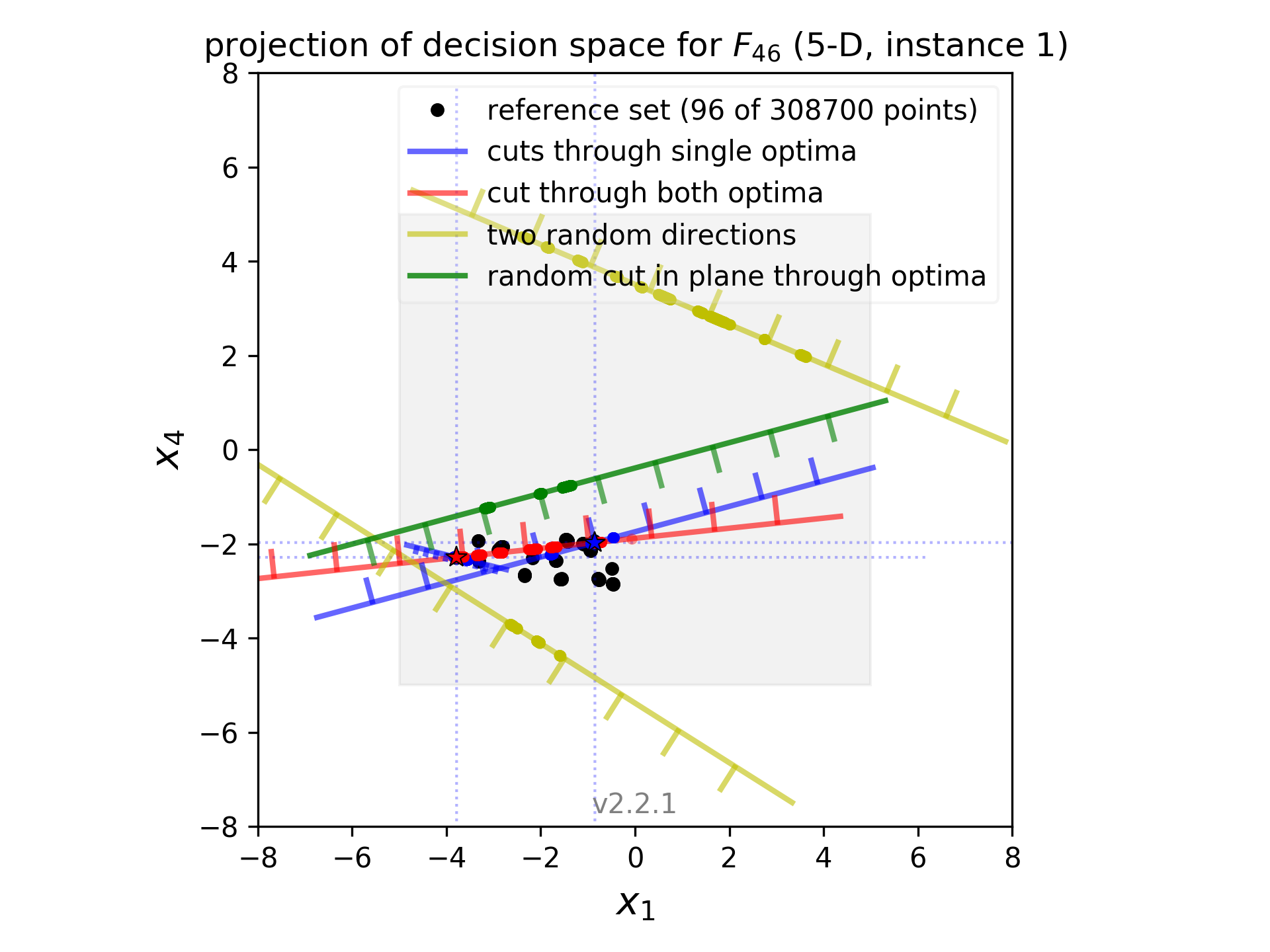}

\includegraphics[trim={1.8cm 0 2.6cm 0},clip,width=0.32\linewidth]{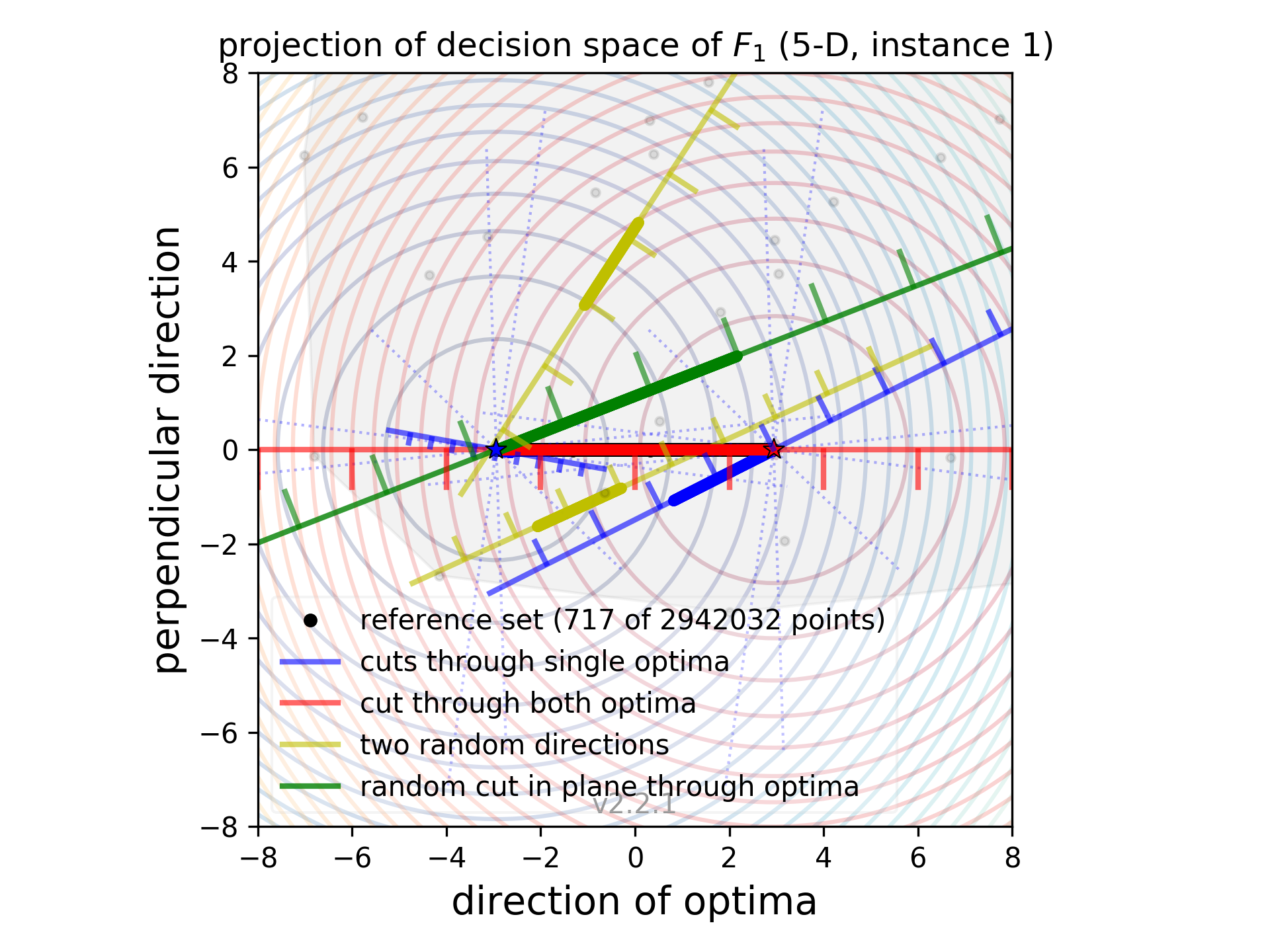} 
\includegraphics[trim={1.8cm 0 2.6cm 0},clip,width=0.32\linewidth]{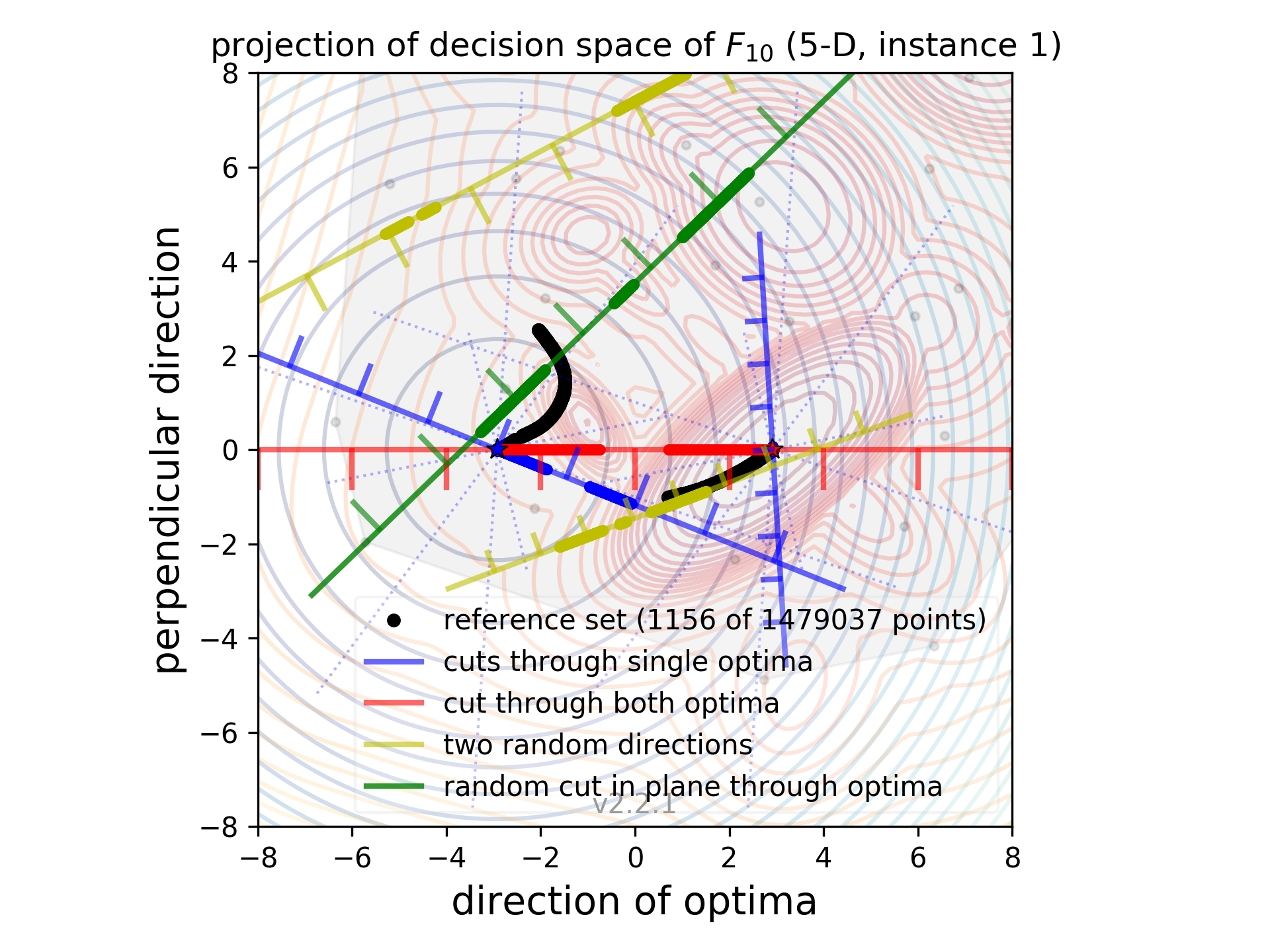}
\includegraphics[trim={1.8cm 0 2.6cm 0},clip,width=0.32\linewidth]{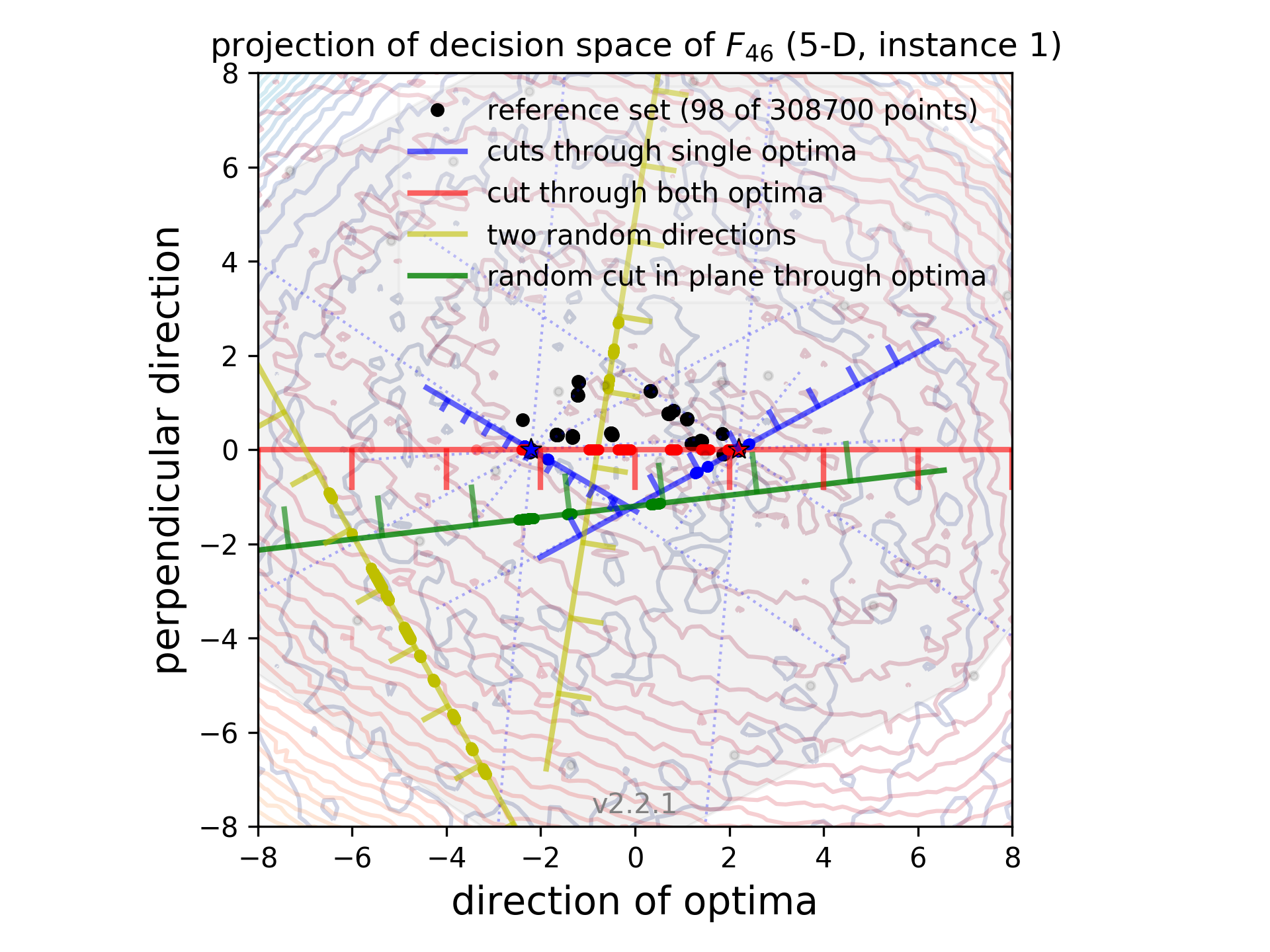}

\includegraphics[trim={1.4cm 0 2.6cm 0},clip,width=0.32\linewidth]{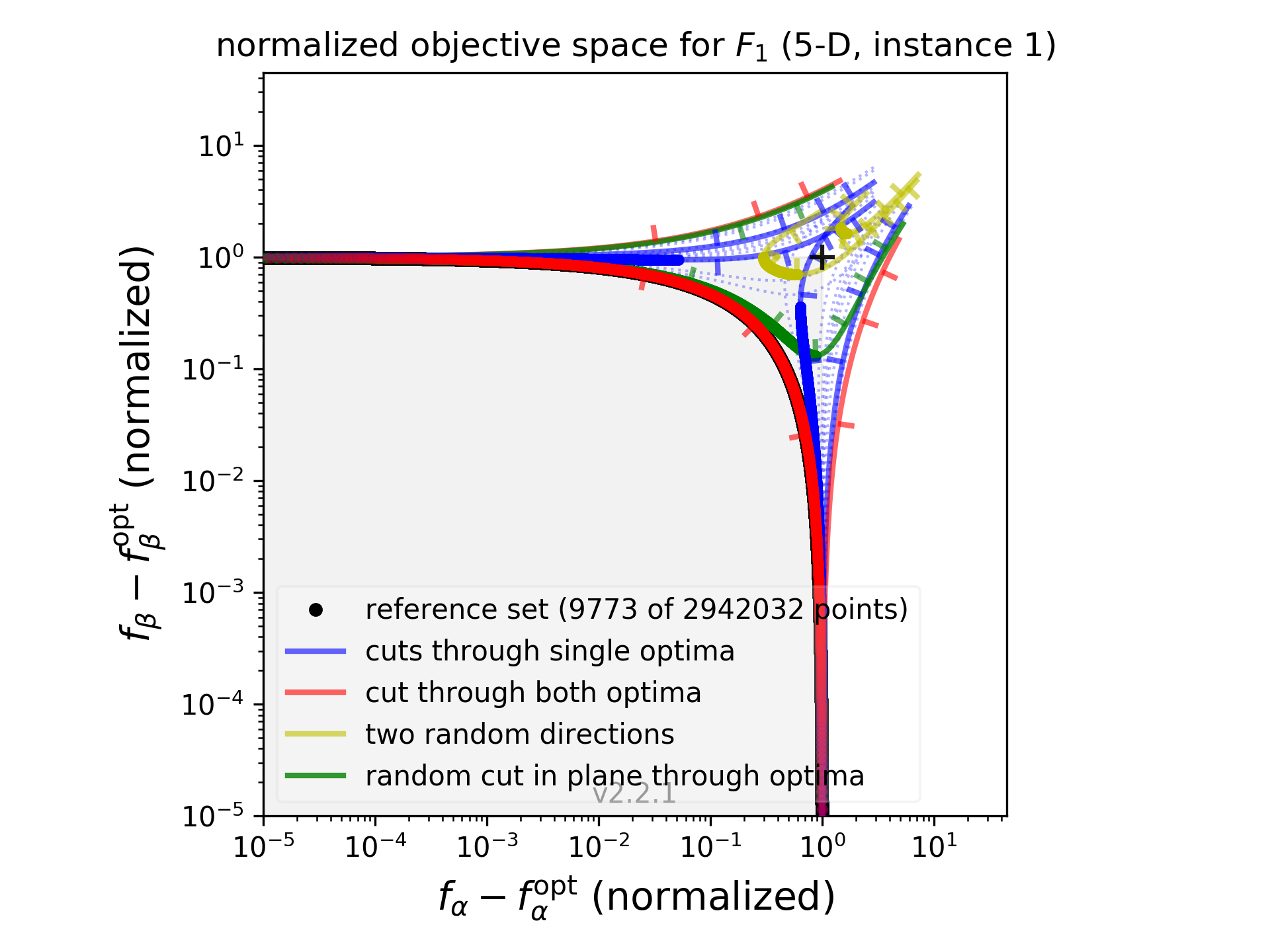}
\includegraphics[trim={1.4cm 0 2.6cm 0},clip,width=0.32\linewidth]{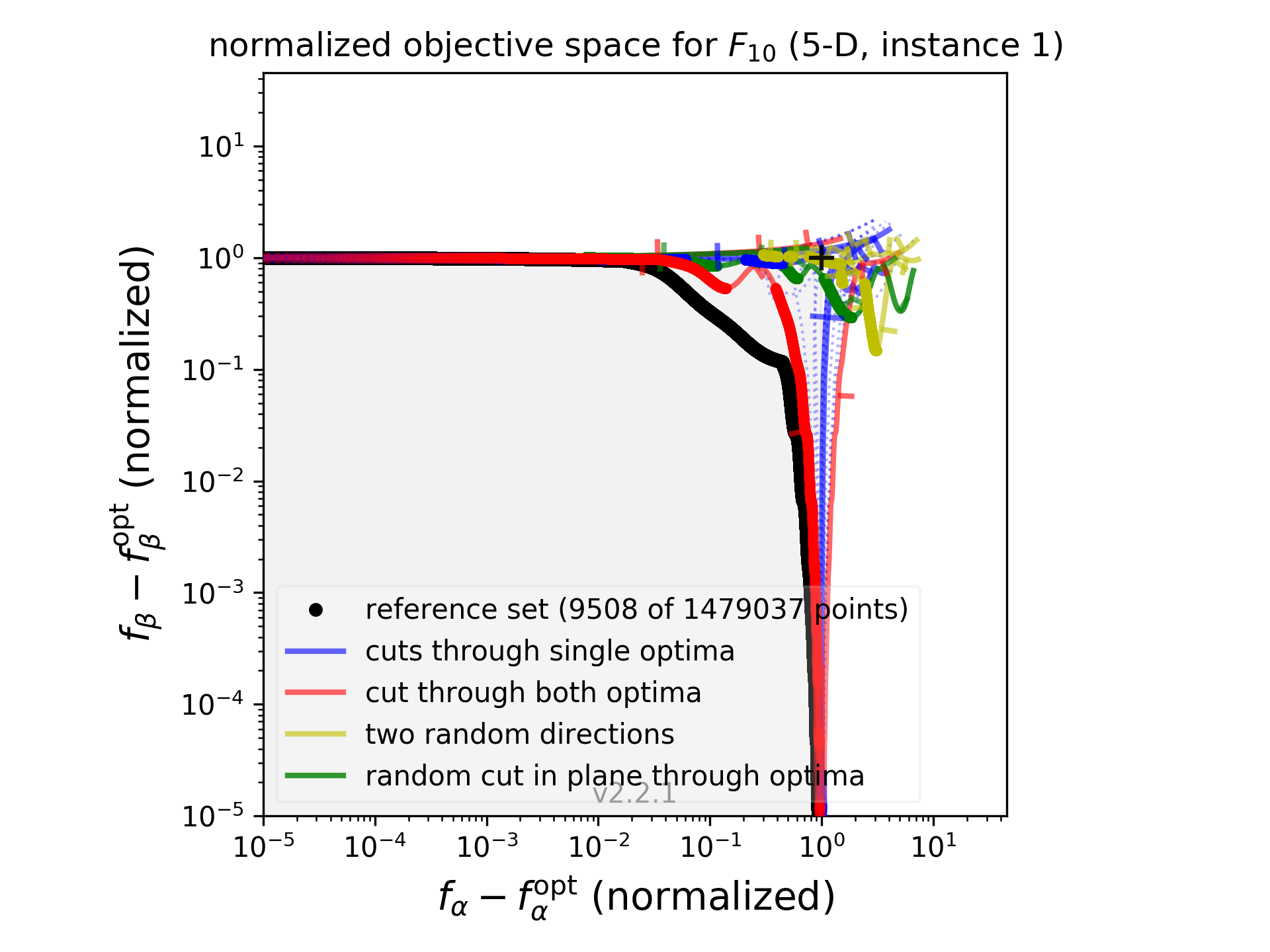}
\includegraphics[trim={1.4cm 0 2.6cm 0},clip,width=0.32\linewidth]{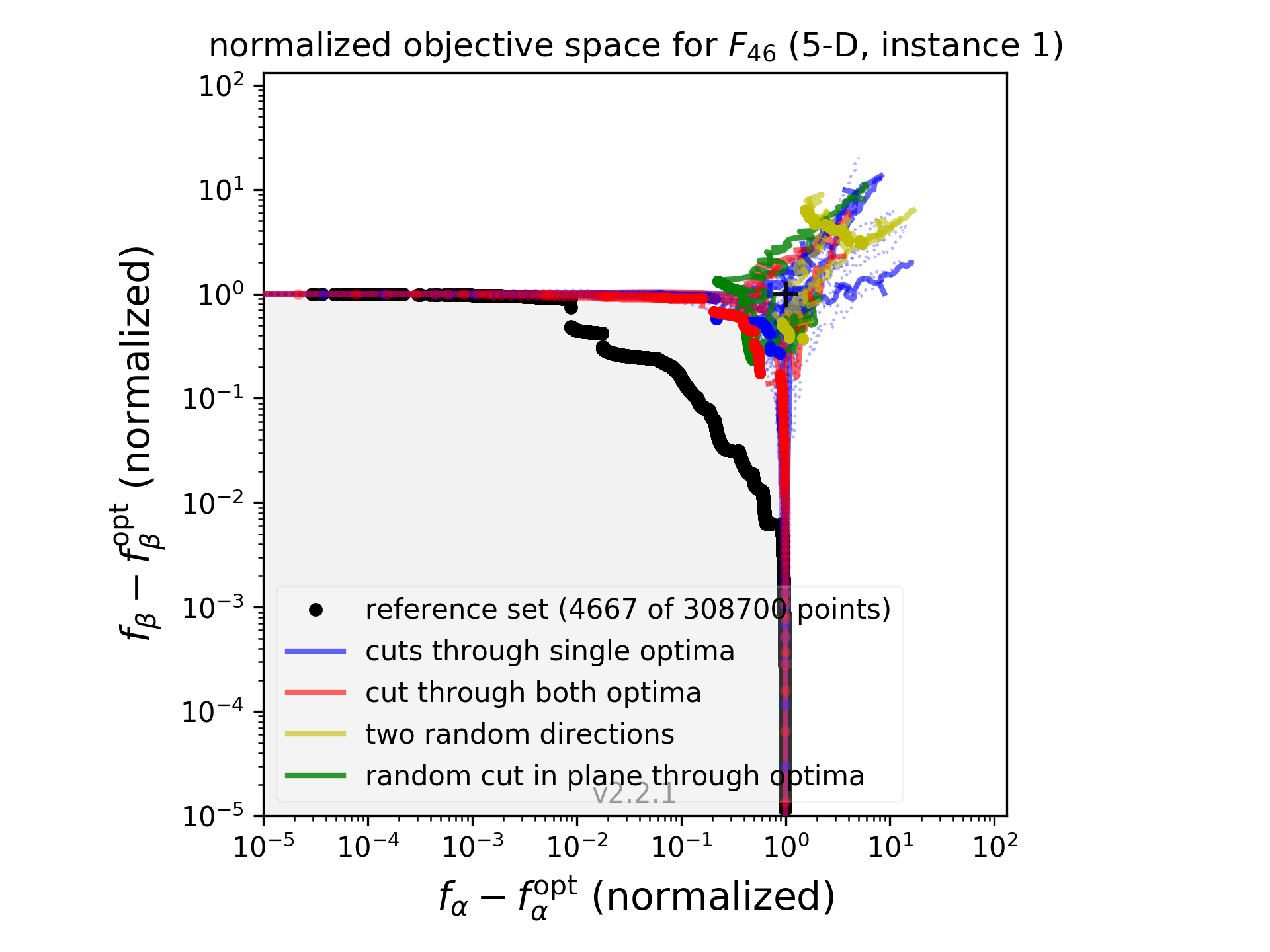}

\includegraphics[trim={1.4cm 0 2.6cm 0},clip,width=0.32\linewidth]{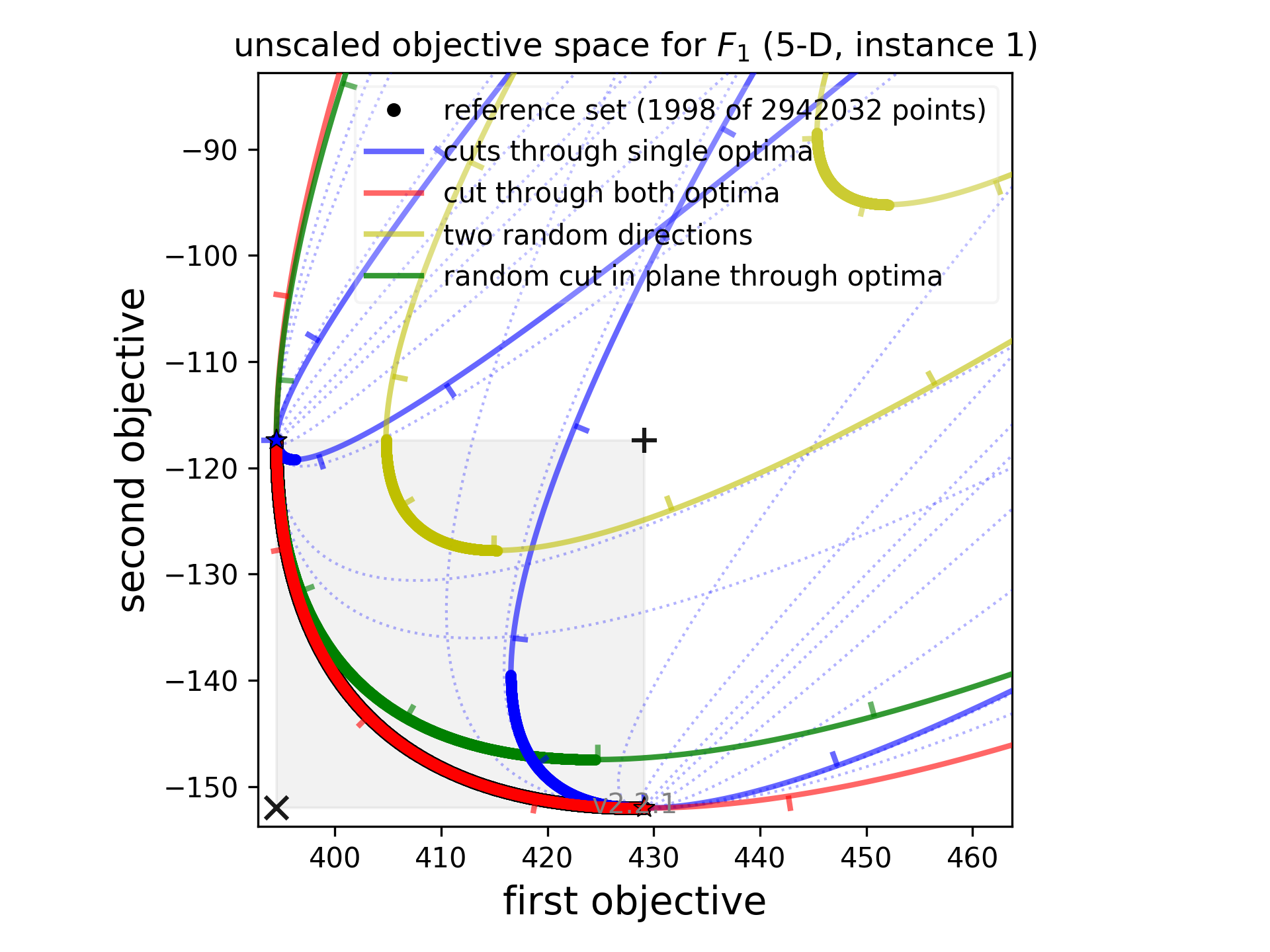}
\includegraphics[trim={1.4cm 0 2.6cm 0},clip,width=0.32\linewidth]{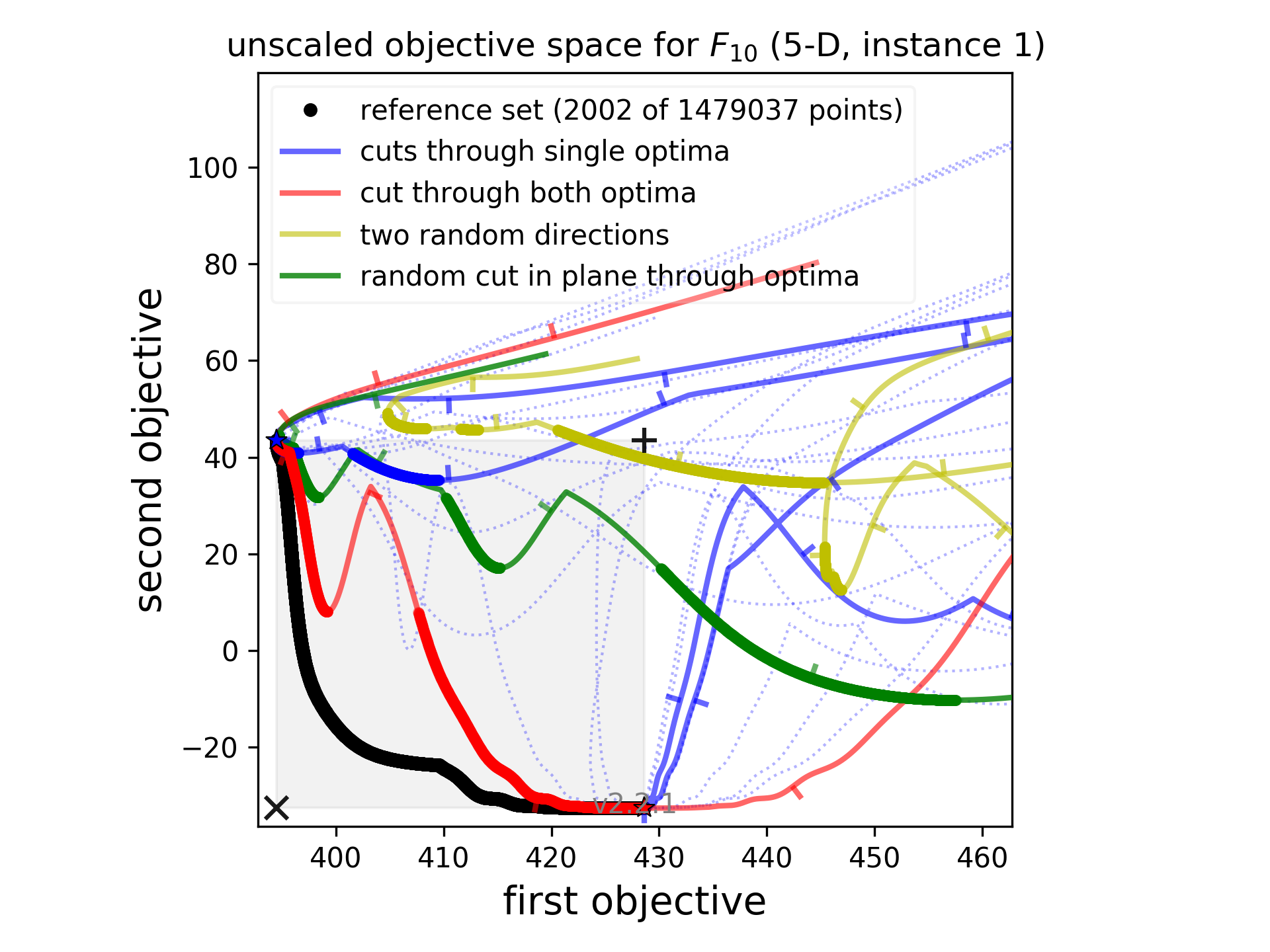}
\includegraphics[trim={1.4cm 0 2.6cm 0},clip,width=0.32\linewidth]{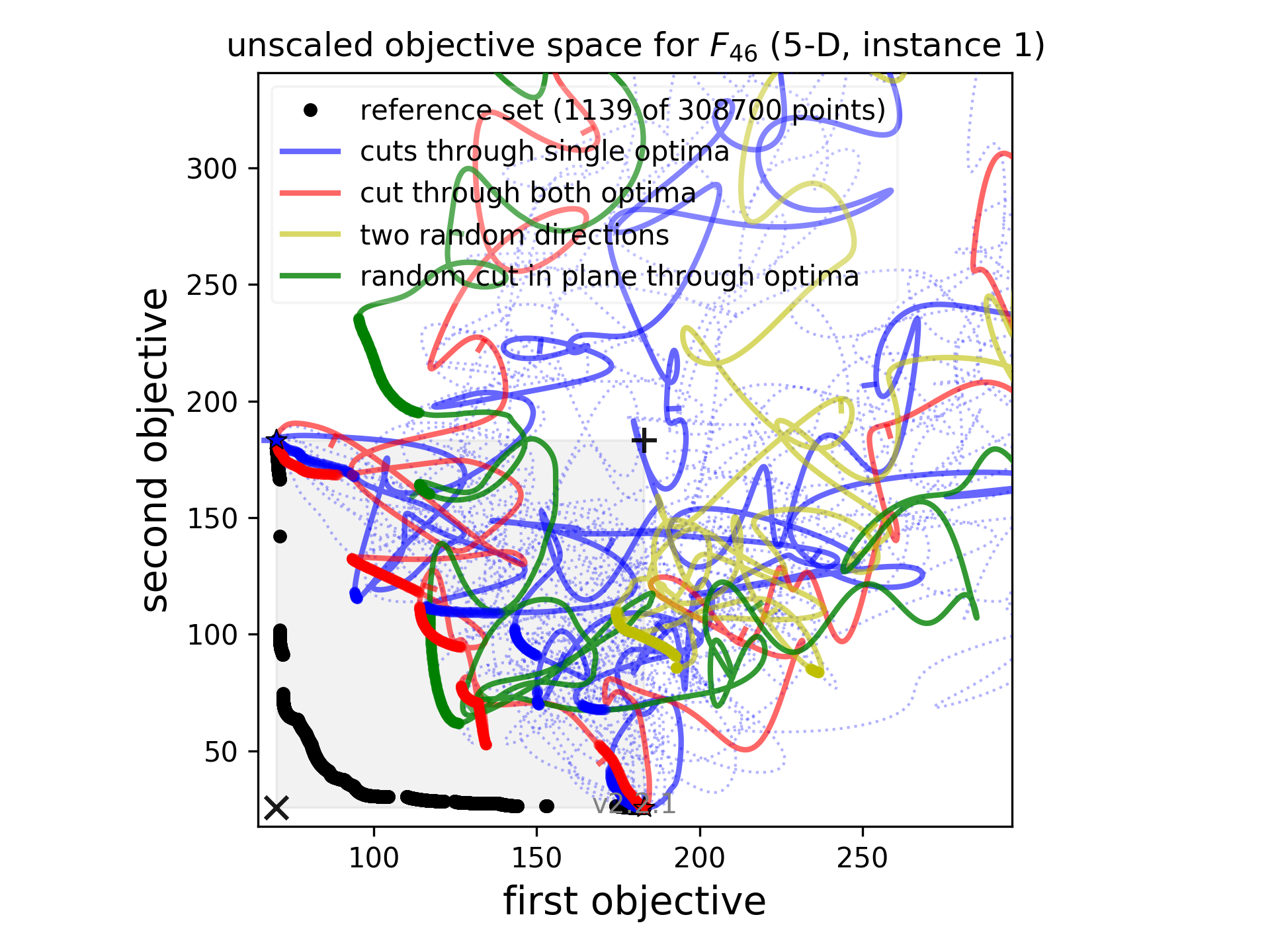}
\caption{
Illustration of the search space (first two rows) and the objective space (third row:
normalized in log-scale; forth row: original scaling) for \code{bbob-biobj}
functions $F_{1}$ (left column), $F_{10}$ (middle column), and $F_{46}$ (right column) in dimension 5 for the first instance.
}
\end{figure}
For links to the illustrations of \sphinxstyleemphasis{all} \sphinxcode{bbob-biobj} and \sphinxcode{bbob-biobj-ext} functions, we refer to the accompanying paper at \url{http://bbobbiobj.gforge.inria.fr/bbob-biobj-functions.pdf} and the accompanying web page \url{http://bbobbiobj.gforge.inria.fr/}.
All plots shown and linked in this paper are provided only for the first instance of dimension 5, but
further plots for more instances and dimensions are provided online at \url{http://bbobbiobj.gforge.inria.fr/}. This web page also provides the link to the python source code that was used to produce the plots.

\section{Extension to Any Number of Objectives}
\label{sec:anyobj}
The above proposed test suites are defined for two objective functions. There is however no general restriction
of the proposed construction to only two objective functions. Yet, the number of resulting test problems
will be practically too high if we extend the \sphinxcode{bbob-biobj} and \sphinxcode{bbob-biobj} suites naively.

The naive approach of combining all potential \(s=24\) \sphinxcode{bbob} functions with each other (with repetitions
but without caring about the order of the chosen functions), results in \(\binom{s+m-1}{m}\)
\(m\)-objective problems overall---which equals the number of \(m\)-combinations with repetitions, or
\(m\)-multicombinations of \(s\) objective functions. Even restricting the number of used objective
functions to \(s=10\), like in the \sphinxcode{bbob-biobj} suite, results in \(\binom{10+3-1}{3}=220\)
combinations for three objectives, \(\binom{10+4-1}{4}=715\) combinations for four objectives and
\(\binom{10+5-1}{5}=2\,002\) combinations for five objectives. Multiplied by
the six dimensions and 15 instances as recommended above, this would require 180\,180
experiments to run the entire \sphinxcode{bbob-biobj} suite in five objectives---an unsuitable number
for practical experiments.

The goal is therefore to find an approach for combining single-objective functions to form \(m\)-objective
problems that results in a suite of manageable size that contains different combinations
of real-world function properties.
Here, we suggest to follow the construction of the \sphinxcode{bbob-biobj-ext} suite and define a number of
function groups (dependent on the number of objectives \(m\)) while restricting the number of
problems within one group. With this approach, the combinatorial explosion of the number of
produced \(m\)-objective problems is less pronounced than with a naive approach, i.e., problem
suites with up to eight objectives still have reasonable size.

Concretely, we propose the following multiobjective test suite \sphinxstyleemphasis{generator} to define
multiobjective test suites with \(m\) objective functions and arbitrary dimensions and instances.
First, we define \(\binom{5+m-1}{m}\) function groups by assigning to each of the \(m\) objective
functions one of the five \sphinxcode{bbob} function groups (with repetitions but without caring about the order).
This will result in groups such as \sphinxstyleemphasis{separable - separable - separable} or \sphinxstyleemphasis{separable - ill-conditioned - multimodal}
for three objectives. Then, we fill these function groups by sampling for each objective uniformly
at random a \sphinxcode{bbob} function from the corresponding single-objective function group (with the exception of \(f_{16}\), as
before). Following the structure of the \sphinxcode{bbob-biobj-ext} suite, we recommend to produce more problems when
all objective functions are from the same function group and significantly less for the other groups. Concretely,
we suggest to sample 12 functions when the function group of all objectives is \sphinxstyleemphasis{separable}, \sphinxstyleemphasis{ill-conditioned} or
\sphinxstyleemphasis{weakly-structured} and 8 functions when the function group of all objectives is \sphinxstyleemphasis{moderate} or \sphinxstyleemphasis{multi-modal}.
In the cases where all objectives' function groups are different, we sample only four problems and
in all other cases, we sample only twice. This way, we reproduce the structure of the \sphinxcode{bbob-biobj-ext} suite
when the number of objectives is two but reduce the number of problems in the suite in higher dimensions.

Since with six objectives or more, all objectives cannot come from different function groups,
our construction
results in \(12+8+12+8+12 + 2\cdot(\binom{5+m-1}{m}-5)\) functions in total if \(m>5\).
The proposed multiobjective suite will therefore have 462 (parametrized) functions for six objectives,
702 for seven objectives, 1032 for eight objectives, and so forth. These numbers are still very high,
especially since multiple search space dimensions and instances need to be considered. For example,
a `standard' setting of the \sphinxcode{bbob-biobj} suite comprises six dimensions and 15 instances per
function. Running experiments
with only four dimensions and five instances, however, would result in a (more) reasonable number of
\(462\cdot 4 \cdot 5 = 9\,240\) experiments for a six-objective test suite compared to the naive approach
with \(\binom{10+6-1}{6} \cdot 4 \cdot 5 = 100\,100\) experiments. Note that, in comparison, an entire
experiment on the 92 \sphinxcode{bbob-biobj-ext} functions with six dimensions and 15 instances also corresponds to 
already \(92\cdot 6 \cdot 15 = 8\,280\) single experiments.

\begin{table}
\centering
\caption{Concrete numbers for the size of the test function suites when combining
the \sphinxstyleliteralintitle{bbob} functions (without the Weierstrass function) to \(m\)-objective problems.
The \sphinxstyleemphasis{all combinations} column gives the numbers \(\binom{s+m-1}{m}\) for the naive
implementation with all possible function combinations (of either \(s=10\) or \(s=23\)
\sphinxstyleliteralintitle{bbob} functions) and the
\sphinxstyleemphasis{proposed suite construction} column gives the numbers for the approach proposed here
which decides first on the objectives' function groups and then samples functions
from the groups with the number of objectives in the first column.
If we assume that the test suites
contain scalable problems in \(n_d\) different dimensions with \(n_i\) instances
per function/dimension pair, the given numbers must be multiplied by \(n_d \cdot n_i\).}\label{index:id73}
\vspace{0.5em}

\begin{tabulary}{\linewidth}{CRRR}
\toprule
\multirow{2}{*}{\#obj} & \multicolumn{2}{c}{all combinations (with repetitions)} & proposed suite construction \\
                       & \multicolumn{1}{c}{based on 10 functions} & \multicolumn{1}{c}{based on 23 functions} & based on 23 functions\\
\midrule
 2 &     55 &        276 &    92\\
 3 &    220 &      2,300 &   132\\
 4 &    715 &     14,950 &   192\\
 5 &  2,002 &     80,730 &   296\\
 6 &  5,005 &    376,740 &   462\\
 7 & 11,440 &  1,560,780 &   702\\
 8 & 24,310 &  5,852,925 & 1,032\\
 9 & 48,620 & 20,160,075 & 1,472\\
10 & 92,378 & 64,512,240 & 2,044\\ \bottomrule
\end{tabulary}
\end{table}

Note that the random choice of objective functions should rather be pseudo-random in an actual
implementation of the proposed multiobjective suite. The random seed to sample the objective functions
can be chosen as the objective function index plus a given constant, e.g. \(100\,000+i\) to sample
the objective function \(\bar{f}_i\) of an \(m\)-objective function
\(F = (\bar{f}_1, \ldots, \bar{f}_i, \ldots, \bar{f}_m)\). Exceptions to this rule are expected, for example,
in order to always have the well-understood multi-sphere function %
\footnote{
In which each objective is an instance of the sphere function.
} in the test suite or to have the \sphinxcode{bbob-biobj-ext} suite
become a special case of the general multiobjective suite if the number of objectives is two.

\section{Conclusions}
\label{sec:conclusions}
Designing test suites is a crucial part of benchmarking optimization algorithms.
Arguably, the most problematic aspect of using \sphinxstyleemphasis{artificial} test functions to assess performance is the
\sphinxstyleemphasis{representativeness} of these regarding difficulties observed in real-world problems.
In this paper, we suggest to address the problem of representativeness in the multiobjective case
by combining established single-objective test functions with known difficulties observed
in practice. Following the concepts of the single-objective \sphinxcode{bbob} test suite, we propose
two concrete bi-objective test suites and a test suite \sphinxstyleemphasis{generator} for arbitrary numbers of objectives
based on the same idea of combining existing single-objective functions.

Our approach contrasts most of the existing test suites for multiobjective optimization.
These are based on the desirable property of having well-understood Pareto sets and Pareto fronts
with analytical forms but have, on the other hand, artificial
characteristics that are arguably under-represented in real-world problems. Examples of such properties
are separability, optima located exactly at the boundary constraints, and the existence of variables
that solely control the distance between a solution and the Pareto front.

The disadvantage of unknown analytical forms of the Pareto sets and Pareto fronts in our
proposal is addressed by collecting the non-dominated solutions from extensive experiments with dozens of
different optimization algorithms and providing and visualizing the
Pareto set and Pareto front approximations for each problem.
These visualizations lead to new insights into how such non-analytical
Pareto sets and Pareto fronts may look in practice.

\section*{Acknowledgments}
This work was supported by the grant ANR-12-MONU-0009 (NumBBO)
of the French National Research Agency. We also thank Ilya Loshchilov and Oswin Krause for their
initial suggestions on how to extend the \sphinxcode{bbob-biobj} test suite.

Tea Tu\v{s}ar acknowledges the financial support from the Slovenian Research Agency (project No.
Z2-8177). This work is part of a project that has received funding from the \sphinxstyleemphasis{European Union's
Horizon 2020 research and innovation program} under grant agreement No.692286.

\bibliographystyle{apalike}
\bibliography{bibtex}

\vfill

\pagebreak

\section{Appendix}
\label{sec:appendix}

\subsection{Definitions and Plots of the Proposed Functions}
\label{sec:definitions-and-plots-of-the-proposed-functions}

\subsubsection[\texorpdfstring{\protect\(F_1\protect\): Sphere/Sphere}{F1: Sphere/Sphere}]{\texorpdfstring{\protect\(F_1\protect\): Sphere/Sphere}{}}
\label{index:f1}\label{index:sphere-sphere}
Combination of two sphere functions (\href{http://coco.lri.fr/downloads/download15.03/bbobdocfunctions.pdf\#page=5}{\(f_1\) in the \sphinxcode{bbob} suite}).

\noindent Both objectives are unimodal, highly symmetric, rotational and scale
invariant. The Pareto set is known to be a straight line and the Pareto
front is convex. Furthermore, the normalized hypervolume value of the
entire Pareto front with respect to the nadir point as reference point
can be computed analytically as the integral
\(1-\int_{0}^{1} (1-\sqrt{x})^2dx = -\frac{1}{2}+\frac{4}{3}=0.833333\ldots\).

Considered as the simplest bi-objective problem in
continuous domain.
Contained in the \sphinxstyleemphasis{separable - separable} function group.

% OLD CODE TO HAVE INLINE GRAPHICS (BUT FILESIZE WAS TOO LARGE FOR SUBMISSION):
%\newcommand{\bbobfigs}[5]{
%\begin{figure}[h]
%\includegraphics[width=0.490\linewidth]{../code/plots/plots-d#2-i#3-March2018/directions-f#1-i#3-d#2-searchspace.png}
%\includegraphics[width=0.490\linewidth]{../code/plots/plots-d#2-i#3-March2018/directions-f#1-i#3-d#2-searchspace-projection.png}
%\caption{Illustration of search space for \code{bbob-biobj} function $F_{#4}$ in dimension #5 for the first instance.}
%\end{figure}
%\begin{figure}[h]
%\includegraphics[width=0.490\linewidth]{../code/plots/plots-d#2-i#3-March2018/directions-f#1-i#3-d#2-logobjspace.png}
%\includegraphics[width=0.490\linewidth]{../code/plots/plots-d#2-i#3-March2018/directions-f#1-i#3-d#2-objspace.png}
%\caption{Illustration of objective space for \code{bbob-biobj} function $F_{#4}$ in dimension #5 for the
%first instance (left: normalized in log-scale; right: original scaling).}
%\end{figure}
%\pagebreak
%}

\newcommand{\bbobfigs}[5]{
\vspace{1cm}
\noindent Links to illustrations of function $F_{#4}$ in dimension #5 for the first instance:
\begin{itemize}
   \item \href{http://bbobbiobj.gforge.inria.fr/plots-d#2-i#3-March2018/directions-f#1-i#3-d#2-searchspace.png}{Search space plot along two coordinate axes}
   \item \href{http://bbobbiobj.gforge.inria.fr/plots-d#2-i#3-March2018/directions-f#1-i#3-d#2-searchspace-projection.png}{Projection of the search space into a hyperplane through both optima}
   \item \href{http://bbobbiobj.gforge.inria.fr/plots-d#2-i#3-March2018/directions-f#1-i#3-d#2-logobjspace.png}{Normalized objective space plot (in log-scale)}
   \item \href{http://bbobbiobj.gforge.inria.fr/plots-d#2-i#3-March2018/directions-f#1-i#3-d#2-logobjspace.png}{Unscaled objective space plot}
\end{itemize}
\vfill
\pagebreak
}

% #1: function string (2 digits)
% #2: dimension string (2 digits)
% #3: instance string (2 digits)
% #4: function number
% #5: dimension
\bbobfigs{01}{05}{01}{1}{5}

\subsubsection[\texorpdfstring{\protect\(F_2\protect\): Sphere/Ellipsoid separable}{F2: Sphere/Ellipsoid separable}]{\texorpdfstring{\protect\(F_2\protect\): Sphere/Ellipsoid separable}{}}
\label{index:f2}\label{index:sphere-ellipsoid-separable}
Combination of the sphere function (\href{http://coco.lri.fr/downloads/download15.03/bbobdocfunctions.pdf\#page=5}{\(f_1\) in the \sphinxcode{bbob} suite})
and the separable ellipsoid function (\href{http://coco.lri.fr/downloads/download15.03/bbobdocfunctions.pdf\#page=10}{\(f_2\) in the \sphinxcode{bbob} suite}).

Both objectives are unimodal and separable. While the first objective is
truly convex-quadratic with a condition number of 1, the second
objective is only globally quadratic with smooth local
irregularities and highly ill-conditioned with a condition number of
about \(10^6\).

Contained in the \sphinxstyleemphasis{separable - separable} function group.

% #1: function string (2 digits)
% #2: dimension string (2 digits)
% #3: instance string (2 digits)
% #4: function number
% #5: dimension
\bbobfigs{02}{05}{01}{2}{5}

\subsubsection[\texorpdfstring{\protect\(F_3\protect\): Sphere/Attractive sector}{F3: Sphere/Attractive sector}]{\texorpdfstring{\protect\(F_3\protect\): Sphere/Attractive sector}{}}
\label{index:f3}\label{index:sphere-attractive-sector}
Combination of the sphere function (\href{http://coco.lri.fr/downloads/download15.03/bbobdocfunctions.pdf\#page=5}{\(f_1\) in the \sphinxcode{bbob} suite})
and the attractive sector function (\href{http://coco.lri.fr/downloads/download15.03/bbobdocfunctions.pdf\#page=30}{\(f_6\) in the \sphinxcode{bbob} suite}).

Both objective functions are unimodal, but only the first objective is
separable and truly convex quadratic. The attractive sector
function is highly asymmetric, where only one \sphinxstyleemphasis{hypercone} (with
angular base area) with a volume of roughly \((1/2)^n\)
yields low function values. The optimum of it is located at the tip
of this cone.

Contained in the \sphinxstyleemphasis{separable - moderate} function group.

% #1: function string (2 digits)
% #2: dimension string (2 digits)
% #3: instance string (2 digits)
% #4: function number
% #5: dimension
\bbobfigs{03}{05}{01}{3}{5}

\subsubsection[\texorpdfstring{\protect\(F_4\protect\): Sphere/Rosenbrock original}{F4: Sphere/Rosenbrock original}]{\texorpdfstring{\protect\(F_4\protect\): Sphere/Rosenbrock original}{}}
\label{index:sphere-rosenbrock-original}\label{index:f4}
Combination of the sphere function (\href{http://coco.lri.fr/downloads/download15.03/bbobdocfunctions.pdf\#page=5}{\(f_1\) in the \sphinxcode{bbob} suite})
and the original, i.e., unrotated Rosenbrock function (\href{http://coco.lri.fr/downloads/download15.03/bbobdocfunctions.pdf\#page=40}{\(f_8\) in the \sphinxcode{bbob} suite}).

The first objective is separable and truly convex, the second
objective is partially separable (tri-band structure). The first
objective is unimodal while the second objective has a local
optimum with an attraction volume of about 25\%.

Contained in the \sphinxstyleemphasis{separable - moderate} function group.

% #1: function string (2 digits)
% #2: dimension string (2 digits)
% #3: instance string (2 digits)
% #4: function number
% #5: dimension
\bbobfigs{04}{05}{01}{4}{5}

\subsubsection[\texorpdfstring{\protect\(F_5\protect\): Sphere/Sharp ridge}{F5: Sphere/Sharp ridge}]{\texorpdfstring{\protect\(F_5\protect\): Sphere/Sharp ridge}{}}
\label{index:sphere-sharp-ridge}\label{index:f5}
Combination of the sphere function (\href{http://coco.lri.fr/downloads/download15.03/bbobdocfunctions.pdf\#page=5}{\(f_1\) in the \sphinxcode{bbob} suite})
and the sharp ridge function (\href{http://coco.lri.fr/downloads/download15.03/bbobdocfunctions.pdf\#page=65}{\(f_{13}\) in the \sphinxcode{bbob} suite}).

Both objective functions are unimodal.
In addition to the simple, separable, and differentiable first
objective, a sharp, i.e., non-differentiable ridge has to be
followed for optimizing the (non-separable) second objective. The
gradient towards the ridge remains constant, when the ridge is
approached from a given point.
Approaching the ridge is initially effective, but becomes ineffective
close to the ridge when the ridge needs to be followed in direction
to its optimum.  The necessary change in \sphinxstyleemphasis{search behavior} close to
the ridge is difficult to diagnose, because the gradient
towards the ridge does not flatten out.

Contained in the \sphinxstyleemphasis{separable - ill-conditioned} function group.

% #1: function string (2 digits)
% #2: dimension string (2 digits)
% #3: instance string (2 digits)
% #4: function number
% #5: dimension
\bbobfigs{05}{05}{01}{5}{5}

\subsubsection[\texorpdfstring{\protect\(F_6\protect\): Sphere/Sum of Different Powers}{F6: Sphere/Sum of Different Powers}]{\texorpdfstring{\protect\(F_6\protect\): Sphere/Sum of Different Powers}{}}
\label{index:sphere-sum-of-different-powers}\label{index:f6}
Combination of the sphere function (\href{http://coco.lri.fr/downloads/download15.03/bbobdocfunctions.pdf\#page=5}{\(f_1\) in the \sphinxcode{bbob} suite})
and the sum of different powers function (\href{http://coco.lri.fr/downloads/download15.03/bbobdocfunctions.pdf\#page=70}{\(f_{14}\) in the \sphinxcode{bbob} suite}).

Both objective functions are unimodal. The first objective is
separable, the second non-separable.
When approaching the second objective's optimum, the difference
in sensitivity between different directions in search space
increases unboundedly.

Contained in the \sphinxstyleemphasis{separable - ill-conditioned} function group.

% #1: function string (2 digits)
% #2: dimension string (2 digits)
% #3: instance string (2 digits)
% #4: function number
% #5: dimension
\bbobfigs{06}{05}{01}{6}{5}

\subsubsection[\texorpdfstring{\protect\(F_7\protect\): Sphere/Rastrigin}{F7: Sphere/Rastrigin}]{\texorpdfstring{\protect\(F_7\protect\): Sphere/Rastrigin}{}}
\label{index:sphere-rastrigin}\label{index:f7}
Combination of the sphere function (\href{http://coco.lri.fr/downloads/download15.03/bbobdocfunctions.pdf\#page=5}{\(f_1\) in the \sphinxcode{bbob} suite})
and the Rastrigin function (\href{http://coco.lri.fr/downloads/download15.03/bbobdocfunctions.pdf\#page=75}{\(f_{15}\) in the \sphinxcode{bbob} suite}).

In addition to the simple sphere function, the prototypical highly
multimodal Rastrigin function needs to be solved which has originally
a very regular and symmetric structure for the placement of the optima.
Here, however, transformations are performed to alleviate
the original symmetry and regularity in the second objective.

The properties of the second objective contain non-separability,
multimodality (roughly \(10^n\) local optima), a conditioning of
about 10, and a large global amplitude compared to the local amplitudes.

Contained in the \sphinxstyleemphasis{separable - multi-modal} function group.

% #1: function string (2 digits)
% #2: dimension string (2 digits)
% #3: instance string (2 digits)
% #4: function number
% #5: dimension
\bbobfigs{07}{05}{01}{7}{5}

\subsubsection[\texorpdfstring{\protect\(F_8\protect\): Sphere/Schaffer F7, condition 10}{F8: Sphere/Schaffer F7, condition 10}]{\texorpdfstring{\protect\(F_8\protect\): Sphere/Schaffer F7, condition 10}{}}
\label{index:f8}\label{index:sphere-schaffer-f7-condition-10}
Combination of the sphere function (\href{http://coco.lri.fr/downloads/download15.03/bbobdocfunctions.pdf\#page=5}{\(f_1\) in the \sphinxcode{bbob} suite})
and the Schaffer F7 function with condition number 10 (\href{http://coco.lri.fr/downloads/download15.03/bbobdocfunctions.pdf\#page=85}{\(f_{17}\) in the \sphinxcode{bbob} suite}).

In addition to the simple sphere function, an asymmetric, non-separable,
and highly multimodal function needs to be solved to approach the Pareto
front/Pareto set where the frequency and amplitude of the modulation
in the second objective vary. The conditioning of the second objective
and thus the entire bi-objective function is low.

Contained in the \sphinxstyleemphasis{separable - multi-modal} function group.

% #1: function string (2 digits)
% #2: dimension string (2 digits)
% #3: instance string (2 digits)
% #4: function number
% #5: dimension
\bbobfigs{08}{05}{01}{8}{5}

\subsubsection[\texorpdfstring{\protect\(F_9\protect\): Sphere/Schwefel x*sin(x)}{F9: Sphere/Schwefel x*sin(x)}]{\texorpdfstring{\protect\(F_9\protect\): Sphere/Schwefel x*sin(x)}{}}
\label{index:f9}\label{index:sphere-schwefel-x-sin-x}
Combination of the sphere function (\href{http://coco.lri.fr/downloads/download15.03/bbobdocfunctions.pdf\#page=5}{\(f_1\) in the \sphinxcode{bbob} suite})
and the Schwefel function (\href{http://coco.lri.fr/downloads/download15.03/bbobdocfunctions.pdf\#page=100}{\(f_{20}\) in the \sphinxcode{bbob} suite}).

While the first objective function is separable and unimodal,
the second objective function is partially separable and highly
multimodal---having the most prominent \(2^n\) minima located
comparatively close to the corners of the unpenalized search area.

Contained in the \sphinxstyleemphasis{separable - weakly-structured} function group.

% #1: function string (2 digits)
% #2: dimension string (2 digits)
% #3: instance string (2 digits)
% #4: function number
% #5: dimension
\bbobfigs{09}{05}{01}{9}{5}

\subsubsection[\texorpdfstring{\protect\(F_{10}\protect\): Sphere/Gallagher 101 peaks}{F10: Sphere/Gallagher 101 peaks}]{\texorpdfstring{\protect\(F_{10}\protect\): Sphere/Gallagher 101 peaks}{}}
\label{index:f10}\label{index:sphere-gallagher-101-peaks}
Combination of the sphere function (\href{http://coco.lri.fr/downloads/download15.03/bbobdocfunctions.pdf\#page=5}{\(f_1\) in the \sphinxcode{bbob} suite})
and the Gallagher function with 101 peaks (\href{http://coco.lri.fr/downloads/download15.03/bbobdocfunctions.pdf\#page=105}{\(f_{21}\) in the \sphinxcode{bbob} suite}).

While the first objective function is separable and unimodal,
the second objective function is non-separable and consists
of 101 optima with position and height being unrelated and
randomly chosen (different for each instantiation of the function).
The conditioning around the global optimum of the second
objective function is about 30.

Contained in the \sphinxstyleemphasis{separable - weakly-structured} function group.

% #1: function string (2 digits)
% #2: dimension string (2 digits)
% #3: instance string (2 digits)
% #4: function number
% #5: dimension
\bbobfigs{10}{05}{01}{10}{5}

\subsubsection[\texorpdfstring{\protect\(F_{11}\protect\): Ellipsoid separable/Ellipsoid separable}{F11: Ellipsoid separable/Ellipsoid separable}]{\texorpdfstring{\protect\(F_{11}\protect\): Ellipsoid separable/Ellipsoid separable}{}}
\label{index:ellipsoid-separable-ellipsoid-separable}\label{index:f11}
Combination of two separable ellipsoid functions (\href{http://coco.lri.fr/downloads/download15.03/bbobdocfunctions.pdf\#page=10}{\(f_2\) in the \sphinxcode{bbob} suite}).

Both objectives are unimodal, separable, only globally
quadratic with smooth local irregularities, and highly
ill-conditioned with a condition number of
about \(10^6\).

Contained in the \sphinxstyleemphasis{separable - separable} function group.

% #1: function string (2 digits)
% #2: dimension string (2 digits)
% #3: instance string (2 digits)
% #4: function number
% #5: dimension
\bbobfigs{11}{05}{01}{11}{5}

\subsubsection[\texorpdfstring{\protect\(F_{12}\protect\): Ellipsoid separable/Attractive sector}{F12: Ellipsoid separable/Attractive sector}]{\texorpdfstring{\protect\(F_{12}\protect\): Ellipsoid separable/Attractive sector}{}}
\label{index:f12}\label{index:ellipsoid-separable-attractive-sector}
Combination of the separable ellipsoid function (\href{http://coco.lri.fr/downloads/download15.03/bbobdocfunctions.pdf\#page=10}{\(f_2\) in the \sphinxcode{bbob} suite})
and the attractive sector function (\href{http://coco.lri.fr/downloads/download15.03/bbobdocfunctions.pdf\#page=30}{\(f_6\) in the \sphinxcode{bbob} suite}).

Both objective functions are unimodal but only the first
one is separable. The first objective function, in addition,
is globally quadratic with smooth local irregularities, and
highly ill-conditioned with a condition number of about
\(10^6\). The second objective function is highly
asymmetric, where only one \sphinxstyleemphasis{hypercone} (with
angular base area) with a volume of roughly \((1/2)^n\)
yields low function values. The optimum of it is located at
the tip of this cone.

Contained in the \sphinxstyleemphasis{separable - moderate} function group.

% #1: function string (2 digits)
% #2: dimension string (2 digits)
% #3: instance string (2 digits)
% #4: function number
% #5: dimension
\bbobfigs{12}{05}{01}{12}{5}

\subsubsection[\texorpdfstring{\protect\(F_{13}\protect\): Ellipsoid separable/Rosenbrock original}{F13: Ellipsoid separable/Rosenbrock original}]{\texorpdfstring{\protect\(F_{13}\protect\): Ellipsoid separable/Rosenbrock original}{}}
\label{index:f13}\label{index:ellipsoid-separable-rosenbrock-original}
Combination of the separable ellipsoid function (\href{http://coco.lri.fr/downloads/download15.03/bbobdocfunctions.pdf\#page=10}{\(f_2\) in the \sphinxcode{bbob} suite}) and the original, i.e., unrotated Rosenbrock function
(\href{http://coco.lri.fr/downloads/download15.03/bbobdocfunctions.pdf\#page=40}{\(f_8\) in the \sphinxcode{bbob} suite}).

Only the first objective is separable and unimodal. The second
objective is partially separable (tri-band structure) and has a local
optimum with an attraction volume of about 25\%.
In addition, the first objective function shows smooth local
irregularities from a globally convex quadratic function and is
highly ill-conditioned with a condition number of about
\(10^6\).

Contained in the \sphinxstyleemphasis{separable - moderate} function group.

% #1: function string (2 digits)
% #2: dimension string (2 digits)
% #3: instance string (2 digits)
% #4: function number
% #5: dimension
\bbobfigs{13}{05}{01}{13}{5}

\subsubsection[\texorpdfstring{\protect\(F_{14}\protect\): Ellipsoid separable/Sharp ridge}{F14: Ellipsoid separable/Sharp ridge}]{\texorpdfstring{\protect\(F_{14}\protect\): Ellipsoid separable/Sharp ridge}{}}
\label{index:f14}\label{index:ellipsoid-separable-sharp-ridge}
Combination of the separable ellipsoid function (\href{http://coco.lri.fr/downloads/download15.03/bbobdocfunctions.pdf\#page=10}{\(f_2\) in the \sphinxcode{bbob} suite}) and the sharp ridge function (\href{http://coco.lri.fr/downloads/download15.03/bbobdocfunctions.pdf\#page=65}{\(f_{13}\) in the \sphinxcode{bbob} suite}).

Both objective functions are unimodal but only the first one is
separable.

The first objective is globally quadratic but with smooth local
irregularities and highly ill-conditioned with a condition number of
about \(10^6\). For optimizing the second objective, a sharp,
i.e., non-differentiable ridge has to be followed.

Contained in the \sphinxstyleemphasis{separable - ill-conditioned} function group.

% #1: function string (2 digits)
% #2: dimension string (2 digits)
% #3: instance string (2 digits)
% #4: function number
% #5: dimension
\bbobfigs{14}{05}{01}{14}{5}

\subsubsection[\texorpdfstring{\protect\(F_{15}\protect\): Ellipsoid separable/Sum of Different Powers}{F15: Ellipsoid separable/Sum of Different Powers}]{\texorpdfstring{\protect\(F_{15}\protect\): Ellipsoid separable/Sum of Different Powers}{}}
\label{index:ellipsoid-separable-sum-of-different-powers}\label{index:f15}
Combination of the separable ellipsoid function (\href{http://coco.lri.fr/downloads/download15.03/bbobdocfunctions.pdf\#page=10}{\(f_2\) in the \sphinxcode{bbob} suite}) and the sum of different powers function
(\href{http://coco.lri.fr/downloads/download15.03/bbobdocfunctions.pdf\#page=70}{\(f_{14}\) in the \sphinxcode{bbob} suite}).

Both objective functions are unimodal but only the first one is
separable.

The first objective is globally quadratic but with smooth local
irregularities and highly ill-conditioned with a condition number of
about \(10^6\). When approaching the second objective's optimum,
the sensitivities of the variables in the rotated search space become
more and more different.

Contained in the \sphinxstyleemphasis{separable - ill-conditioned} function group.

% #1: function string (2 digits)
% #2: dimension string (2 digits)
% #3: instance string (2 digits)
% #4: function number
% #5: dimension
\bbobfigs{15}{05}{01}{15}{5}

\subsubsection[\texorpdfstring{\protect\(F_{16}\protect\): Ellipsoid separable/Rastrigin}{F16: Ellipsoid separable/Rastrigin}]{\texorpdfstring{\protect\(F_{16}\protect\): Ellipsoid separable/Rastrigin}{}}
\label{index:ellipsoid-separable-rastrigin}\label{index:f16}
Combination of the separable ellipsoid function (\href{http://coco.lri.fr/downloads/download15.03/bbobdocfunctions.pdf\#page=10}{\(f_2\) in the \sphinxcode{bbob} suite}) and the Rastrigin function (\href{http://coco.lri.fr/downloads/download15.03/bbobdocfunctions.pdf\#page=75}{\(f_{15}\) in the \sphinxcode{bbob} suite}).

The objective functions show rather opposite properties.
The first one is separable, the second not. The first one
is unimodal, the second highly multimodal (roughly \(10^n\) local
optima). The first one is highly ill-conditioning (condition number of
\(10^6\)), the second one has a conditioning of about 10. Local
non-linear transformations are performed in both objective functions
to alleviate the original symmetry and regularity of the two
baseline functions.

Contained in the \sphinxstyleemphasis{separable - multi-modal} function group.

% #1: function string (2 digits)
% #2: dimension string (2 digits)
% #3: instance string (2 digits)
% #4: function number
% #5: dimension
\bbobfigs{16}{05}{01}{16}{5}

\subsubsection[\texorpdfstring{\protect\(F_{17}\protect\): Ellipsoid separable/Schaffer F7, condition 10}{F17: Ellipsoid separable/Schaffer F7, condition 10}]{\texorpdfstring{\protect\(F_{17}\protect\): Ellipsoid separable/Schaffer F7, condition 10}{}}
\label{index:ellipsoid-separable-schaffer-f7-condition-10}\label{index:f17}
Combination of the separable ellipsoid function (\href{http://coco.lri.fr/downloads/download15.03/bbobdocfunctions.pdf\#page=10}{\(f_2\) in the \sphinxcode{bbob} suite}) and the Schaffer F7 function with condition number 10
(\href{http://coco.lri.fr/downloads/download15.03/bbobdocfunctions.pdf\#page=85}{\(f_{17}\) in the \sphinxcode{bbob} suite}).

Also here, both single objectives possess opposing properties.
The first objective is unimodal, besides small local non-linearities symmetric,
separable and highly ill-conditioned while the second objective is highly
multi-modal, asymmetric, and non-separable, with only a low conditioning.

Contained in the \sphinxstyleemphasis{separable - multi-modal} function group.

% #1: function string (2 digits)
% #2: dimension string (2 digits)
% #3: instance string (2 digits)
% #4: function number
% #5: dimension
\bbobfigs{17}{05}{01}{17}{5}

\subsubsection[\texorpdfstring{\protect\(F_{18}\protect\): Ellipsoid separable/Schwefel x*sin(x)}{F18: Ellipsoid separable/Schwefel x*sin(x)}]{\texorpdfstring{\protect\(F_{18}\protect\): Ellipsoid separable/Schwefel x*sin(x)}{}}
\label{index:ellipsoid-separable-schwefel-x-sin-x}\label{index:f18}
Combination of the separable ellipsoid function (\href{http://coco.lri.fr/downloads/download15.03/bbobdocfunctions.pdf\#page=10}{\(f_2\) in the \sphinxcode{bbob} suite}) and the Schwefel function (\href{http://coco.lri.fr/downloads/download15.03/bbobdocfunctions.pdf\#page=100}{\(f_{20}\) in the \sphinxcode{bbob} suite}).

The first objective is unimodal, separable and highly ill-conditioned.
The second objective is partially separable and highly multimodal---having
the most prominent \(2^n\) minima located comparatively close to the
corners of the unpenalized search area.

Contained in the \sphinxstyleemphasis{separable - weakly-structured} function group.

% #1: function string (2 digits)
% #2: dimension string (2 digits)
% #3: instance string (2 digits)
% #4: function number
% #5: dimension
\bbobfigs{18}{05}{01}{18}{5}

\subsubsection[\texorpdfstring{\protect\(F_{19}\protect\): Ellipsoid separable/Gallagher 101 peaks}{F19: Ellipsoid separable/Gallagher 101 peaks}]{\texorpdfstring{\protect\(F_{19}\protect\): Ellipsoid separable/Gallagher 101 peaks}{}}
\label{index:ellipsoid-separable-gallagher-101-peaks}\label{index:f19}
Combination of the separable ellipsoid function (\href{http://coco.lri.fr/downloads/download15.03/bbobdocfunctions.pdf\#page=10}{\(f_2\) in the \sphinxcode{bbob} suite}) and the Gallagher function with 101 peaks (\href{http://coco.lri.fr/downloads/download15.03/bbobdocfunctions.pdf\#page=105}{\(f_{21}\) in the \sphinxcode{bbob} suite}).

While the first objective function is separable, unimodal, and
highly ill-conditioned (condition number of about \(10^6\)),
the second objective function is non-separable and consists
of 101 optima with position and height being unrelated and
randomly chosen (different for each instantiation of the function).
The conditioning around the global optimum of the second
objective function is about 30.

Contained in the \sphinxstyleemphasis{separable - weakly-structured} function group.

% #1: function string (2 digits)
% #2: dimension string (2 digits)
% #3: instance string (2 digits)
% #4: function number
% #5: dimension
\bbobfigs{19}{05}{01}{19}{5}

\subsubsection[\texorpdfstring{\protect\(F_{20}\protect\): Attractive sector/Attractive sector}{F20: Attractive sector/Attractive sector}]{\texorpdfstring{\protect\(F_{20}\protect\): Attractive sector/Attractive sector}{}}
\label{index:f20}\label{index:attractive-sector-attractive-sector}
Combination of two attractive sector functions (\href{http://coco.lri.fr/downloads/download15.03/bbobdocfunctions.pdf\#page=30}{\(f_6\) in the \sphinxcode{bbob} suite}).
Both functions are unimodal and highly asymmetric, where only one
\sphinxstyleemphasis{hypercone} (with angular base area) per objective with a volume of
roughly \((1/2)^n\) yields low function values. The objective
functions' optima are located at the tips of those two cones.

Contained in the \sphinxstyleemphasis{moderate - moderate} function group.

% #1: function string (2 digits)
% #2: dimension string (2 digits)
% #3: instance string (2 digits)
% #4: function number
% #5: dimension
\bbobfigs{20}{05}{01}{20}{5}

\subsubsection[\texorpdfstring{\protect\(F_{21}\protect\): Attractive sector/Rosenbrock original}{F21: Attractive sector/Rosenbrock original}]{\texorpdfstring{\protect\(F_{21}\protect\): Attractive sector/Rosenbrock original}{}}
\label{index:f21}\label{index:attractive-sector-rosenbrock-original}
Combination of the attractive sector function (\href{http://coco.lri.fr/downloads/download15.03/bbobdocfunctions.pdf\#page=30}{\(f_6\) in the \sphinxcode{bbob} suite}) and the Rosenbrock function (\href{http://coco.lri.fr/downloads/download15.03/bbobdocfunctions.pdf\#page=40}{\(f_8\) in the \sphinxcode{bbob} suite}).

The first function is unimodal but highly asymmetric, where only one
\sphinxstyleemphasis{hypercone} (with angular base area) with a volume of
roughly \((1/2)^n\) yields low function values (with the
optimum at the tip of the cone). The second
objective is partially separable (tri-band structure) and has a local
optimum with an attraction volume of about 25\%.

Contained in the \sphinxstyleemphasis{moderate - moderate} function group.

% #1: function string (2 digits)
% #2: dimension string (2 digits)
% #3: instance string (2 digits)
% #4: function number
% #5: dimension
\bbobfigs{21}{05}{01}{21}{5}

\subsubsection[\texorpdfstring{\protect\(F_{22}\protect\): Attractive sector/Sharp ridge}{F22: Attractive sector/Sharp ridge}]{\texorpdfstring{\protect\(F_{22}\protect\): Attractive sector/Sharp ridge}{}}
\label{index:f22}\label{index:attractive-sector-sharp-ridge}
Combination of the attractive sector function (\href{http://coco.lri.fr/downloads/download15.03/bbobdocfunctions.pdf\#page=30}{\(f_6\) in the \sphinxcode{bbob} suite}) and the sharp ridge function (\href{http://coco.lri.fr/downloads/download15.03/bbobdocfunctions.pdf\#page=65}{\(f_{13}\) in the \sphinxcode{bbob} suite}).

Both objective functions are unimodal and non-separable. The
first objective is highly asymmetric in the sense that only one
\sphinxstyleemphasis{hypercone} (with angular base area) with a volume of
roughly \((1/2)^n\) yields low function values (with the
optimum at the tip of the cone). For optimizing the second
objective, a sharp, i.e., non-differentiable ridge has to be followed.

Contained in the \sphinxstyleemphasis{moderate - ill-conditioned} function group.

% #1: function string (2 digits)
% #2: dimension string (2 digits)
% #3: instance string (2 digits)
% #4: function number
% #5: dimension
\bbobfigs{22}{05}{01}{22}{5}

\subsubsection[\texorpdfstring{\protect\(F_{23}\protect\): Attractive sector/Sum of Different Powers}{F23: Attractive sector/Sum of Different Powers}]{\texorpdfstring{\protect\(F_{23}\protect\): Attractive sector/Sum of Different Powers}{}}
\label{index:f23}\label{index:attractive-sector-sum-of-different-powers}
Combination of the attractive sector function (\href{http://coco.lri.fr/downloads/download15.03/bbobdocfunctions.pdf\#page=30}{\(f_6\) in the \sphinxcode{bbob} suite}) and the sum of different powers function
(\href{http://coco.lri.fr/downloads/download15.03/bbobdocfunctions.pdf\#page=70}{\(f_{14}\) in the \sphinxcode{bbob} suite}).

Both objective functions are unimodal and non-separable. The
first objective is highly asymmetric in the sense that only one
\sphinxstyleemphasis{hypercone} (with angular base area) with a volume of
roughly \((1/2)^n\) yields low function values (with the
optimum at the tip of the cone). When approaching the second
objective's optimum, the sensitivities of the variables in the
rotated search space become more and more different.

Contained in the \sphinxstyleemphasis{moderate - ill-conditioned} function group.

% #1: function string (2 digits)
% #2: dimension string (2 digits)
% #3: instance string (2 digits)
% #4: function number
% #5: dimension
\bbobfigs{23}{05}{01}{23}{5}

\subsubsection[\texorpdfstring{\protect\(F_{24}\protect\): Attractive sector/Rastrigin}{F24: Attractive sector/Rastrigin}]{\texorpdfstring{\protect\(F_{24}\protect\): Attractive sector/Rastrigin}{}}
\label{index:attractive-sector-rastrigin}\label{index:f24}
Combination of the attractive sector function (\href{http://coco.lri.fr/downloads/download15.03/bbobdocfunctions.pdf\#page=30}{\(f_6\) in the \sphinxcode{bbob} suite}) and the Rastrigin function
(\href{http://coco.lri.fr/downloads/download15.03/bbobdocfunctions.pdf\#page=75}{\(f_{15}\) in the \sphinxcode{bbob} suite}).

Both objectives are non-separable, and the second one
is highly multi-modal (roughly \(10^n\) local
optima) while the first one is unimodal. Further
properties are that the first objective is highly
asymmetric and the second has a conditioning of about 10.

Contained in the \sphinxstyleemphasis{moderate - multi-modal} function group.

% #1: function string (2 digits)
% #2: dimension string (2 digits)
% #3: instance string (2 digits)
% #4: function number
% #5: dimension
\bbobfigs{24}{05}{01}{24}{5}

\subsubsection[\texorpdfstring{\protect\(F_{25}\protect\): Attractive sector/Schaffer F7, condition 10}{F25: Attractive sector/Schaffer F7, condition 10}]{\texorpdfstring{\protect\(F_{25}\protect\): Attractive sector/Schaffer F7, condition 10}{}}
\label{index:attractive-sector-schaffer-f7-condition-10}\label{index:f25}
Combination of the attractive sector function (\href{http://coco.lri.fr/downloads/download15.03/bbobdocfunctions.pdf\#page=30}{\(f_6\) in the \sphinxcode{bbob} suite}) and the Schaffer F7 function with condition number 10
(\href{http://coco.lri.fr/downloads/download15.03/bbobdocfunctions.pdf\#page=85}{\(f_{17}\) in the \sphinxcode{bbob} suite}).

Both objectives are non-separable and asymmetric.
While the first objective is unimodal, the second one is
a highly multi-modal function with a low conditioning where
frequency and amplitude of the modulation vary.

Contained in the \sphinxstyleemphasis{moderate - multi-modal} function group.

% #1: function string (2 digits)
% #2: dimension string (2 digits)
% #3: instance string (2 digits)
% #4: function number
% #5: dimension
\bbobfigs{25}{05}{01}{25}{5}

\subsubsection[\texorpdfstring{\protect\(F_{26}\protect\): Attractive sector/Schwefel x*sin(x)}{F26: Attractive sector/Schwefel x*sin(x)}]{\texorpdfstring{\protect\(F_{26}\protect\): Attractive sector/Schwefel x*sin(x)}{}}
\label{index:f26}\label{index:attractive-sector-schwefel-x-sin-x}
Combination of the attractive sector function (\href{http://coco.lri.fr/downloads/download15.03/bbobdocfunctions.pdf\#page=30}{\(f_6\) in the \sphinxcode{bbob} suite}) and the Schwefel function (\href{http://coco.lri.fr/downloads/download15.03/bbobdocfunctions.pdf\#page=100}{\(f_{20}\) in the \sphinxcode{bbob} suite}).

The first objective is non-separable, unimodal, and asymmetric.
The second objective is partially separable and highly multimodal---having
the most prominent \(2^n\) minima located comparatively close to the
corners of the unpenalized search area.

Contained in the \sphinxstyleemphasis{moderate - weakly-structured} function group.

% #1: function string (2 digits)
% #2: dimension string (2 digits)
% #3: instance string (2 digits)
% #4: function number
% #5: dimension
\bbobfigs{26}{05}{01}{26}{5}

\subsubsection[\texorpdfstring{\protect\(F_{27}\protect\): Attractive sector/Gallagher 101 peaks}{F27: Attractive sector/Gallagher 101 peaks}]{\texorpdfstring{\protect\(F_{27}\protect\): Attractive sector/Gallagher 101 peaks}{}}
\label{index:f27}\label{index:attractive-sector-gallagher-101-peaks}
Combination of the attractive sector function (\href{http://coco.lri.fr/downloads/download15.03/bbobdocfunctions.pdf\#page=30}{\(f_6\) in the \sphinxcode{bbob} suite}) and the Gallagher function with 101 peaks (\href{http://coco.lri.fr/downloads/download15.03/bbobdocfunctions.pdf\#page=105}{\(f_{21}\) in the \sphinxcode{bbob} suite}).

Both objective functions are non-separable but only the first
is unimodal. The first objective function is furthermore asymmetric.
The second objective function has 101 optima with position and height
being unrelated and randomly chosen (different for each instantiation
of the function). The conditioning around the global optimum of the second
objective function is about 30.

Contained in the \sphinxstyleemphasis{moderate - weakly-structured} function group.

% #1: function string (2 digits)
% #2: dimension string (2 digits)
% #3: instance string (2 digits)
% #4: function number
% #5: dimension
\bbobfigs{27}{05}{01}{27}{5}

\subsubsection[\texorpdfstring{\protect\(F_{28}\protect\): Rosenbrock original/Rosenbrock original}{F28: Rosenbrock original/Rosenbrock original}]{\texorpdfstring{\protect\(F_{28}\protect\): Rosenbrock original/Rosenbrock original}{}}
\label{index:rosenbrock-original-rosenbrock-original}\label{index:f28}
Combination of two Rosenbrock functions (\href{http://coco.lri.fr/downloads/download15.03/bbobdocfunctions.pdf\#page=40}{\(f_8\) in the \sphinxcode{bbob} suite}).

Both objectives are partially separable (tri-band structure) and have
a local optimum with an attraction volume of about 25\%.

Contained in the \sphinxstyleemphasis{moderate - moderate} function group.

% #1: function string (2 digits)
% #2: dimension string (2 digits)
% #3: instance string (2 digits)
% #4: function number
% #5: dimension
\bbobfigs{28}{05}{01}{28}{5}

\subsubsection[\texorpdfstring{\protect\(F_{29}\protect\): Rosenbrock original/Sharp ridge}{F29: Rosenbrock original/Sharp ridge}]{\texorpdfstring{\protect\(F_{29}\protect\): Rosenbrock original/Sharp ridge}{}}
\label{index:rosenbrock-original-sharp-ridge}\label{index:f29}
Combination of the Rosenbrock function (\href{http://coco.lri.fr/downloads/download15.03/bbobdocfunctions.pdf\#page=40}{\(f_8\) in the \sphinxcode{bbob} suite}) and the
sharp ridge function (\href{http://coco.lri.fr/downloads/download15.03/bbobdocfunctions.pdf\#page=65}{\(f_{13}\) in the \sphinxcode{bbob} suite}).

The first objective function is partially separable (tri-band structure)
and has a local optimum with an attraction volume of about 25\%.
The second objective is unimodal and non-separable and, for
optimizing it, a sharp, i.e., non-differentiable ridge has to be followed.

Contained in the \sphinxstyleemphasis{moderate - ill-conditioned} function group.

% #1: function string (2 digits)
% #2: dimension string (2 digits)
% #3: instance string (2 digits)
% #4: function number
% #5: dimension
\bbobfigs{29}{05}{01}{29}{5}

\subsubsection[\texorpdfstring{\protect\(F_{30}\protect\): Rosenbrock original/Sum of Different Powers}{F30: Rosenbrock original/Sum of Different Powers}]{\texorpdfstring{\protect\(F_{30}\protect\): Rosenbrock original/Sum of Different Powers}{}}
\label{index:f30}\label{index:rosenbrock-original-sum-of-different-powers}
Combination of the Rosenbrock function (\href{http://coco.lri.fr/downloads/download15.03/bbobdocfunctions.pdf\#page=40}{\(f_8\) in the \sphinxcode{bbob} suite}) and the sum of different powers function
(\href{http://coco.lri.fr/downloads/download15.03/bbobdocfunctions.pdf\#page=70}{\(f_{14}\) in the \sphinxcode{bbob} suite}).

The first objective function is partially separable (tri-band structure)
and has a local optimum with an attraction volume of about 25\%.
The second objective function is unimodal and non-separable. When
approaching the second objective's optimum, the sensitivities of the
variables in the rotated search space become more and more different.

Contained in the \sphinxstyleemphasis{moderate - ill-conditioned} function group.

% #1: function string (2 digits)
% #2: dimension string (2 digits)
% #3: instance string (2 digits)
% #4: function number
% #5: dimension
\bbobfigs{30}{05}{01}{30}{5}

\subsubsection[\texorpdfstring{\protect\(F_{31}\protect\): Rosenbrock original/Rastrigin}{F31: Rosenbrock original/Rastrigin}]{\texorpdfstring{\protect\(F_{31}\protect\): Rosenbrock original/Rastrigin}{}}
\label{index:f31}\label{index:rosenbrock-original-rastrigin}
Combination of the Rosenbrock function (\href{http://coco.lri.fr/downloads/download15.03/bbobdocfunctions.pdf\#page=40}{\(f_8\) in the \sphinxcode{bbob} suite}) and the Rastrigin function
(\href{http://coco.lri.fr/downloads/download15.03/bbobdocfunctions.pdf\#page=75}{\(f_{15}\) in the \sphinxcode{bbob} suite}).

The first objective function is partially separable (tri-band structure)
and has a local optimum with an attraction volume of about 25\%.
The second objective function is non-separable and
highly multi-modal (roughly \(10^n\) local
optima).

Contained in the \sphinxstyleemphasis{moderate - multi-modal} function group.

% #1: function string (2 digits)
% #2: dimension string (2 digits)
% #3: instance string (2 digits)
% #4: function number
% #5: dimension
\bbobfigs{31}{05}{01}{31}{5}

\subsubsection[\texorpdfstring{\protect\(F_{32}\protect\): Rosenbrock original/Schaffer F7, condition 10}{F32: Rosenbrock original/Schaffer F7, condition 10}]{\texorpdfstring{\protect\(F_{32}\protect\): Rosenbrock original/Schaffer F7, condition 10}{}}
\label{index:f32}\label{index:rosenbrock-original-schaffer-f7-condition-10}
Combination of the Rosenbrock function (\href{http://coco.lri.fr/downloads/download15.03/bbobdocfunctions.pdf\#page=40}{\(f_8\) in the \sphinxcode{bbob} suite}) and the
Schaffer F7 function with condition number 10
(\href{http://coco.lri.fr/downloads/download15.03/bbobdocfunctions.pdf\#page=85}{\(f_{17}\) in the \sphinxcode{bbob} suite}).

The first objective function is partially separable (tri-band structure)
and has a local optimum with an attraction volume of about 25\%.
The second objective function is non-separable, asymmetric, and
highly multi-modal with a low conditioning where
frequency and amplitude of the modulation vary.

Contained in the \sphinxstyleemphasis{moderate - multi-modal} function group.

% #1: function string (2 digits)
% #2: dimension string (2 digits)
% #3: instance string (2 digits)
% #4: function number
% #5: dimension
\bbobfigs{32}{05}{01}{32}{5}

\subsubsection[\texorpdfstring{\protect\(F_{33}\protect\): Rosenbrock original/Schwefel x*sin(x)}{F33: Rosenbrock original/Schwefel x*sin(x)}]{\texorpdfstring{\protect\(F_{33}\protect\): Rosenbrock original/Schwefel x*sin(x)}{}}
\label{index:rosenbrock-original-schwefel-x-sin-x}\label{index:f33}
Combination of the Rosenbrock function (\href{http://coco.lri.fr/downloads/download15.03/bbobdocfunctions.pdf\#page=40}{\(f_8\) in the \sphinxcode{bbob} suite}) and the
Schwefel function (\href{http://coco.lri.fr/downloads/download15.03/bbobdocfunctions.pdf\#page=100}{\(f_{20}\) in the \sphinxcode{bbob} suite}).

Both objective functions are partially separable.
While the first objective function has a local optimum with an attraction
volume of about 25\%, the second objective function is highly
multimodal---having the most prominent \(2^n\) minima located
comparatively close to the corners of its unpenalized search area.

Contained in the \sphinxstyleemphasis{moderate - weakly-structured} function group.

% #1: function string (2 digits)
% #2: dimension string (2 digits)
% #3: instance string (2 digits)
% #4: function number
% #5: dimension
\bbobfigs{33}{05}{01}{33}{5}

\subsubsection[\texorpdfstring{\protect\(F_{34}\protect\): Rosenbrock original/Gallagher 101 peaks}{F34: Rosenbrock original/Gallagher 101 peaks}]{\texorpdfstring{\protect\(F_{34}\protect\): Rosenbrock original/Gallagher 101 peaks}{}}
\label{index:f34}\label{index:rosenbrock-original-gallagher-101-peaks}
Combination of the Rosenbrock function (\href{http://coco.lri.fr/downloads/download15.03/bbobdocfunctions.pdf\#page=40}{\(f_8\) in the \sphinxcode{bbob} suite}) and
the Gallagher function with 101 peaks (\href{http://coco.lri.fr/downloads/download15.03/bbobdocfunctions.pdf\#page=105}{\(f_{21}\) in the \sphinxcode{bbob} suite}).

The first objective function is partially separable, the second one
non-separable. While the first objective function has a local optimum
with an attraction volume of about 25\%, the second objective function
has 101 optima with position and height being unrelated and randomly
chosen (different for each instantiation of the function). The
conditioning around the global optimum of the second objective function
is about 30.

Contained in the \sphinxstyleemphasis{moderate - weakly-structured} function group.

% #1: function string (2 digits)
% #2: dimension string (2 digits)
% #3: instance string (2 digits)
% #4: function number
% #5: dimension
\bbobfigs{34}{05}{01}{34}{5}

\subsubsection[\texorpdfstring{\protect\(F_{35}\protect\): Sharp ridge/Sharp ridge}{F35: Sharp ridge/Sharp ridge}]{\texorpdfstring{\protect\(F_{35}\protect\): Sharp ridge/Sharp ridge}{}}
\label{index:sharp-ridge-sharp-ridge}\label{index:f35}
Combination of two sharp ridge functions (\href{http://coco.lri.fr/downloads/download15.03/bbobdocfunctions.pdf\#page=65}{\(f_{13}\) in the \sphinxcode{bbob} suite}).

Both objective functions are unimodal and non-separable and, for
optimizing them, two sharp, i.e., non-differentiable ridges have to be
followed.

Contained in the \sphinxstyleemphasis{ill-conditioned - ill-conditioned} function group.

% #1: function string (2 digits)
% #2: dimension string (2 digits)
% #3: instance string (2 digits)
% #4: function number
% #5: dimension
\bbobfigs{35}{05}{01}{35}{5}

\subsubsection[\texorpdfstring{\protect\(F_{36}\protect\): Sharp ridge/Sum of Different Powers}{F36: Sharp ridge/Sum of Different Powers}]{\texorpdfstring{\protect\(F_{36}\protect\): Sharp ridge/Sum of Different Powers}{}}
\label{index:sharp-ridge-sum-of-different-powers}\label{index:f36}
Combination of the sharp ridge function (\href{http://coco.lri.fr/downloads/download15.03/bbobdocfunctions.pdf\#page=65}{\(f_{13}\) in the \sphinxcode{bbob} suite}) and the
sum of different powers function
(\href{http://coco.lri.fr/downloads/download15.03/bbobdocfunctions.pdf\#page=70}{\(f_{14}\) in the \sphinxcode{bbob} suite}).

Both functions are uni-modal and non-separable.
For optimizing the first objective, a sharp, i.e., non-differentiable
ridge has to be followed.
When approaching the second objective's optimum, the sensitivities of the
variables in the rotated search space become more and more different.

Contained in the \sphinxstyleemphasis{ill-conditioned - ill-conditioned} function group.

% #1: function string (2 digits)
% #2: dimension string (2 digits)
% #3: instance string (2 digits)
% #4: function number
% #5: dimension
\bbobfigs{36}{05}{01}{36}{5}

\subsubsection[\texorpdfstring{\protect\(F_{37}\protect\): Sharp ridge/Rastrigin}{F37: Sharp ridge/Rastrigin}]{\texorpdfstring{\protect\(F_{37}\protect\): Sharp ridge/Rastrigin}{}}
\label{index:sharp-ridge-rastrigin}\label{index:f37}
Combination of the sharp ridge function (\href{http://coco.lri.fr/downloads/download15.03/bbobdocfunctions.pdf\#page=65}{\(f_{13}\) in the \sphinxcode{bbob} suite}) and the Rastrigin function
(\href{http://coco.lri.fr/downloads/download15.03/bbobdocfunctions.pdf\#page=75}{\(f_{15}\) in the \sphinxcode{bbob} suite}).

Both functions are non-separable. While the first one
is unimodal and non-differentiable at its ridge, the second objective
function is highly multi-modal (roughly \(10^n\) local optima).

Contained in the \sphinxstyleemphasis{ill-conditioned - multi-modal} function group.

% #1: function string (2 digits)
% #2: dimension string (2 digits)
% #3: instance string (2 digits)
% #4: function number
% #5: dimension
\bbobfigs{37}{05}{01}{37}{5}

\subsubsection[\texorpdfstring{\protect\(F_{38}\protect\): Sharp ridge/Schaffer F7, condition 10}{F38: Sharp ridge/Schaffer F7, condition 10}]{\texorpdfstring{\protect\(F_{38}\protect\): Sharp ridge/Schaffer F7, condition 10}{}}
\label{index:f38}\label{index:sharp-ridge-schaffer-f7-condition-10}
Combination of the sharp ridge function (\href{http://coco.lri.fr/downloads/download15.03/bbobdocfunctions.pdf\#page=65}{\(f_{13}\) in the \sphinxcode{bbob} suite}) and the
Schaffer F7 function with condition number 10
(\href{http://coco.lri.fr/downloads/download15.03/bbobdocfunctions.pdf\#page=85}{\(f_{17}\) in the \sphinxcode{bbob} suite}).

Both functions are non-separable. While the first one
is unimodal and non-differentiable at its ridge, the second objective
function is asymmetric and highly multi-modal with a low conditioning where
frequency and amplitude of the modulation vary.

Contained in the \sphinxstyleemphasis{ill-conditioned - multi-modal} function group.

% #1: function string (2 digits)
% #2: dimension string (2 digits)
% #3: instance string (2 digits)
% #4: function number
% #5: dimension
\bbobfigs{38}{05}{01}{38}{5}

\subsubsection[\texorpdfstring{\protect\(F_{39}\protect\): Sharp ridge/Schwefel x*sin(x)}{F39: Sharp ridge/Schwefel x*sin(x)}]{\texorpdfstring{\protect\(F_{39}\protect\): Sharp ridge/Schwefel x*sin(x)}{}}
\label{index:f39}\label{index:sharp-ridge-schwefel-x-sin-x}
Combination of the sharp ridge function (\href{http://coco.lri.fr/downloads/download15.03/bbobdocfunctions.pdf\#page=65}{\(f_{13}\) in the \sphinxcode{bbob} suite}) and the
Schwefel function (\href{http://coco.lri.fr/downloads/download15.03/bbobdocfunctions.pdf\#page=100}{\(f_{20}\) in the \sphinxcode{bbob} suite}).

While the first objective function is unimodal, non-separable, and
non-differentiable at its ridge, the second objective function is highly
multimodal---having the most prominent \(2^n\) minima located
comparatively close to the corners of its unpenalized search area.

Contained in the \sphinxstyleemphasis{ill-conditioned - weakly-structured} function group.

% #1: function string (2 digits)
% #2: dimension string (2 digits)
% #3: instance string (2 digits)
% #4: function number
% #5: dimension
\bbobfigs{39}{05}{01}{39}{5}

\subsubsection[\texorpdfstring{\protect\(F_{40}\protect\): Sharp ridge/Gallagher 101 peaks}{F40: Sharp ridge/Gallagher 101 peaks}]{\texorpdfstring{\protect\(F_{40}\protect\): Sharp ridge/Gallagher 101 peaks}{}}
\label{index:f40}\label{index:sharp-ridge-gallagher-101-peaks}
Combination of the sharp ridge function (\href{http://coco.lri.fr/downloads/download15.03/bbobdocfunctions.pdf\#page=65}{\(f_{13}\) in the \sphinxcode{bbob} suite}) and the
Gallagher function with 101 peaks (\href{http://coco.lri.fr/downloads/download15.03/bbobdocfunctions.pdf\#page=105}{\(f_{21}\) in the \sphinxcode{bbob} suite}).

Both objective functions are non-separable.
While the first objective function is unimodal and non-differentiable at
its ridge, the second objective function
has 101 optima with position and height being unrelated and randomly
chosen (different for each instantiation of the function). The
conditioning around the global optimum of the second objective function
is about 30.

Contained in the \sphinxstyleemphasis{ill-conditioned - weakly-structured} function group.

% #1: function string (2 digits)
% #2: dimension string (2 digits)
% #3: instance string (2 digits)
% #4: function number
% #5: dimension
\bbobfigs{40}{05}{01}{40}{5}

\subsubsection[\texorpdfstring{\protect\(F_{41}\protect\): Sum of Different Powers/Sum of Different Powers}{F41: Sum of Different Powers/Sum of Different Powers}]{\texorpdfstring{\protect\(F_{41}\protect\): Sum of Different Powers/Sum of Different Powers}{}}
\label{index:f41}\label{index:sum-of-different-powers-sum-of-different-powers}
Combination of two sum of different powers functions
(\href{http://coco.lri.fr/downloads/download15.03/bbobdocfunctions.pdf\#page=70}{\(f_{14}\) in the \sphinxcode{bbob} suite}).

Both functions are uni-modal and non-separable where the sensitivities of
the variables in the rotated search space become more and more different
when approaching the objectives' optima.

Contained in the \sphinxstyleemphasis{ill-conditioned - ill-conditioned} function group.

% #1: function string (2 digits)
% #2: dimension string (2 digits)
% #3: instance string (2 digits)
% #4: function number
% #5: dimension
\bbobfigs{41}{05}{01}{41}{5}

\subsubsection[\texorpdfstring{\protect\(F_{42}\protect\): Sum of Different Powers/Rastrigin}{F42: Sum of Different Powers/Rastrigin}]{\texorpdfstring{\protect\(F_{42}\protect\): Sum of Different Powers/Rastrigin}{}}
\label{index:f42}\label{index:sum-of-different-powers-rastrigin}
Combination of the sum of different powers functions
(\href{http://coco.lri.fr/downloads/download15.03/bbobdocfunctions.pdf\#page=70}{\(f_{14}\) in the \sphinxcode{bbob} suite}) and the Rastrigin function
(\href{http://coco.lri.fr/downloads/download15.03/bbobdocfunctions.pdf\#page=75}{\(f_{15}\) in the \sphinxcode{bbob} suite}).

Both objective functions are non-separable. While the first one
is unimodal, the second objective
function is highly multi-modal (roughly \(10^n\) local optima).

Contained in the \sphinxstyleemphasis{ill-conditioned - multi-modal} function group.

% #1: function string (2 digits)
% #2: dimension string (2 digits)
% #3: instance string (2 digits)
% #4: function number
% #5: dimension
\bbobfigs{42}{05}{01}{42}{5}

\subsubsection[\texorpdfstring{\protect\(F_{43}\protect\): Sum of Different Powers/Schaffer F7, condition 10}{F43: Sum of Different Powers/Schaffer F7, condition 10}]{\texorpdfstring{\protect\(F_{43}\protect\): Sum of Different Powers/Schaffer F7, condition 10}{}}
\label{index:sum-of-different-powers-schaffer-f7-condition-10}\label{index:f43}
Combination of the sum of different powers functions
(\href{http://coco.lri.fr/downloads/download15.03/bbobdocfunctions.pdf\#page=70}{\(f_{14}\) in the \sphinxcode{bbob} suite}) and the Schaffer F7 function with
condition number 10 (\href{http://coco.lri.fr/downloads/download15.03/bbobdocfunctions.pdf\#page=85}{\(f_{17}\) in the \sphinxcode{bbob} suite}).

Both objective functions are non-separable. While the first one
is unimodal with an increasing conditioning once the optimum is approached,
the second objective function is asymmetric and highly multi-modal with a
low conditioning where frequency and amplitude of the modulation vary.

Contained in the \sphinxstyleemphasis{ill-conditioned - multi-modal} function group.

% #1: function string (2 digits)
% #2: dimension string (2 digits)
% #3: instance string (2 digits)
% #4: function number
% #5: dimension
\bbobfigs{43}{05}{01}{43}{5}

\subsubsection[\texorpdfstring{\protect\(F_{44}\protect\): Sum of Different Powers/Schwefel x*sin(x)}{F44: Sum of Different Powers/Schwefel x*sin(x)}]{\texorpdfstring{\protect\(F_{44}\protect\): Sum of Different Powers/Schwefel x*sin(x)}{}}
\label{index:f44}\label{index:sum-of-different-powers-schwefel-x-sin-x}
Combination of the sum of different powers functions
(\href{http://coco.lri.fr/downloads/download15.03/bbobdocfunctions.pdf\#page=70}{\(f_{14}\) in the \sphinxcode{bbob} suite}) and the Schwefel function (\href{http://coco.lri.fr/downloads/download15.03/bbobdocfunctions.pdf\#page=100}{\(f_{20}\) in the \sphinxcode{bbob} suite}).

Both objectives are non-separable.
While the first objective function is unimodal,
the second objective function is highly multimodal---having the most
prominent \(2^n\) minima located comparatively close to the corners
of its unpenalized search area.

Contained in the \sphinxstyleemphasis{ill-conditioned - weakly-structured} function group.

% #1: function string (2 digits)
% #2: dimension string (2 digits)
% #3: instance string (2 digits)
% #4: function number
% #5: dimension
\bbobfigs{44}{05}{01}{44}{5}

\subsubsection[\texorpdfstring{\protect\(F_{45}\protect\): Sum of Different Powers/Gallagher 101 peaks}{F45: Sum of Different Powers/Gallagher 101 peaks}]{\texorpdfstring{\protect\(F_{45}\protect\): Sum of Different Powers/Gallagher 101 peaks}{}}
\label{index:sum-of-different-powers-gallagher-101-peaks}\label{index:f45}
Combination of the sum of different powers functions
(\href{http://coco.lri.fr/downloads/download15.03/bbobdocfunctions.pdf\#page=70}{\(f_{14}\) in the \sphinxcode{bbob} suite}) and the Gallagher function with
101 peaks (\href{http://coco.lri.fr/downloads/download15.03/bbobdocfunctions.pdf\#page=105}{\(f_{21}\) in the \sphinxcode{bbob} suite}).

Both objective functions are non-separable.
While the first objective function is unimodal, the second objective function
has 101 optima with position and height being unrelated and randomly
chosen (different for each instantiation of the function). The
conditioning around the global optimum of the second objective function
is about 30.

Contained in the \sphinxstyleemphasis{ill-conditioned - weakly-structured} function group.

% #1: function string (2 digits)
% #2: dimension string (2 digits)
% #3: instance string (2 digits)
% #4: function number
% #5: dimension
\bbobfigs{45}{05}{01}{45}{5}

\subsubsection[\texorpdfstring{\protect\(F_{46}\protect\): Rastrigin/Rastrigin}{F46: Rastrigin/Rastrigin}]{\texorpdfstring{\protect\(F_{46}\protect\): Rastrigin/Rastrigin}{}}
\label{index:rastrigin-rastrigin}\label{index:f46}
Combination of two Rastrigin functions
(\href{http://coco.lri.fr/downloads/download15.03/bbobdocfunctions.pdf\#page=75}{\(f_{15}\) in the \sphinxcode{bbob} suite}).

Both objective functions are non-separable and highly multi-modal
(roughly \(10^n\) local optima).

Contained in the \sphinxstyleemphasis{multi-modal - multi-modal} function group.

% #1: function string (2 digits)
% #2: dimension string (2 digits)
% #3: instance string (2 digits)
% #4: function number
% #5: dimension
\bbobfigs{46}{05}{01}{46}{5}

\subsubsection[\texorpdfstring{\protect\(F_{47}\protect\): Rastrigin/Schaffer F7, condition 10}{F47: Rastrigin/Schaffer F7, condition 10}]{\texorpdfstring{\protect\(F_{47}\protect\): Rastrigin/Schaffer F7, condition 10}{}}
\label{index:rastrigin-schaffer-f7-condition-10}\label{index:f47}
Combination of the Rastrigin function
(\href{http://coco.lri.fr/downloads/download15.03/bbobdocfunctions.pdf\#page=75}{\(f_{15}\) in the \sphinxcode{bbob} suite}) and the Schaffer F7 function with
condition number 10 (\href{http://coco.lri.fr/downloads/download15.03/bbobdocfunctions.pdf\#page=85}{\(f_{17}\) in the \sphinxcode{bbob} suite}).

Both objective functions are non-separable and highly multi-modal.

Contained in the \sphinxstyleemphasis{multi-modal - multi-modal} function group.

% #1: function string (2 digits)
% #2: dimension string (2 digits)
% #3: instance string (2 digits)
% #4: function number
% #5: dimension
\bbobfigs{47}{05}{01}{47}{5}

\subsubsection[\texorpdfstring{\protect\(F_{48}\protect\): Rastrigin/Schwefel x*sin(x)}{F48: Rastrigin/Schwefel x*sin(x)}]{\texorpdfstring{\protect\(F_{48}\protect\): Rastrigin/Schwefel x*sin(x)}{}}
\label{index:f48}\label{index:rastrigin-schwefel-x-sin-x}
Combination of the Rastrigin function
(\href{http://coco.lri.fr/downloads/download15.03/bbobdocfunctions.pdf\#page=75}{\(f_{15}\) in the \sphinxcode{bbob} suite}) and the Schwefel function (\href{http://coco.lri.fr/downloads/download15.03/bbobdocfunctions.pdf\#page=100}{\(f_{20}\) in the \sphinxcode{bbob} suite}).

Both objective functions are non-separable and highly multi-modal where
the first has roughly \(10^n\) local optima and the most prominent
\(2^n\) minima of the second objective function are located
comparatively close to the corners of its unpenalized search area.

Contained in the \sphinxstyleemphasis{multi-modal - weakly-structured} function group.

% #1: function string (2 digits)
% #2: dimension string (2 digits)
% #3: instance string (2 digits)
% #4: function number
% #5: dimension
\bbobfigs{48}{05}{01}{48}{5}

\subsubsection[\texorpdfstring{\protect\(F_{49}\protect\): Rastrigin/Gallagher 101 peaks}{F49: Rastrigin/Gallagher 101 peaks}]{\texorpdfstring{\protect\(F_{49}\protect\): Rastrigin/Gallagher 101 peaks}{}}
\label{index:f49}\label{index:rastrigin-gallagher-101-peaks}
Combination of the Rastrigin function
(\href{http://coco.lri.fr/downloads/download15.03/bbobdocfunctions.pdf\#page=75}{\(f_{15}\) in the \sphinxcode{bbob} suite}) and the Gallagher function with
101 peaks (\href{http://coco.lri.fr/downloads/download15.03/bbobdocfunctions.pdf\#page=105}{\(f_{21}\) in the \sphinxcode{bbob} suite}).

Both objective functions are non-separable and highly multi-modal where
the first has roughly \(10^n\) local optima and the second has
101 optima with position and height being unrelated and randomly
chosen (different for each instantiation of the function).

Contained in the \sphinxstyleemphasis{multi-modal - weakly-structured} function group.

% #1: function string (2 digits)
% #2: dimension string (2 digits)
% #3: instance string (2 digits)
% #4: function number
% #5: dimension
\bbobfigs{49}{05}{01}{49}{5}

\subsubsection[\texorpdfstring{\protect\(F_{50}\protect\): Schaffer F7, condition 10/Schaffer F7, condition 10}{F50: Schaffer F7, condition 10/Schaffer F7, condition 10}]{\texorpdfstring{\protect\(F_{50}\protect\): Schaffer F7, condition 10/Schaffer F7, condition 10}{}}
\label{index:schaffer-f7-condition-10-schaffer-f7-condition-10}\label{index:f50}
Combination of two Schaffer F7 functions with
condition number 10 (\href{http://coco.lri.fr/downloads/download15.03/bbobdocfunctions.pdf\#page=85}{\(f_{17}\) in the \sphinxcode{bbob} suite}).

Both objective functions are non-separable and highly multi-modal.

Contained in the \sphinxstyleemphasis{multi-modal - multi-modal} function group.

% #1: function string (2 digits)
% #2: dimension string (2 digits)
% #3: instance string (2 digits)
% #4: function number
% #5: dimension
\bbobfigs{50}{05}{01}{50}{5}

\subsubsection[\texorpdfstring{\protect\(F_{51}\protect\): Schaffer F7, condition 10/Schwefel x*sin(x)}{F51: Schaffer F7, condition 10/Schwefel x*sin(x)}]{\texorpdfstring{\protect\(F_{51}\protect\): Schaffer F7, condition 10/Schwefel x*sin(x)}{}}
\label{index:schaffer-f7-condition-10-schwefel-x-sin-x}\label{index:f51}
Combination of the Schaffer F7 function with
condition number 10 (\href{http://coco.lri.fr/downloads/download15.03/bbobdocfunctions.pdf\#page=85}{\(f_{17}\) in the \sphinxcode{bbob} suite})
and the Schwefel function (\href{http://coco.lri.fr/downloads/download15.03/bbobdocfunctions.pdf\#page=100}{\(f_{20}\) in the \sphinxcode{bbob} suite}).

Both objective functions are non-separable and highly multi-modal.
While frequency and amplitude of the modulation vary in an almost
regular fashion in the first objective function, the second objective
function possesses less global structure.

Contained in the \sphinxstyleemphasis{multi-modal - weakly-structured} function group.

% #1: function string (2 digits)
% #2: dimension string (2 digits)
% #3: instance string (2 digits)
% #4: function number
% #5: dimension
\bbobfigs{51}{05}{01}{51}{5}

\subsubsection[\texorpdfstring{\protect\(F_{52}\protect\): Schaffer F7, condition 10/Gallagher 101 peaks}{F52: Schaffer F7, condition 10/Gallagher 101 peaks}]{\texorpdfstring{\protect\(F_{52}\protect\): Schaffer F7, condition 10/Gallagher 101 peaks}{}}
\label{index:schaffer-f7-condition-10-gallagher-101-peaks}\label{index:f52}
Combination of the Schaffer F7 function with
condition number 10 (\href{http://coco.lri.fr/downloads/download15.03/bbobdocfunctions.pdf\#page=85}{\(f_{17}\) in the \sphinxcode{bbob} suite})
and the Gallagher function with
101 peaks (\href{http://coco.lri.fr/downloads/download15.03/bbobdocfunctions.pdf\#page=105}{\(f_{21}\) in the \sphinxcode{bbob} suite}).

Both objective functions are non-separable and highly multi-modal.
While frequency and amplitude of the modulation vary in an almost
regular fashion in the first objective function, the second has
101 optima with position and height being unrelated and randomly
chosen (different for each instantiation of the function).

Contained in the \sphinxstyleemphasis{multi-modal - weakly-structured} function group.

% #1: function string (2 digits)
% #2: dimension string (2 digits)
% #3: instance string (2 digits)
% #4: function number
% #5: dimension
\bbobfigs{52}{05}{01}{52}{5}

\subsubsection[\texorpdfstring{\protect\(F_{53}\protect\): Schwefel x*sin(x)/Schwefel x*sin(x)}{F53: Schwefel x*sin(x)/Schwefel x*sin(x)}]{\texorpdfstring{\protect\(F_{53}\protect\): Schwefel x*sin(x)/Schwefel x*sin(x)}{}}
\label{index:schwefel-x-sin-x-schwefel-x-sin-x}\label{index:f53}
Combination of two Schwefel functions (\href{http://coco.lri.fr/downloads/download15.03/bbobdocfunctions.pdf\#page=100}{\(f_{20}\) in the \sphinxcode{bbob} suite}).

Both objective functions are non-separable and highly multi-modal where
the most prominent \(2^n\) minima of each objective function are
located comparatively close to the corners of its unpenalized search area.
Due to the combinatorial nature of the Schwefel function, it is likely
in low dimensions that the Pareto set goes through the origin of the
search space.

Contained in the \sphinxstyleemphasis{weakly-structured - weakly-structured} function group.

% #1: function string (2 digits)
% #2: dimension string (2 digits)
% #3: instance string (2 digits)
% #4: function number
% #5: dimension
\bbobfigs{53}{05}{01}{53}{5}

\subsubsection[\texorpdfstring{\protect\(F_{54}\protect\): Schwefel x*sin(x)/Gallagher 101 peaks}{F54: Schwefel x*sin(x)/Gallagher 101 peaks}]{\texorpdfstring{\protect\(F_{54}\protect\): Schwefel x*sin(x)/Gallagher 101 peaks}{}}
\label{index:f54}\label{index:schwefel-x-sin-x-gallagher-101-peaks}
Combination of the Schwefel function (\href{http://coco.lri.fr/downloads/download15.03/bbobdocfunctions.pdf\#page=100}{\(f_{20}\) in the \sphinxcode{bbob} suite}) and the Gallagher function with
101 peaks (\href{http://coco.lri.fr/downloads/download15.03/bbobdocfunctions.pdf\#page=105}{\(f_{21}\) in the \sphinxcode{bbob} suite}).

Both objective functions are non-separable and highly multi-modal.
For the first objective function, the most prominent \(2^n\) minima
are located comparatively close to the corners of its unpenalized search
area. For the second objective, position and height of all
101 optima are unrelated and randomly
chosen (different for each instantiation of the function).

Contained in the \sphinxstyleemphasis{weakly-structured - weakly-structured} function group.

% #1: function string (2 digits)
% #2: dimension string (2 digits)
% #3: instance string (2 digits)
% #4: function number
% #5: dimension
\bbobfigs{54}{05}{01}{54}{5}

\subsubsection[\texorpdfstring{\protect\(F_{55}\protect\): Gallagher 101 peaks/Gallagher 101 peaks}{F55: Gallagher 101 peaks/Gallagher 101 peaks}]{\texorpdfstring{\protect\(F_{55}\protect\): Gallagher 101 peaks/Gallagher 101 peaks}{}}
\label{index:gallagher-101-peaks-gallagher-101-peaks}\label{index:f55}
Combination of two Gallagher functions with
101 peaks (\href{http://coco.lri.fr/downloads/download15.03/bbobdocfunctions.pdf\#page=105}{\(f_{21}\) in the \sphinxcode{bbob} suite}).

Both objective functions are non-separable and highly multi-modal.
Position and height of all 101 optima in each objective function
are unrelated and randomly chosen and thus, no global structure
is present.

Contained in the \sphinxstyleemphasis{weakly-structured - weakly-structured} function group.

% #1: function string (2 digits)
% #2: dimension string (2 digits)
% #3: instance string (2 digits)
% #4: function number
% #5: dimension
\bbobfigs{55}{05}{01}{55}{5}

\subsubsection[\texorpdfstring{\protect\(F_{56}\protect\): Sphere/Rastrigin separable}{F56: Sphere/Rastrigin separable}]{\texorpdfstring{\protect\(F_{56}\protect\): Sphere/Rastrigin separable}{}}
\label{index:f56}\label{index:sphere-rastrigin-separable}
Combination of the Sphere function (\href{http://coco.lri.fr/downloads/download15.03/bbobdocfunctions.pdf\#page=5}{\(f_1\) in the \sphinxcode{bbob} suite}) and the
separable Rastrigin function (\href{http://coco.lri.fr/downloads/download15.03/bbobdocfunctions.pdf\#page=15}{\(f_3\) in the \sphinxcode{bbob} suite}).

While the first objective function is unimodal, highly symmetric,
rotational and scale invariant, the second one is highly multimodal
with a comparatively regular structure for the placement of the optima.
Note that the non-linear transformations of the second objective's
Rastrigin function alleviate the symmetry and regularity of the
original Rastrigin function.

Contained in the \sphinxstyleemphasis{separable - separable} function group.

% #1: function string (2 digits)
% #2: dimension string (2 digits)
% #3: instance string (2 digits)
% #4: function number
% #5: dimension
\bbobfigs{56}{05}{01}{56}{5}

\subsubsection[\texorpdfstring{\protect\(F_{57}\protect\): Sphere/Rastrigin-B\"{u}che}{F57: Sphere/Rastrigin-B\"{u}che}]{\texorpdfstring{\protect\(F_{57}\protect\): Sphere/Rastrigin-B\"{u}che}{}}
\label{index:f57}\label{index:sphere-rastrigin-buche}
Combination of the Sphere function (\href{http://coco.lri.fr/downloads/download15.03/bbobdocfunctions.pdf\#page=5}{\(f_1\) in the \sphinxcode{bbob} suite}) and the
separable B\"{u}che-Rastrigin function (\href{http://coco.lri.fr/downloads/download15.03/bbobdocfunctions.pdf\#page=20}{\(f_4\) in the \sphinxcode{bbob} suite}).

While the first objective function is unimodal, highly symmetric,
rotational and scale invariant, the second one is highly multimodal
with a structured but highly asymmetric placement of the optima.
Constructed as a deceptive function for symmetrically distributed search
operators, the second objective function has roughly 10D
local optima, a conditioning of about 10, and a skew factor of about
10 in x-space and 100 in f-space.

Contained in the \sphinxstyleemphasis{separable - separable} function group.

% #1: function string (2 digits)
% #2: dimension string (2 digits)
% #3: instance string (2 digits)
% #4: function number
% #5: dimension
\bbobfigs{57}{05}{01}{57}{5}

\subsubsection[\texorpdfstring{\protect\(F_{58}\protect\): Sphere/Linear slope}{F58: Sphere/Linear slope}]{\texorpdfstring{\protect\(F_{58}\protect\): Sphere/Linear slope}{}}
\label{index:sphere-linear-slope}\label{index:f58}
Combination of the Sphere function (\href{http://coco.lri.fr/downloads/download15.03/bbobdocfunctions.pdf\#page=5}{\(f_1\) in the \sphinxcode{bbob} suite}) and the
Linear Slope function (\href{http://coco.lri.fr/downloads/download15.03/bbobdocfunctions.pdf\#page=25}{\(f_5\) in the \sphinxcode{bbob} suite}).

Both objective functions are separable and amongst the simplest
continuous functions to optimize. The first objective function is fully
quadratic and symmetric around the optimum, the second objective function
is fully linear within the hypercube \([-5,5]^n\) and has a region of
constant \(f\)-value outside the hypercube by definition to ensure
that a solution at one corner of \([-5,5]^n\) has optimal function value.

Contained in the \sphinxstyleemphasis{separable - separable} function group.

% #1: function string (2 digits)
% #2: dimension string (2 digits)
% #3: instance string (2 digits)
% #4: function number
% #5: dimension
\bbobfigs{58}{05}{01}{58}{5}

\subsubsection[\texorpdfstring{\protect\(F_{59}\protect\): Separable Ellipsoid/Separable Rastrigin}{F59: Separable Ellipsoid/Separable Rastrigin}]{\texorpdfstring{\protect\(F_{59}\protect\): Separable Ellipsoid/Separable Rastrigin}{}}
\label{index:separable-ellipsoid-separable-rastrigin}\label{index:f59}
Combination of the separable Ellipsoid function (\href{http://coco.lri.fr/downloads/download15.03/bbobdocfunctions.pdf\#page=10}{\(f_2\) in the \sphinxcode{bbob} suite}) and the
separable Rastrigin function (\href{http://coco.lri.fr/downloads/download15.03/bbobdocfunctions.pdf\#page=15}{\(f_3\) in the \sphinxcode{bbob} suite}).

Besides being both separable, the two objective functions are quite opposite:
the first objective function is unimodal, globally quadratic and ill-conditioned
with a conditioning of about \(10^6\) with smooth local irregularities while
the second objective function is highly multimodal with roughly \(10n\) local
optima and only small conditioning of about \(10\). Note that the separable
Rastrigin function has a comparatively regular structure for the
placement of the optima but asymmetric and oscillating non-linear transformations
of this function alleviates the symmetry and regularity of the original Rastrigin function.

Contained in the \sphinxstyleemphasis{separable - separable} function group.

% #1: function string (2 digits)
% #2: dimension string (2 digits)
% #3: instance string (2 digits)
% #4: function number
% #5: dimension
\bbobfigs{59}{05}{01}{59}{5}

\subsubsection[\texorpdfstring{\protect\(F_{60}\protect\): separable Ellipsoid/B\"{u}che-Rastrigin}{F60: separable Ellipsoid/B\"{u}che-Rastrigin}]{\texorpdfstring{\protect\(F_{60}\protect\): separable Ellipsoid/B\"{u}che-Rastrigin}{}}
\label{index:separable-ellipsoid-buche-rastrigin}\label{index:f60}
Combination of the separable Ellipsoid function (\href{http://coco.lri.fr/downloads/download15.03/bbobdocfunctions.pdf\#page=10}{\(f_2\) in the \sphinxcode{bbob} suite}) and the
separable B\"{u}che-Rastrigin function (\href{http://coco.lri.fr/downloads/download15.03/bbobdocfunctions.pdf\#page=20}{\(f_4\) in the \sphinxcode{bbob} suite}).

Besides being both separable, the two objective functions are quite opposite:
the first objective function is unimodal, globally quadratic and ill-conditioned
with a conditioning of about \(10^6\) with smooth local irregularities while
the second objective is highly multimodal with a structured but highly asymmetric
placement of the optima. Constructed as a deceptive function for symmetrically
distributed search operators.

Contained in the \sphinxstyleemphasis{separable - separable} function group.

% #1: function string (2 digits)
% #2: dimension string (2 digits)
% #3: instance string (2 digits)
% #4: function number
% #5: dimension
\bbobfigs{60}{05}{01}{60}{5}

\subsubsection[\texorpdfstring{\protect\(F_{61}\protect\): Separable Ellipsoid/Linear Slope}{F61: Separable Ellipsoid/Linear Slope}]{\texorpdfstring{\protect\(F_{61}\protect\): Separable Ellipsoid/Linear Slope}{}}
\label{index:f61}\label{index:separable-ellipsoid-linear-slope}
Combination of the separable Ellipsoid function (\href{http://coco.lri.fr/downloads/download15.03/bbobdocfunctions.pdf\#page=10}{\(f_2\) in the \sphinxcode{bbob} suite}) and the
Linear Slope function (\href{http://coco.lri.fr/downloads/download15.03/bbobdocfunctions.pdf\#page=25}{\(f_5\) in the \sphinxcode{bbob} suite}).

Both objective functions are separable. The first objective function
is unimodal with a high condition number of about \(10^6\). The second
objective function is fully linear within the hypercube \([-5,5]^n\) and
has a region of constant \(f\)-value outside the hypercube by definition to ensure
that a solution at one corner of \([-5,5]^n\) has optimal function value.

Contained in the \sphinxstyleemphasis{separable - separable} function group.

% #1: function string (2 digits)
% #2: dimension string (2 digits)
% #3: instance string (2 digits)
% #4: function number
% #5: dimension
\bbobfigs{61}{05}{01}{61}{5}

\subsubsection[\texorpdfstring{\protect\(F_{62}\protect\): separable Rastrigin/B\"{u}che-Rastrigin}{F62: separable Rastrigin/B\"{u}che-Rastrigin}]{\texorpdfstring{\protect\(F_{62}\protect\): separable Rastrigin/B\"{u}che-Rastrigin}{}}
\label{index:separable-rastrigin-buche-rastrigin}\label{index:f62}
Combination of the separable Rastrigin function (\href{http://coco.lri.fr/downloads/download15.03/bbobdocfunctions.pdf\#page=15}{\(f_3\) in the \sphinxcode{bbob} suite}) and the
separable B\"{u}che-Rastrigin function (\href{http://coco.lri.fr/downloads/download15.03/bbobdocfunctions.pdf\#page=20}{\(f_4\) in the \sphinxcode{bbob} suite}).

Both objective functions are separable and highly multimodal with an underlying
structure for the placements of the optima. While for the separable Rastrigin function,
the placements of the optima is symmetric, the optima for the B\"{u}che-Rastrigin function
are highly asymmetrically placed.

Contained in the \sphinxstyleemphasis{separable - separable} function group.

% #1: function string (2 digits)
% #2: dimension string (2 digits)
% #3: instance string (2 digits)
% #4: function number
% #5: dimension
\bbobfigs{62}{05}{01}{62}{5}

\subsubsection[\texorpdfstring{\protect\(F_{63}\protect\): Separable Rastrigin/Linear Slope}{F63: Separable Rastrigin/Linear Slope}]{\texorpdfstring{\protect\(F_{63}\protect\): Separable Rastrigin/Linear Slope}{}}
\label{index:separable-rastrigin-linear-slope}\label{index:f63}
Combination of the separable Rastrigin function (\href{http://coco.lri.fr/downloads/download15.03/bbobdocfunctions.pdf\#page=15}{\(f_3\) in the \sphinxcode{bbob} suite}) and the
Linear Slope function (\href{http://coco.lri.fr/downloads/download15.03/bbobdocfunctions.pdf\#page=25}{\(f_5\) in the \sphinxcode{bbob} suite}).

Both objective functions are separable, but while the first objective function
is highly multi-modal with an underlying symmetric structure, the second objective
function is purely linear with plateaus of constant function value outside
the region \([-5,5]^n\).

Contained in the \sphinxstyleemphasis{separable - separable} function group.

% #1: function string (2 digits)
% #2: dimension string (2 digits)
% #3: instance string (2 digits)
% #4: function number
% #5: dimension
\bbobfigs{63}{05}{01}{63}{5}

\subsubsection[\texorpdfstring{\protect\(F_{64}\protect\): B\"{u}che-Rastrigin/Linear slope}{F64: B\"{u}che-Rastrigin/Linear slope}]{\texorpdfstring{\protect\(F_{64}\protect\): B\"{u}che-Rastrigin/Linear slope}{}}
\label{index:f64}\label{index:buche-rastrigin-linear-slope}
Combination of the B\"{u}che-Rastrigin function (\href{http://coco.lri.fr/downloads/download15.03/bbobdocfunctions.pdf\#page=20}{\(f_4\) in the \sphinxcode{bbob} suite}) and the
Linear Slope function (\href{http://coco.lri.fr/downloads/download15.03/bbobdocfunctions.pdf\#page=25}{\(f_5\) in the \sphinxcode{bbob} suite}).

Both objective functions are separable, but while the first objective function
is highly multi-modal with an underlying asymmetric structure, the second objective
function is purely linear with plateaus of constant function value outside
the region \([-5,5]^n\).

Contained in the \sphinxstyleemphasis{separable - separable} function group.

% #1: function string (2 digits)
% #2: dimension string (2 digits)
% #3: instance string (2 digits)
% #4: function number
% #5: dimension
\bbobfigs{64}{05}{01}{64}{5}

\subsubsection[\texorpdfstring{\protect\(F_{65}\protect\): Attractive Sector/Step-ellipsoid}{F65: Attractive Sector/Step-ellipsoid}]{\texorpdfstring{\protect\(F_{65}\protect\): Attractive Sector/Step-ellipsoid}{}}
\label{index:f65}\label{index:attractive-sector-step-ellipsoid}
Combination of the Attractive Sector function (\href{http://coco.lri.fr/downloads/download15.03/bbobdocfunctions.pdf\#page=30}{\(f_6\) in the \sphinxcode{bbob} suite}) and the
Step Ellipsoidal function (\href{http://coco.lri.fr/downloads/download15.03/bbobdocfunctions.pdf\#page=35}{\(f_7\) in the \sphinxcode{bbob} suite}).

Both objective functions are unimodal and of moderate conditioning.
The first objective function is highly asymmetric, where only one \emph{hypercone}
(with angular base area) with a volume of roughly \(1/2^n\) yields low function values.
The optimum of the first objective is located at the tip of this cone. This function can
be deceptive for cumulative step size adaptation. The second objective function consists
of many plateaus of different sizes. Apart from a small area close to the global optimum,
the gradient is zero almost everywhere.

Contained in the \sphinxstyleemphasis{moderate - moderate} function group.

% #1: function string (2 digits)
% #2: dimension string (2 digits)
% #3: instance string (2 digits)
% #4: function number
% #5: dimension
\bbobfigs{65}{05}{01}{65}{5}

\subsubsection[\texorpdfstring{\protect\(F_{66}\protect\): Attractive Sector/rotated Rosenbrock}{F66: Attractive Sector/rotated Rosenbrock}]{\texorpdfstring{\protect\(F_{66}\protect\): Attractive Sector/rotated Rosenbrock}{}}
\label{index:f66}\label{index:attractive-sector-rotated-rosenbrock}
Combination of the Attractive Sector function (\href{http://coco.lri.fr/downloads/download15.03/bbobdocfunctions.pdf\#page=30}{\(f_6\) in the \sphinxcode{bbob} suite}) and the
rotated Rosenbrock function (\href{http://coco.lri.fr/downloads/download15.03/bbobdocfunctions.pdf\#page=45}{\(f_9\) in the \sphinxcode{bbob} suite}).

The first objective function is highly asymmetric, where only one \emph{hypercone}
(with angular base area) with a volume of roughly \(1/2^n\) yields low function values.
The optimum of the first objective is located at the tip of this cone.
The second objective function is the so-called banana function due to its 2-D contour lines
as a bent ridge (or valley) and partially separable (tri-band structure). In larger dimensions,
the second objective function has a local optimum with an attraction volume of about 25\%.
Note that, compared to the original Rosenbrock function, a rotation in the search space
is applied, such that the second objective function is non-separable.

Contained in the \sphinxstyleemphasis{moderate - moderate} function group.

% #1: function string (2 digits)
% #2: dimension string (2 digits)
% #3: instance string (2 digits)
% #4: function number
% #5: dimension
\bbobfigs{66}{05}{01}{66}{5}

\subsubsection[\texorpdfstring{\protect\(F_{67}\protect\): Step-ellipsoid/separable Rosenbrock}{F67: Step-ellipsoid/separable Rosenbrock}]{\texorpdfstring{\protect\(F_{67}\protect\): Step-ellipsoid/separable Rosenbrock}{}}
\label{index:f67}\label{index:step-ellipsoid-separable-rosenbrock}
Combination of the Step Ellipsoidal function (\href{http://coco.lri.fr/downloads/download15.03/bbobdocfunctions.pdf\#page=35}{\(f_7\) in the \sphinxcode{bbob} suite}) and the
separable Rosenbrock function (\href{http://coco.lri.fr/downloads/download15.03/bbobdocfunctions.pdf\#page=40}{\(f_8\) in the \sphinxcode{bbob} suite}).

The first objective function is unimodal, non-separable, and has a
conditioning of about 100. It actually consists of many plateaus of different sizes.
Apart from a small area close to the global optimum, the gradient is zero almost everywhere.
The second objective function is the so-called banana function due to its 2-D contour lines
as a bent ridge (or valley). It is partially separable (tri-band structure) and
in larger dimensions, the function has a local optimum with an attraction volume of about 25\%.

Contained in the \sphinxstyleemphasis{moderate - moderate} function group.

% #1: function string (2 digits)
% #2: dimension string (2 digits)
% #3: instance string (2 digits)
% #4: function number
% #5: dimension
\bbobfigs{67}{05}{01}{67}{5}

\subsubsection[\texorpdfstring{\protect\(F_{68}\protect\): Step-ellipsoid/rotated Rosenbrock}{F68: Step-ellipsoid/rotated Rosenbrock}]{\texorpdfstring{\protect\(F_{68}\protect\): Step-ellipsoid/rotated Rosenbrock}{}}
\label{index:step-ellipsoid-rotated-rosenbrock}\label{index:f68}
Combination of the Step Ellipsoidal function (\href{http://coco.lri.fr/downloads/download15.03/bbobdocfunctions.pdf\#page=35}{\(f_7\) in the \sphinxcode{bbob} suite}) and the
rotated Rosenbrock function (\href{http://coco.lri.fr/downloads/download15.03/bbobdocfunctions.pdf\#page=45}{\(f_9\) in the \sphinxcode{bbob} suite}).

The first objective function is unimodal, non-separable, and has a
conditioning of about 100. It actually consists of many plateaus of different sizes.
Apart from a small area close to the global optimum, the gradient is zero almost everywhere.
The second objective function is a rotated version of the original
so-called banana function (due to its 2-D contour lines as a bent ridge or valley) and
in larger dimensions, has a local optimum with an attraction volume of about 25\%.

This function resembles \(F_{67}\) except for the additional search space
rotation for the second objective function which makes both objective
function fully non-separable.

Contained in the \sphinxstyleemphasis{moderate - moderate} function group.

% #1: function string (2 digits)
% #2: dimension string (2 digits)
% #3: instance string (2 digits)
% #4: function number
% #5: dimension
\bbobfigs{68}{05}{01}{68}{5}

\subsubsection[\texorpdfstring{\protect\(F_{69}\protect\): separable Rosenbrock/rotated Rosenbrock}{F69: separable Rosenbrock/rotated Rosenbrock}]{\texorpdfstring{\protect\(F_{69}\protect\): separable Rosenbrock/rotated Rosenbrock}{}}
\label{index:separable-rosenbrock-rotated-rosenbrock}\label{index:f69}
Combination of the separable Rosenbrock function (\href{http://coco.lri.fr/downloads/download15.03/bbobdocfunctions.pdf\#page=40}{\(f_8\) in the \sphinxcode{bbob} suite}) and the
rotated Rosenbrock function (\href{http://coco.lri.fr/downloads/download15.03/bbobdocfunctions.pdf\#page=45}{\(f_9\) in the \sphinxcode{bbob} suite}).

Both objective functions are Rosenbrock functions (also known under the name
banana function due to its 2-D contour lines forming a bent ridge or valley)
with a local optimum in large dimension that has about 25\% attraction volume.
The first objective function is partially separable while the second objective
function is fully non-separable.

Contained in the \sphinxstyleemphasis{moderate - moderate} function group.

% #1: function string (2 digits)
% #2: dimension string (2 digits)
% #3: instance string (2 digits)
% #4: function number
% #5: dimension
\bbobfigs{69}{05}{01}{69}{5}

\subsubsection[\texorpdfstring{\protect\(F_{70}\protect\): Ellipsoid/Discus}{F70: Ellipsoid/Discus}]{\texorpdfstring{\protect\(F_{70}\protect\): Ellipsoid/Discus}{}}
\label{index:ellipsoid-discus}\label{index:f70}
Combination of the Ellipsoid function (\href{http://coco.lri.fr/downloads/download15.03/bbobdocfunctions.pdf\#page=50}{\(f_{10}\) in the \sphinxcode{bbob} suite}) and the
Discus (or Tablet) function (\href{http://coco.lri.fr/downloads/download15.03/bbobdocfunctions.pdf\#page=55}{\(f_{11}\) in the \sphinxcode{bbob} suite}).

Both objective functions are globally quadratic (unimodal) ill-conditioned functions with
condition numbers of \(10^6\) with  smooth local irregularities. A single direction
in search space is a thousand times more sensitive than all others for the Discus function.

Contained in the \sphinxstyleemphasis{ill-conditioned - ill-conditioned} function group.

% #1: function string (2 digits)
% #2: dimension string (2 digits)
% #3: instance string (2 digits)
% #4: function number
% #5: dimension
\bbobfigs{70}{05}{01}{70}{5}

\subsubsection[\texorpdfstring{\protect\(F_{71}\protect\): Ellipsoid/Bent Cigar}{F71: Ellipsoid/Bent Cigar}]{\texorpdfstring{\protect\(F_{71}\protect\): Ellipsoid/Bent Cigar}{}}
\label{index:ellipsoid-bent-cigar}\label{index:f71}
Combination of the Ellipsoid function (\href{http://coco.lri.fr/downloads/download15.03/bbobdocfunctions.pdf\#page=50}{\(f_{10}\) in the \sphinxcode{bbob} suite}) and the
Bent Cigar function (\href{http://coco.lri.fr/downloads/download15.03/bbobdocfunctions.pdf\#page=60}{\(f_{12}\) in the \sphinxcode{bbob} suite}).

Both objective functions are unimodal, non-separable, and have a conditioning
of about \(10^6\). The Ellipsoid function is globally quadratic with
smooth local irregularities while the Bent Cigar function deviates remarkably
from being quadratic due to an asymmetric transformation. To optimize the
Bent Cigar function, a smooth, but very narrow ridge has to be followed.

Contained in the \sphinxstyleemphasis{ill-conditioned - ill-conditioned} function group.

% #1: function string (2 digits)
% #2: dimension string (2 digits)
% #3: instance string (2 digits)
% #4: function number
% #5: dimension
\bbobfigs{71}{05}{01}{71}{5}

\subsubsection[\texorpdfstring{\protect\(F_{72}\protect\): Ellipsoid/Sharp Ridge}{F72: Ellipsoid/Sharp Ridge}]{\texorpdfstring{\protect\(F_{72}\protect\): Ellipsoid/Sharp Ridge}{}}
\label{index:ellipsoid-sharp-ridge}\label{index:f72}
Combination of the Ellipsoid function (\href{http://coco.lri.fr/downloads/download15.03/bbobdocfunctions.pdf\#page=50}{\(f_{10}\) in the \sphinxcode{bbob} suite}) and the
Sharp Ridge function (\href{http://coco.lri.fr/downloads/download15.03/bbobdocfunctions.pdf\#page=65}{\(f_{13}\) in the \sphinxcode{bbob} suite}).

Both objective functions are unimodal, non-separable, and have a conditioning
of about \(10^6\). Compared to the previous function, the ridge of the
here is sharp (non-differentiable) and the gradient remains constant, when the
ridge is approached from a given point. Approaching the ridge is
initially effective, but search behavior becomes difficult to diagnose
close to the ridge because the gradient towards the ridge does not flatten out.

Contained in the \sphinxstyleemphasis{ill-conditioned - ill-conditioned} function group.

% #1: function string (2 digits)
% #2: dimension string (2 digits)
% #3: instance string (2 digits)
% #4: function number
% #5: dimension
\bbobfigs{72}{05}{01}{72}{5}

\subsubsection[\texorpdfstring{\protect\(F_{73}\protect\): Ellipsoid/Sum of Different Powers}{F73: Ellipsoid/Sum of Different Powers}]{\texorpdfstring{\protect\(F_{73}\protect\): Ellipsoid/Sum of Different Powers}{}}
\label{index:ellipsoid-sum-of-different-powers}\label{index:f73}
Combination of the Ellipsoid function (\href{http://coco.lri.fr/downloads/download15.03/bbobdocfunctions.pdf\#page=50}{\(f_{10}\) in the \sphinxcode{bbob} suite}) and the
Sum of Different Powers function (\href{http://coco.lri.fr/downloads/download15.03/bbobdocfunctions.pdf\#page=70}{\(f_{14}\) in the \sphinxcode{bbob} suite}).

Both objective functions are unimodal and non-separable. While the
Ellipsoid function has a constant conditioning of \(10^6\) everywhere,
the sensitivities of the \(z_i\)-variables (in the rotated search space)
for the Different Powers function become more and more different when
approaching the optimum. The latter function has furthermore a small
solution volume.

Contained in the \sphinxstyleemphasis{ill-conditioned - ill-conditioned} function group.

% #1: function string (2 digits)
% #2: dimension string (2 digits)
% #3: instance string (2 digits)
% #4: function number
% #5: dimension
\bbobfigs{73}{05}{01}{73}{5}

\subsubsection[\texorpdfstring{\protect\(F_{74}\protect\): Discus/Bent Cigar}{F74: Discus/Bent Cigar}]{\texorpdfstring{\protect\(F_{74}\protect\): Discus/Bent Cigar}{}}
\label{index:f74}\label{index:discus-bent-cigar}
Combination of the Discus function (\href{http://coco.lri.fr/downloads/download15.03/bbobdocfunctions.pdf\#page=55}{\(f_{11}\) in the \sphinxcode{bbob} suite}) and the
Bent cigar function (\href{http://coco.lri.fr/downloads/download15.03/bbobdocfunctions.pdf\#page=60}{\(f_{12}\) in the \sphinxcode{bbob} suite}).

Both objective functions are unimodal, non-separable, and have a conditioning
of about \(10^6\).
The Discus function is globally quadratic with smooth local irregularities
and has a single direction in search space that is a thousand times more
sensitive than all others. The Bent Cigar function deviates remarkably
from being quadratic due to an asymmetric transformation and a smooth,
but very narrow ridge has to be followed to optimize it.

Contained in the \sphinxstyleemphasis{ill-conditioned - ill-conditioned} function group.

% #1: function string (2 digits)
% #2: dimension string (2 digits)
% #3: instance string (2 digits)
% #4: function number
% #5: dimension
\bbobfigs{74}{05}{01}{74}{5}

\subsubsection[\texorpdfstring{\protect\(F_{75}\protect\): Discus/Sharp Ridge}{F75: Discus/Sharp Ridge}]{\texorpdfstring{\protect\(F_{75}\protect\): Discus/Sharp Ridge}{}}
\label{index:discus-sharp-ridge}\label{index:f75}
Combination of the Discus function (\href{http://coco.lri.fr/downloads/download15.03/bbobdocfunctions.pdf\#page=55}{\(f_{11}\) in the \sphinxcode{bbob} suite}) and the
Sharp Ridge function (\href{http://coco.lri.fr/downloads/download15.03/bbobdocfunctions.pdf\#page=65}{\(f_{13}\) in the \sphinxcode{bbob} suite}).

Both objective functions are unimodal, non-separable, and have a conditioning
of about \(10^6\).
The Discus function is globally quadratic with smooth local irregularities
and has a single direction in search space that is a thousand times more
sensitive than all others.
To optimize the Sharp Ridge function, a sharp (i.e. non-differentiable) ridge
has to be followed around which the gradient remains constant, when the
ridge is approached from a given point. Approaching the ridge is
initially effective, but search behavior becomes difficult to diagnose
close to the ridge because the gradient towards the ridge does not flatten out.

Contained in the \sphinxstyleemphasis{ill-conditioned - ill-conditioned} function group.

% #1: function string (2 digits)
% #2: dimension string (2 digits)
% #3: instance string (2 digits)
% #4: function number
% #5: dimension
\bbobfigs{75}{05}{01}{75}{5}

\subsubsection[\texorpdfstring{\protect\(F_{76}\protect\): Discus/Sum of Different Powers}{F76: Discus/Sum of Different Powers}]{\texorpdfstring{\protect\(F_{76}\protect\): Discus/Sum of Different Powers}{}}
\label{index:f76}\label{index:discus-sum-of-different-powers}
Combination of the Discus function (\href{http://coco.lri.fr/downloads/download15.03/bbobdocfunctions.pdf\#page=55}{\(f_{11}\) in the \sphinxcode{bbob} suite}) and the
Sum of Different Powers function (\href{http://coco.lri.fr/downloads/download15.03/bbobdocfunctions.pdf\#page=70}{\(f_{14}\) in the \sphinxcode{bbob} suite}).

Both objective functions are unimodal and non-separable. While the
globally quadratic Discus function has a constant conditioning
of about \(10^6\) everywhere
with a single direction in search space that is a thousand times more
sensitive than all others, the sensitivities of the
\(z_i\)-variables (in the rotated search space)
for the Different Powers function become more and more different when
approaching the optimum. The latter function has furthermore a small
solution volume.

Contained in the \sphinxstyleemphasis{ill-conditioned - ill-conditioned} function group.

% #1: function string (2 digits)
% #2: dimension string (2 digits)
% #3: instance string (2 digits)
% #4: function number
% #5: dimension
\bbobfigs{76}{05}{01}{76}{5}

\subsubsection[\texorpdfstring{\protect\(F_{77}\protect\): Bent Cigar/Sharp Ridge}{F77: Bent Cigar/Sharp Ridge}]{\texorpdfstring{\protect\(F_{77}\protect\): Bent Cigar/Sharp Ridge}{}}
\label{index:bent-cigar-sharp-ridge}\label{index:f77}
Combination of the Bent Cigar function (\href{http://coco.lri.fr/downloads/download15.03/bbobdocfunctions.pdf\#page=60}{\(f_{12}\) in the \sphinxcode{bbob} suite}) and the
Sharp Ridge function (\href{http://coco.lri.fr/downloads/download15.03/bbobdocfunctions.pdf\#page=65}{\(f_{13}\) in the \sphinxcode{bbob} suite}).

Both objective functions are unimodal, non-separable, and have a conditioning
of about \(10^6\).
The Bent Cigar function deviates remarkably
from being quadratic due to an asymmetric transformation and a smooth,
but very narrow ridge has to be followed to optimize it.
To optimize the Sharp Ridge function, in turn, the ridge to be followed
is even sharper (i.e. non-differentiable), around which the gradient remains
constant, when the ridge is approached from a given point. Approaching the ridge is
initially effective, but search behavior becomes difficult to diagnose
close to the ridge because the gradient towards the ridge does not flatten out.

Contained in the \sphinxstyleemphasis{ill-conditioned - ill-conditioned} function group.

% #1: function string (2 digits)
% #2: dimension string (2 digits)
% #3: instance string (2 digits)
% #4: function number
% #5: dimension
\bbobfigs{77}{05}{01}{77}{5}

\subsubsection[\texorpdfstring{\protect\(F_{78}\protect\): Bent Cigar/Sum of Different Powers}{F78: Bent Cigar/Sum of Different Powers}]{\texorpdfstring{\protect\(F_{78}\protect\): Bent Cigar/Sum of Different Powers}{}}
\label{index:f78}\label{index:bent-cigar-sum-of-different-powers}
Combination of the Bent Cigar function (\href{http://coco.lri.fr/downloads/download15.03/bbobdocfunctions.pdf\#page=60}{\(f_{12}\) in the \sphinxcode{bbob} suite}) and the
Sum of Different Powers function (\href{http://coco.lri.fr/downloads/download15.03/bbobdocfunctions.pdf\#page=70}{\(f_{14}\) in the \sphinxcode{bbob} suite}).

Both objective functions are unimodal, non-separable, and have a conditioning
of about \(10^6\).

Both objective functions are unimodal and non-separable but differ in the
difficulties provided to an optimization algorithm.
The Bent Cigar function, on the one hand, deviates remarkably
from being quadratic due to an asymmetric transformation and a smooth,
but very narrow ridge has to be followed to optimize it.
The sensitivities of the
\(z_i\)-variables (in the rotated search space)
for the Different Powers function, on the other hand,
become more and more different when
approaching the optimum.

Contained in the \sphinxstyleemphasis{ill-conditioned - ill-conditioned} function group.

% #1: function string (2 digits)
% #2: dimension string (2 digits)
% #3: instance string (2 digits)
% #4: function number
% #5: dimension
\bbobfigs{78}{05}{01}{78}{5}

\subsubsection[\texorpdfstring{\protect\(F_{79}\protect\): Rastrigin/Schaffer F7 with conditioning of 1000}{F79: Rastrigin/Schaffer F7 with conditioning of 1000}]{\texorpdfstring{\protect\(F_{79}\protect\): Rastrigin/Schaffer F7 with conditioning of 1000}{}}
\label{index:rastrigin-schaffer-f7-with-conditioning-of-1000}\label{index:f79}
Combination of the Rastrigin function (\href{http://coco.lri.fr/downloads/download15.03/bbobdocfunctions.pdf\#page=75}{\(f_{15}\) in the \sphinxcode{bbob} suite}) and the
Schaffer F7 function with conditioning 1000 (\href{http://coco.lri.fr/downloads/download15.03/bbobdocfunctions.pdf\#page=90}{\(f_{18}\) in the \sphinxcode{bbob} suite}).

Both objective functions are non-separable and highly multimodal.
The problem's Rastrigin function alleviates the symmetry and regularity of
the originally proposed Rastrigin function via asymmetric and oscillating
transformations of the search space. It has roughly \(10^n\) local optima,
a low conditioning, and the global amplitude of function values is large
compared to the local amplitudes.
On the contrary, frequency  and  amplitude  of  the  function value modulation
vary for the Schaffer F7 function. It is furthermore asymmetric as well
but, compared to the other objective function is moderately ill-conditioned
with a conditioning of 1000.

Contained in the \sphinxstyleemphasis{multi-modal - multi-modal} function group.

% #1: function string (2 digits)
% #2: dimension string (2 digits)
% #3: instance string (2 digits)
% #4: function number
% #5: dimension
\bbobfigs{79}{05}{01}{79}{5}

\subsubsection[\texorpdfstring{\protect\(F_{80}\protect\): Rastrigin/Griewank-Rosenbrock}{F80: Rastrigin/Griewank-Rosenbrock}]{\texorpdfstring{\protect\(F_{80}\protect\): Rastrigin/Griewank-Rosenbrock}{}}
\label{index:rastrigin-griewank-rosenbrock}\label{index:f80}
Combination of the Rastrigin function (\href{http://coco.lri.fr/downloads/download15.03/bbobdocfunctions.pdf\#page=75}{\(f_{15}\) in the \sphinxcode{bbob} suite}) and the
Griewank-Rosenbrock function (\href{http://coco.lri.fr/downloads/download15.03/bbobdocfunctions.pdf\#page=95}{\(f_{19}\) in the \sphinxcode{bbob} suite}).

Both objective functions are non-separable and highly multimodal.
Both objective functions furthermore are variants of the original Rosenbrock
function:
The problem's Rastrigin function alleviates the symmetry and regularity of
the originally proposed Rastrigin function via asymmetric and oscillating
transformations of the search space. The Griewank-Rosenbrock function
resembles the original Rosenbrock function in a highly multimodal way.

Contained in the \sphinxstyleemphasis{multi-modal multi-modal} function group.

% #1: function string (2 digits)
% #2: dimension string (2 digits)
% #3: instance string (2 digits)
% #4: function number
% #5: dimension
\bbobfigs{80}{05}{01}{80}{5}

\subsubsection[\texorpdfstring{\protect\(F_{81}\protect\): Schaffer F7/Schaffer F7 with conditioning 1000}{F81: Schaffer F7/Schaffer F7 with conditioning 1000}]{\texorpdfstring{\protect\(F_{81}\protect\): Schaffer F7/Schaffer F7 with conditioning 1000}{}}
\label{index:schaffer-f7-schaffer-f7-with-conditioning-1000}\label{index:f81}
Combination of the Schaffer F7 function (\href{http://coco.lri.fr/downloads/download15.03/bbobdocfunctions.pdf\#page=85}{\(f_{17}\) in the \sphinxcode{bbob} suite}) and the
Schaffer F7 with conditioning 1000 function (\href{http://coco.lri.fr/downloads/download15.03/bbobdocfunctions.pdf\#page=90}{\(f_{18}\) in the \sphinxcode{bbob} suite}).

Both objective functions are of the same type (asymmetric, non-separable,
highly multimodal where frequency and amplitude of the modulation vary).
The main difference is in the conditioning, which is about 10 in one case
and 1000 in the other.

Contained in the \sphinxstyleemphasis{multi-modal - multi-modal} function group.

% #1: function string (2 digits)
% #2: dimension string (2 digits)
% #3: instance string (2 digits)
% #4: function number
% #5: dimension
\bbobfigs{81}{05}{01}{81}{5}

\subsubsection[\texorpdfstring{\protect\(F_{82}\protect\): Schaffer F7/Griewank-Rosenbrock}{F82: Schaffer F7/Griewank-Rosenbrock}]{\texorpdfstring{\protect\(F_{82}\protect\): Schaffer F7/Griewank-Rosenbrock}{}}
\label{index:schaffer-f7-griewank-rosenbrock}\label{index:f82}
Combination of the Schaffer F7 function (\href{http://coco.lri.fr/downloads/download15.03/bbobdocfunctions.pdf\#page=85}{\(f_{17}\) in the \sphinxcode{bbob} suite}) and the
Griewank-Rosenbrock function (\href{http://coco.lri.fr/downloads/download15.03/bbobdocfunctions.pdf\#page=95}{\(f_{19}\) in the \sphinxcode{bbob} suite}).

Both objective functions are non-separable and highly multimodal.
For the asymmetric Schaffer F7 function, frequency and amplitude of
the function value modulation vary and it has a low conditioning of
about 10. The Griewank-Rosenbrock function
resembles the original Rosenbrock function in a highly multimodal way.

Contained in the \sphinxstyleemphasis{multi-modal - multi-modal} function group.

% #1: function string (2 digits)
% #2: dimension string (2 digits)
% #3: instance string (2 digits)
% #4: function number
% #5: dimension
\bbobfigs{82}{05}{01}{82}{5}

\subsubsection[\texorpdfstring{\protect\(F_{83}\protect\): Schaffer F7 with conditioning 1000/Griewank-Rosenbrock}{F83: Schaffer F7 with conditioning 1000/Griewank-Rosenbrock}]{\texorpdfstring{\protect\(F_{83}\protect\): Schaffer F7 with conditioning 1000/Griewank-Rosenbrock}{}}
\label{index:schaffer-f7-with-conditioning-1000-griewank-rosenbrock}\label{index:f83}
Combination of the Schaffer F7 function with conditioning 1000 (\href{http://coco.lri.fr/downloads/download15.03/bbobdocfunctions.pdf\#page=90}{\(f_{18}\) in the \sphinxcode{bbob} suite}) and the
Griewank-Rosenbrock function (\href{http://coco.lri.fr/downloads/download15.03/bbobdocfunctions.pdf\#page=95}{\(f_{19}\) in the \sphinxcode{bbob} suite}).

Compared to \(F_{82}\), the only difference is the higher condition number of
about 1000 (compared to 10) in the Schaffer F7 function.

Contained in the \sphinxstyleemphasis{multi-modal - multi-modal} function group.

% #1: function string (2 digits)
% #2: dimension string (2 digits)
% #3: instance string (2 digits)
% #4: function number
% #5: dimension
\bbobfigs{83}{05}{01}{83}{5}

\subsubsection[\texorpdfstring{\protect\(F_{84}\protect\): Schwefel/Gallagher 21}{F84: Schwefel/Gallagher 21}]{\texorpdfstring{\protect\(F_{84}\protect\): Schwefel/Gallagher 21}{}}
\label{index:f84}\label{index:schwefel-gallagher-21}
Combination of the Schwefel function (\href{http://coco.lri.fr/downloads/download15.03/bbobdocfunctions.pdf\#page=100}{\(f_{20}\) in the \sphinxcode{bbob} suite}) and the
Gallagher 21 function (\href{http://coco.lri.fr/downloads/download15.03/bbobdocfunctions.pdf\#page=110}{\(f_{22}\) in the \sphinxcode{bbob} suite}).

Both objective functions are multi-modal with only a weak global structure.
The most prominent \(2^n\) minima of the Schwefel function
are located comparatively close to the corners of
the unpenalized search area. The penalization is essential, as otherwise more
and better minima occur further away from the search space origin. The function
is furthermore partially separable, a kind of combinatorial problem, and has two
search regimes. The Gallagher function consists of 21 optima with position and
height being unrelated and randomly chosen (different for each instantiation
of the function). The conditioning of the Gallagher function around the global
optimum is about 1000.

Contained in the \sphinxstyleemphasis{weakly-structured - weakly-structured} function group.

% #1: function string (2 digits)
% #2: dimension string (2 digits)
% #3: instance string (2 digits)
% #4: function number
% #5: dimension
\bbobfigs{84}{05}{01}{84}{5}

\subsubsection[\texorpdfstring{\protect\(F_{85}\protect\): Schwefel/Katsuuras}{F85: Schwefel/Katsuuras}]{\texorpdfstring{\protect\(F_{85}\protect\): Schwefel/Katsuuras}{}}
\label{index:f85}\label{index:schwefel-katsuuras}
Combination of the Schwefel function (\href{http://coco.lri.fr/downloads/download15.03/bbobdocfunctions.pdf\#page=100}{\(f_{20}\) in the \sphinxcode{bbob} suite}) and the
Katsuuras function (\href{http://coco.lri.fr/downloads/download15.03/bbobdocfunctions.pdf\#page=115}{\(f_{23}\) in the \sphinxcode{bbob} suite}).

Both objective functions are highly multi-modal with an exponential number
(in the dimension) of (global) optima and only a weak global structure.
The most prominent \(2^n\) minima of the Schwefel function
are located comparatively close to the corners of
the unpenalized search area.
The Katsuuras function, in turn, is highly repetitive with more than \(10^n\)
global optima.

% #1: function string (2 digits)
% #2: dimension string (2 digits)
% #3: instance string (2 digits)
% #4: function number
% #5: dimension
\bbobfigs{85}{05}{01}{85}{5}

\subsubsection[\texorpdfstring{\protect\(F_{86}\protect\): Schwefel/Lunacek bi-Rastrigin}{F86: Schwefel/Lunacek bi-Rastrigin}]{\texorpdfstring{\protect\(F_{86}\protect\): Schwefel/Lunacek bi-Rastrigin}{}}
\label{index:f86}\label{index:schwefel-lunacek-bi-rastrigin}
Combination of the Schwefel function (\href{http://coco.lri.fr/downloads/download15.03/bbobdocfunctions.pdf\#page=100}{\(f_{20}\) in the \sphinxcode{bbob} suite}) and the
Lunacek bi-Rastrigin function (\href{http://coco.lri.fr/downloads/download15.03/bbobdocfunctions.pdf\#page=120}{\(f_{24}\) in the \sphinxcode{bbob} suite}).

Both objective functions are highly multi-modal with only a weak global structure.
While the most prominent \(2^n\) minima of the Schwefel function
are located comparatively close to the corners of the unpenalized search area,
the Lunacek bi-Rastrigin function has two superimposed funnels. Presumably,
different approaches need to be used for ``selecting the funnel''
and for searching the highly multimodal function ``within'' the funnel.
The single-objective Lunacek bi-Rastrigin function was constructed
to be deceptive for some evolutionary algorithms with large population size.

Contained in the \sphinxstyleemphasis{weakly-structure - weakly-structured} function group.

% #1: function string (2 digits)
% #2: dimension string (2 digits)
% #3: instance string (2 digits)
% #4: function number
% #5: dimension
\bbobfigs{86}{05}{01}{86}{5}

\subsubsection[\texorpdfstring{\protect\(F_{87}\protect\): Gallagher 101/Gallagher 21}{F87: Gallagher 101/Gallagher 21}]{\texorpdfstring{\protect\(F_{87}\protect\): Gallagher 101/Gallagher 21}{}}
\label{index:f87}\label{index:gallagher-101-gallagher-21}
Combination of Gallagher’s Gaussian 101-me Peaks function (\href{http://coco.lri.fr/downloads/download15.03/bbobdocfunctions.pdf\#page=105}{\(f_{21}\) in the \sphinxcode{bbob} suite}) and the
Gallagher’s Gaussian 21-hi Peaks function (\href{http://coco.lri.fr/downloads/download15.03/bbobdocfunctions.pdf\#page=110}{\(f_{22}\) in the \sphinxcode{bbob} suite}).

Both objective functions are multi-modal and non-separable. Both consist of a set of
optima with position and height being unrelated and randomly chosen. The number of
optima is 101 and 21 respectively and the condition number around the (unique) global
optima are about 30 and about 1000 respectively.

Contained in the \sphinxstyleemphasis{weakly-structured - weakly-structured} function group.

% #1: function string (2 digits)
% #2: dimension string (2 digits)
% #3: instance string (2 digits)
% #4: function number
% #5: dimension
\bbobfigs{87}{05}{01}{87}{5}

\subsubsection[\texorpdfstring{\protect\(F_{88}\protect\): Gallagher 101/Katsuuras}{F88: Gallagher 101/Katsuuras}]{\texorpdfstring{\protect\(F_{88}\protect\): Gallagher 101/Katsuuras}{}}
\label{index:f88}\label{index:gallagher-101-katsuuras}
Combination of Gallagher’s Gaussian 101-me Peaks function (\href{http://coco.lri.fr/downloads/download15.03/bbobdocfunctions.pdf\#page=105}{\(f_{21}\) in the \sphinxcode{bbob} suite}) and the
Katsuuras function (\href{http://coco.lri.fr/downloads/download15.03/bbobdocfunctions.pdf\#page=115}{\(f_{23}\) in the \sphinxcode{bbob} suite}).

Both objective functions are non-separable and highly multi-modal with
only a weak global structure. Gallagher's Gaussian 101-me Peaks
function consists of a set of 101 optima with position and height being
unrelated and randomly chosen. The conditioning is about 30.
The Katsuuras function, in turn, is highly repetitive with more than \(10^n\)
global optima.

Contained in the \sphinxstyleemphasis{weakly-structured - weakly-structured} function group.

% #1: function string (2 digits)
% #2: dimension string (2 digits)
% #3: instance string (2 digits)
% #4: function number
% #5: dimension
\bbobfigs{88}{05}{01}{88}{5}

\subsubsection[\texorpdfstring{\protect\(F_{89}\protect\): Gallagher 101/Lunacek bi-Rastrigin}{F89: Gallagher 101/Lunacek bi-Rastrigin}]{\texorpdfstring{\protect\(F_{89}\protect\): Gallagher 101/Lunacek bi-Rastrigin}{}}
\label{index:gallagher-101-lunacek-bi-rastrigin}\label{index:f89}
Combination of Gallagher’s Gaussian 101-me Peaks function (\href{http://coco.lri.fr/downloads/download15.03/bbobdocfunctions.pdf\#page=105}{\(f_{21}\) in the \sphinxcode{bbob} suite}) and the
Lunacek bi-Rastrigin function (\href{http://coco.lri.fr/downloads/download15.03/bbobdocfunctions.pdf\#page=120}{\(f_{24}\) in the \sphinxcode{bbob} suite}).

Both objective functions are non-separable and highly multi-modal with
only a weak global structure. Gallagher's Gaussian 101-me Peaks
function consists of a set of 101 optima with position and height being
unrelated and randomly chosen. The conditioning is about 30.
The Lunacek bi-Rastrigin function has two superimposed funnels. Presumably,
different approaches need to be used for ``selecting the funnel''
and for searching the highly multimodal function ``within'' the funnel.
The single-objective Lunacek bi-Rastrigin function was constructed
to be deceptive for some evolutionary algorithms with large population size.

Contained in the \sphinxstyleemphasis{weakly-structured - weakly-structured} function group.

% #1: function string (2 digits)
% #2: dimension string (2 digits)
% #3: instance string (2 digits)
% #4: function number
% #5: dimension
\bbobfigs{89}{05}{01}{89}{5}

\subsubsection[\texorpdfstring{\protect\(F_{90}\protect\): Gallagher 21/Katsuuras}{F90: Gallagher 21/Katsuuras}]{\texorpdfstring{\protect\(F_{90}\protect\): Gallagher 21/Katsuuras}{}}
\label{index:gallagher-21-katsuuras}\label{index:f90}
Combination of Gallagher’s Gaussian 21-hi Peaks function (\href{http://coco.lri.fr/downloads/download15.03/bbobdocfunctions.pdf\#page=110}{\(f_{22}\) in the \sphinxcode{bbob} suite}) and the
Katsuuras function (\href{http://coco.lri.fr/downloads/download15.03/bbobdocfunctions.pdf\#page=115}{\(f_{23}\) in the \sphinxcode{bbob} suite}).

Both objective functions are non-separable and multi-modal with
only a weak global structure. Gallagher's Gaussian 21-hi Peaks
function consists of a set of 21 optima with position and height being
unrelated and randomly chosen. The conditioning is about 1000.
The Katsuuras function, in turn, is highly repetitive with more than \(10^n\)
global optima.

Contained in the \sphinxstyleemphasis{weakly-structured - weakly-structured} function group.

% #1: function string (2 digits)
% #2: dimension string (2 digits)
% #3: instance string (2 digits)
% #4: function number
% #5: dimension
\bbobfigs{90}{05}{01}{90}{5}

\subsubsection[\texorpdfstring{\protect\(F_{91}\protect\): Gallagher 21/Lunacek bi-Rastrigin}{F91: Gallagher 21/Lunacek bi-Rastrigin}]{\texorpdfstring{\protect\(F_{91}\protect\): Gallagher 21/Lunacek bi-Rastrigin}{}}
\label{index:f91}\label{index:gallagher-21-lunacek-bi-rastrigin}
Combination of Gallagher’s Gaussian 21-hi Peaks function (\href{http://coco.lri.fr/downloads/download15.03/bbobdocfunctions.pdf\#page=110}{\(f_{22}\) in the \sphinxcode{bbob} suite}) and the
Lunacek bi-Rastrigin function (\href{http://coco.lri.fr/downloads/download15.03/bbobdocfunctions.pdf\#page=120}{\(f_{24}\) in the \sphinxcode{bbob} suite}).

Both objective functions are non-separable and multi-modal with
only a weak global structure. Gallagher's Gaussian 21-hi Peaks
function consists of a set of 21 optima with position and height being
unrelated and randomly chosen. The conditioning is about 1000.
The Lunacek bi-Rastrigin function has two superimposed funnels. Presumably,
different approaches need to be used for ``selecting the funnel''
and for searching the highly multimodal function ``within'' the funnel.
The single-objective Lunacek bi-Rastrigin function was constructed
to be deceptive for some evolutionary algorithms with large population size.

Contained in the \sphinxstyleemphasis{weakly-structured - weakly-structured} function group.

% #1: function string (2 digits)
% #2: dimension string (2 digits)
% #3: instance string (2 digits)
% #4: function number
% #5: dimension
\bbobfigs{91}{05}{01}{91}{5}

\subsubsection[\texorpdfstring{\protect\(F_{92}\protect\): Katsuuras/Lunacek bi-Rastrigin}{F92: Katsuuras/Lunacek bi-Rastrigin}]{\texorpdfstring{\protect\(F_{92}\protect\): Katsuuras/Lunacek bi-Rastrigin}{}}
\label{index:f92}\label{index:katsuuras-lunacek-bi-rastrigin}
Combination of the Katsuuras function (\href{http://coco.lri.fr/downloads/download15.03/bbobdocfunctions.pdf\#page=115}{\(f_{23}\) in the \sphinxcode{bbob} suite}) and the
Lunacek bi-Rastrigin function (\href{http://coco.lri.fr/downloads/download15.03/bbobdocfunctions.pdf\#page=120}{\(f_{24}\) in the \sphinxcode{bbob} suite}).

Both objective functions are non-separable and highly multi-modal with
only a weak global structure.
The Katsuuras function is highly repetitive with more than \(10^n\)
global optima.
The Lunacek bi-Rastrigin function has two superimposed funnels. Presumably,
different approaches need to be used for ``selecting the funnel''
and for searching the highly multimodal function ``within'' the funnel.
The single-objective Lunacek bi-Rastrigin function was constructed
to be deceptive for some evolutionary algorithms with large population size.

Contained in the \sphinxstyleemphasis{weakly-structured - weakly-structured} function group.

% #1: function string (2 digits)
% #2: dimension string (2 digits)
% #3: instance string (2 digits)
% #4: function number
% #5: dimension
\bbobfigs{92}{05}{01}{92}{5}

\end{document}